\documentclass[examplefnt,biber]{nowfnt} % creates the journal version, needs biber version
\usepackage[utf8]{inputenc}

\usepackage{amsmath}
\usepackage{lscape}
\usepackage{multirow}
\usepackage{graphicx}
\usepackage{amsfonts}

\title{Multi-hop Question Answering}

\maintitleauthorlist{
Vaibhav Mavi \\
New York University, United States of America \\
vaibhavg152@gmail.com
\and
Anubhav Jangra \\
Indian Institute of Technology Patna, India \\
anubhav0603@gmail.com
\and
Adam Jatowt \\
University of Innsbruck, Austria \\
jatowt@acm.org
}

\issuesetup
{%
 copyrightowner={V.~Mavi and A~Jangra and A.~Jatowt},
 volume        = xx,
 issue         = xx,
 pubyear       = 2023,
 isbn          = xxx-x-xxxxx-xxx-x,
 eisbn         = xxx-x-xxxxx-xxx-x,
 doi           = 10.1561/XXXXXXXXX,
 firstpage     = 1, %Explain
 lastpage      = 75
 }

\usepackage{mwe}

\author[1]{Mavi,Vaibhav}
\author[2]{Jangra,Anubhav}
\author[3]{Jatowt,Adam}

\affil[1]{New York University, United States of America; vaibhavg152@gmail.com}
\affil[2]{Indian Institute of Technology Patna, India;  anubhav0603@gmail.com}
\affil[3]{University of Innsbruck, Austria;  jatowt@acm.org}

\articledatabox{\nowfntstandardcitation}

\begin{document}

\makeabstracttitle

\begin{abstract}
The task of Question Answering (QA) has attracted significant research interest for long. Its relevance to language understanding and knowledge retrieval tasks, along with the simple setting makes the task of QA crucial for strong AI systems. Recent success on simple QA tasks has shifted the focus to more complex settings. Among these, Multi-Hop QA (MHQA) is one of the most researched tasks over the recent years. In broad terms, MHQA is the task of answering natural language questions that involve extracting and combining multiple pieces of information and doing multiple steps of reasoning. An example of a multi-hop question would be ``The Argentine PGA Championship record holder has won how many tournaments worldwide?''. Answering the question would need two pieces of information: ``Who is the record holder for Argentine PGA Championship tournaments?'' and ``How many tournaments did [Answer of Sub Q1] win?''. The ability to answer multi-hop questions and perform multi step reasoning can significantly improve the utility of NLP systems. Consequently, the field has seen a surge with high quality datasets, models and evaluation strategies. The notion of `multiple hops' is somewhat abstract which results in a large variety of tasks that require multi-hop reasoning. This leads to different datasets and models that differ significantly from each other and makes the field challenging to generalize and survey. We aim to provide a general and formal definition of the MHQA task, and organize and summarize existing MHQA frameworks. We also outline some best practices for building MHQA datasets. This book provides a systematic and thorough introduction as well as the structuring of the existing attempts to this highly interesting, yet quite challenging task. 

\end{abstract}

\chapter{Introduction} \label{sec:intro}

\section{Question Answering}

An eventual goal of artificial intelligence (AI) is to impart the ability to reason over natural language to machines. In order to achieve this, several natural language understanding and generation tasks have been proposed that require an agent to do some reasoning to get to the goal. One such example is the task of Question Answering (QA) where given a question and some relevant context, the goal is to predict the correct answer. The question answering task provides a quantifiable way to evaluate a system’s capability of language understanding and reasoning \citep{SP25, squad, hermann2015teaching}. It is a critical problem in the fields of natural language processing (NLP) and information retrieval (IR), and a long-standing AI milestone. 

Abundance of readily-available, high-quality information on the internet facilitates the need of automated QA systems that help probe this rich content based on individual needs. Due to recent advancements in Deep Learning techniques \citep{albert}, the machines have become able to successfully beat human performance on datasets like SQUAD 2.0 \citep{squad2}. However, we have only scratched the surface of what these modern systems are capable of achieving. Depending on the user requirements, the complexity of QA tasks may vary. Some questions can be answered in brief (\textit{e.g.}, ``\textit{Which color do you get when you mix red and yellow paints?}'') - such questions are called \textit{objective questions} or \textit{factoid questions}. On the other hand, there exist \textit{subjective questions} that demand detailed explanations to meet user requirements (\textit{e.g.}, ``\textit{Why does mixing red, green and blue paints give black color paint, but projecting red, green, and blue light on a white surface return white light?}''). A question can also be considered complex, if it requires a very niche domain expertise to answer the question (\textit{e.g.}, ``\textit{What symptoms help diagnose chickenpox?}'').

% The more traditional work focuses on performing this reasoning over a single context \citep{SP25, bidaf, liu-etal-2018-stochastic, wang-etal-2017-gated}. 

\par

\section{What is Multi-hop Question Answering (MHQA)?}

For questions mentioned above, there might exist a single document or a single passage (formally referred as a `\textit{context}') that can provide a justifiable answer. However, there exists certain questions that cannot be answered using a single \textit{context} (\textit{e.g.}, ``\textit{What is the national bird of the nation that has a negative carbon footprint?}''). The task of answering such questions is called multi-hop question answering (MHQA). The goal of MHQA is to predict the correct answer to a question that requires multiple reasoning `hops' across given contexts (text, table, knowledge graph etc). We look at a more detailed definition of the task in Chapter \ref{sec:prob_def}.

The success in simple QA systems (also referred as \textit{single hop QA}) does not necessarily entail success of MHQA systems. \citet{min-etal-2018-efficient} and \citet{SP25} observe that most questions in existing single-hop QA datasets are answerable without much reasoning, by retrieving a small set of sentences. Moreover, multi-step reasoning is required by the models to answer complex questions (refer to Table \ref{tab:mhqa-examples}). Humans can easily perform these multi-step reasoning in their everyday tasks, yet this is still a difficult task for machines. 
% More challenging tasks such as complex QA require the models to perform multi-step reasoning. Humans can easily perform multi-step reasoning in their everyday tasks, yet this is still difficult for machines. 
An agent can be said to perform multi-step reasoning if it reaches one or more intermediate conclusions before deriving the final answer and each of the intermediate conclusions serves as a necessary premise for some other conclusion. This sequence of intermediate conclusions, including the final answer, is called a \textit{reasoning chain} and each step from one conclusion to the next can be referred to as a \textit{hop}.
\par 
\begin{table}[h]
\scriptsize
\caption{Examples of various types of multi-hop questions.} \label{tab:mhqa-examples}
\begin{tabular}{p{0.3\linewidth}|p{0.4\linewidth}|p{0.2\linewidth}}
\hline
\textbf{Type of question } & \textbf{Question}& \textbf{Answer}                 \\
\hline
\begin{tabular}[c]{@{}l@{}}Bridge Entity-based\\ (temporal entity)\end{tabular}     & \begin{tabular}[c]{@{}l@{}}Who was the president of United States\\ in the year in which Mike Tyson\\ declared his retirement?\end{tabular} & George W. Bush         \\[0.2cm]
\hline
\begin{tabular}[c]{@{}l@{}}Bridge Entity-based\\ (geographical entity)\end{tabular} & \begin{tabular}[c]{@{}l@{}}What is the national bird of the nation\\ that has a negative carbon footprint?\end{tabular}                     & The Raven              \\
\hline
\begin{tabular}[c]{@{}l@{}}Bridge Entity-based\\ (named entity)\end{tabular}        & \begin{tabular}[c]{@{}l@{}}What is the birth place of the tennis \\ player who has won the most grand slams?\end{tabular}                   & Belgrade, Serbia         \\
\hline
Intersection                                                                        & \begin{tabular}[c]{@{}l@{}}Who is the only person to win\\ an olympic medal and a Nobel prize?\end{tabular}                                 & Philip John Noel-Baker \\
\hline
Comparison                                                                          & \begin{tabular}[c]{@{}l@{}}Which country has won more\\ soccer world cups - Argentina or Brazil?\end{tabular}                               & Brazil \\
\hline
Commonsense Reasoning                                                                          & \begin{tabular}[c]{@{}l@{}}If A prefers fruits over meat,\\ when given an option of apple and\\ chicken sandwich, what will A prefer?\end{tabular}                               & Apple \\[0.2cm]
\hline
\end{tabular}
\end{table}

\begin{figure}[ht]
    \includegraphics[width=0.7\textwidth]{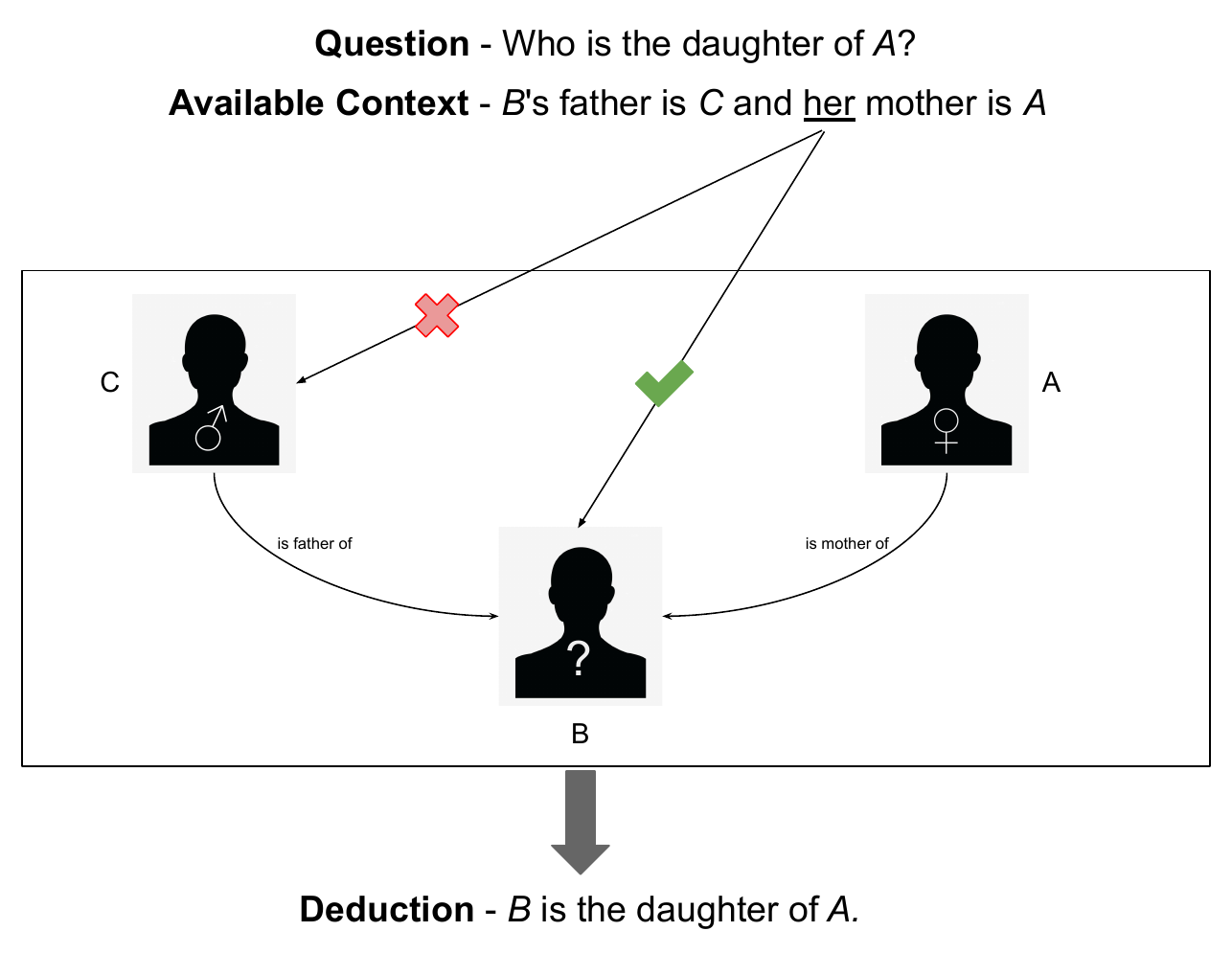}
    \centering
    \caption{An example of multi-hop reasoning}
    \label{fig:example}
\end{figure}

It is important to note that the inability of AI systems to perform multiple steps of reasoning can be severely limiting, significantly reducing their usability. One such instance can be as shown in Fig. \ref{fig:example}. Say a user is interested in knowing more about `the daughter of \textit{A}' and the only relevant information available in this context is `\textit{B}'s father is \textit{C} and her mother is \textit{A}'. In this case, the AI system has to first infer that \textit{B} is female and her mother is \textit{A}. The system will then have to use common sense reasoning to conclude that \textit{B} is the entity of interest and then retrieve the required information (refer to Fig. \ref{fig:example} for visual aid). Something like this seems trivial to humans but it may fatally confuse many existing AI systems. Therefore, we argue that multi-step reasoning is a crucial challenge and solving it can be a giant leap towards the goals of AI.
%\par

\section{Applications of MHQA}
As discussed above, MHQA serves as an appropriate benchmark task for evaluating an agent's ability to perform multi-step reasoning. Along with this scientific significance, the task of MHQA has various practical applications. Queries given to current web search systems can often require multi-hop reasoning to reach the relevant documents. 
User satisfaction when using such systems can be greatly improved by utilizing multi-hop reasoning models. Furthermore, conversations between humans and agents can be smoother and more informative if the latter can handle complex questions.
Answering a multi-hop question requires systems to aggregate information over multiple contexts. Therefore, techniques that are successful for MHQA can inspire progress in tasks such as sentence fusion \citep{weiss2021extending, geva2019discofuse} and abstractive summarization \citep{nayeem2018abstractive, lebanoff2019analyzing}, event occurrence time prediction \citep{wangSIGIR21}, as well as multi-document summarization \citep{ma2020multi, goldstein2000multi, haghighi2009exploring, barzilay1999information} or timeline summarization \citep{yan2011evolutionary, ghalandari2020examining, steen2019abstractive,yu-etal-2021-multi} that require information aggregation over multiple documents.
Additionally, most applications of QA such as information extraction (IE) and entailment, can be immensely benefited by multi-hop reasoning abilities \citep{qa-ie}.

\citet{QG1} argue that MHQA is a challenging task to an extent that they quantify the difficulty of a question as the number of inference steps (or hops) required to answer the question. This illustrates the direct utility of MHQA for the task of Difficulty controllable Question Generation (DQG) \citep{dqg} that has various applications including curriculum-learning based methods for QA systems \citep{qg4edu} and designing school exams of certain difficulty levels \citep{easyques1}.

\par
Another problem closely related to MHQA consists of generating clarifying questions for conversational QA (chatbots) \citep{SP22, zaib2021conversational}. In this setting, the original question/query can be ambiguous and hence more information is needed to disambiguate it. The model is supposed to generate a clarifying question in natural language, asking the user for the missing information. This can be considered as another task involving multi-step reasoning and can be greatly helped by improvements in MHQA.
\section{Overview}
Recently, a variety of datasets and techniques have been proposed for MHQA, including ones designed for MHQA over Knowledge Bases and Knowledge Graphs as well as those designed for QA over tables and text. A substantial number of recent works have focused on the task of MHQA and contributed to significant advancements. High quality datasets \citep{DP1, DP2, DP3, DP4, DP5, DP6, DP7} have encouraged better models to be proposed which in turn have achieved impressive accuracy on these benchmarks.
There has been a significant research in the recent years to solve the task. A variety of methods model the task as performing inference over static or dynamic graphs to find the reasoning paths \citep{SP1, SP6, SP8, SP12, SP13, SP15, SP16, SP25, SP20, SP26, SP21}. A number of works have also attempted to decompose the multi-hop questions into single hop questions or generate follow-up questions based on the retrieved information \citep{SP14, SP21, SP22, SP8, SP17}. The recent success of large language models (LLMs) has significantly influenced MHQA as well, with multiple attempts of using LLMs' strong natural understanding and emergent abilities for answering complex multi-hop questions \citep{llm-survey, LL2, LL3, LL33, LL13, LL14}. We discuss all these methods in a detailed and organized manner in Chapters \ref{sec:method}, \ref{sec:llm} and \ref{sec:tax}.
\par

Due to the surge in the attention received by the task over the last decade, we believe that the community would benefit from an extensive survey encompassing recent advancements in MHQA. In this work, we closely cover $\sim $75 works from top venues including but not limited to EMNLP, ACL, NAACL, TACL, AAAI, EACL, SIGIR, ICLR, COLING, CoRR etc. published from 2016 to 2024. 
The research community has already several surveys in the field of question-answering, such as for single-hop QA \citep{allam2012question, bouziane2015question, mishra2016survey, hoffner2017survey, soares2020literature, dimitrakis2020survey}, open-domain QA \citep{DBLP:series/synthesis/2021Roy, surveyodqa, zhu2021retrieving}, medical QA \citep{lin2021medical, jin2022biomedical}, visual QA \citep{visual-qa-survey, wu2017visual}, etc. 
The surveys that are most relevant to MHQA are the ones focused on QA over knowledge bases \citep{complex-survey-1, complex-survey-2, diefenbach2018core,DBLP:series/synthesis/2021Roy} and visual QA \citep{visual-qa-survey, lin2021medical, wu2017visual}. However, these can be considered as sub-domains of the more general formulation of the MHQA field that this book aims to survey. 
Since the existing works go a long way in summarizing their intended domains, we choose to exclude Visual MHQA and MHQA over Knowledge Bases and Knowledge Graphs from the scope of this work.
\par
We observe that despite the impressive accuracy of recent models on MHQA benchmarks,  significant concerns have been raised regarding whether the models are actually able to perform multi-step reasoning in order to answer the multi-hop questions. Several works \citep{AP1, AP2, AP3, AP4, AP5, AP6, AP7, AP8, SP32} conduct experiments and demonstrate that a significant portion of the accuracy can be ascribed to pattern matching and single step reasoning (also termed as \textit{shortcut reasoning}). This points to new challenges and future directions for research in MHQA. Above all, it is fair to say that despite the inspiring progress made so far, the task of MHQA is still a long way from being solved.
% creating good datasets for the task, since even the carefully created high quality datasets suffer from issues such as annotator bias and annotator errors. In order to minimize such errors and biases, we provide a concise set of annotation guidelines that the existing works have effectively used. We believe this set can be a useful reference while creating new good quality datasets. We also take a closer look at the above-mentioned findings which reveal weaknesses of current MHQA systems.
\par
A promising direction for solving some of these challenges is the task of explainable MHQA, a particular setting of MHQA that requires the model to output the correct reasoning chain (or equivalently, some kind of representation of the reasoning chain) along with the correct answer. This increases the model's accountability and interpretability to the end user since the model now has to also explain how it reached the answer. Interpretability of the AI systems is crucial for their wide adoption for most high-stake applications such as finance, law and healthcare \citep{samek2017explainable, alvarez-melis-jaakkola-2017-causal, arras2016relevant, biran2017explanation, gilpin2018explaining}. Consequently, more recent works \citep{SP18, SP19, DP1, AP7, AP6} have focused on this setting. \citet{DP1} have also argued that training the model to output reasoning chain can further help in training to predict the correct answer as it serves as a useful auxiliary task. \citet{SP32} also find that using the reasoning chain as a supervision signal during training improves the performance on adversarial examples as well.
\par
% The contributions of this manuscript can be summarized as:
% \begin{itemize}
%     \item To the best of our knowledge, our work is the first to survey the recent works in the domain of MHQA.
%     \item To the best of our knowledge, we are first to provide a formal definition for MHQA that is general enough to encompass existing variants of the task and also to promote new variants.
%     \item We propose a taxonomy of the existing works in MHQA that allows to systematically organize and structure the current works.
%     \item We provide a concise set of data annotation guidelines that existing works have adopted and found beneficial.
%     \item We propose candidate future directions for continuing MHQA research.
% \end{itemize}
% \par
The remainder of this book is structured as follows:
Chapter \ref{sec:prob_def} aims to formalize the task of MHQA in a way that encompasses most existing variants.
Chapter \ref{sec:data} describes existing MHQA datasets, their creation techniques, critiques and challenges\footnote{We discuss the datasets before methods as doing so provides on overview of the existing variants of the tasks which would be helpful to understand the intuition behind the proposed architectures.}.
Chapter \ref{sec:method} discusses traditional pre-LLM models in-depth in a structured way that leads to a taxonomy for existing methods in Chapter \ref{sec:tax}.
Chapter \ref{sec:llm} is dedicated to recent LLM based methods for MHQA, challenges of incorporating LLMs and their proposed solutions.
Chapter \ref{sec:eval} discusses the standard evaluation metrics along with evaluation methods specifically designed for evaluating multi-step reasoning/retrieval.
Chapter \ref{sec:mhqg} touches upon the multi-hop question generation problem.
Chapter \ref{sec:future} then summarizes the insights of the book and critiques of the existing methods and datasets, to propose promising directions for future research in MHQA.
\chapter{Formulating the Multi-Hop Question Answering Task} \label{sec:prob_def}
Before diving into the advancements, we try to provide a formal and descriptive definition of the task. Although many attempts at formally defining the task have been made, we observe that they tend to focus on specific cases of the broader task. However, we aim to cover a large variety of tasks that can be considered as variants of the MHQA task. Therefore, we try to propose a broader definition that encompasses many variants of the task that have been tackled. By doing so, we also aim to clearly define the scope of what concerns Multi-Hop Question Answering for the rest of this book.
\par
Formally defining the task of multi-hop question answering is not straightforward, since the definition of a \textit{hop} is ambiguous in itself. For instance, in the context of open-domain QA on text documents, a hop could signify reasoning across different documents \citep{DP1} whereas for QA over long documents, reasoning across different sections or paragraphs is a hop \citep{SP22}. To the best of our knowledge, existing works do not provide a general definition of the task that encompasses its different variants. We argue that in order to systematically tackle the problem, and to obtain a good understanding of the progress in MHQA, it is crucial to first have a general definition.
\par 
We attribute the generality of the MHQA by keeping the notion of a context abstract. Depending on the task, a context can be any single independent piece of information: a sentence, a document, an image or an entity in a knowledge graph. Keeping this in mind, we formally define the task of multi-hop question answering as:\par
Let $\mathbb{C}$ denote the set of all contexts, $\mathcal{S}$ denote the set of all questions and $\mathcal{A}$ denote the set of all possible answers. Given a question $q\in\mathcal{S}$ and a set of related contexts, $C\subseteq\mathbb{C}$, the task is to approximate a function $f: \mathcal{S}\times\mathbb {C}^n \mapsto \mathcal{A} \cup \{\Phi\}$, that satisfies:

\begin{equation} \label{eq:def}
        f(q, C) = 
        \begin{cases} 
            a\in \mathcal{A} & \exists~ P_q = \{p_1,\cdots, p_k\} \subseteq C, k>1 \\
            & \And P_q \models (a~ answers~ q) \\
            \Phi & otherwise 
   \end{cases}
\end{equation}
% \begin{gather}\label{eq:def}
%     f(q, C) = \begin{cases} 
%       a\in \mathcal{A} & \exists~ P_q = \{p_1,\cdots, p_k\} \subseteq C, k>1 \And P_q \models (a~ answers~ q) \\
%       \Phi & otherwise 
%    \end{cases}
% \end{gather}
% \begin{equation*}
       where~ $\models$ represents entailment, and $\Phi$ is the output when $q$ is unanswerable using $\mathbb{C}$\footnote{In case of multiple correct answers to $q$, Eq. \ref{eq:def} allows $f$ to output any $a$ that answers $q$, which is the case for datasets like SQuAD \citep{squad, squad2}. However, some applications may require $f$ to output all correct answers \citep{min2020ambigqa} when multiple correct answers exist.}.
% \end{equation*}
Given a question $q$ and a set of contexts $C$, $f$ returns an answer $a$ that answers $q$ by using a subset of `gold' supporting contexts $P$ from $C$. The number of gold supporting contexts $k$ is restricted to be more than 1 to ensure that the question is not solvable using a single hop ($k=1$ reduces the task to traditional QA). 
\par
This definition captures the commonly adopted breakdown of the task to two sub-problems: Information Retrieval (IR) and Reading Comprehension (RC). Typically, $f$ can be decomposed into (IR) $g: \mathcal{S}\times\mathbb{C}^n\mapsto\mathbb{C}^k$ and (RC) $h:\mathcal{S}\times\mathbb{C}^k\mapsto\mathcal{A}\cup\{\Phi\}$, for some $k\in \mathbb{N},~ k>1$ such that: 
\begin{gather}
    g(q, C) = P_q 
\end{gather}
where $P_q \subseteq C$ is the set of contexts relevant to $q$.
\begin{gather}
    h(q, P_q) = \begin{cases} a & P_q \models (a~ answers~ q) \\
      \Phi & otherwise 
   \end{cases} and \\
   f(q, C) = h(q, g(q, C))
\end{gather} 
\par
\textbf{Reasoning chain:} A reasoning chain for a question $P_q' = \{p'_{q,i}\}_{i=1}^k$ is defined as an ordered permutation of the set $P_q$ defined above, such that:
\begin{gather*}
    \forall~ j,~ 1\leq j<k,~ p'_{q,j} \xrightarrow{} p'_{q,j+1}\text{ represents the $j^{th}$ reasoning step and } \\
    p'_{q, k} \xrightarrow{} a \text{ is the $k^{th}$ reasoning step.}
\end{gather*}
It is important to note that the granularity of what constitutes a reasoning chain can also be smaller than that of the contexts. For example, when the granularity of a context is passage, the reasoning chain may consist of particular sentences or particular entities belonging to those passages. 
\par 
\textbf{Hop:} Each reasoning step of the reasoning chain can be termed as a hop. Furthermore, some commonsense knowledge might additionally be required to perform a reasoning step from one context (i.e., a document, table, etc.) to another. In that case, the commonsense reasoning can also be considered as a hop. The definition provided here does not consider reasoning hops over external/commonsense knowledge although it can be accommodated by allowing  $P_q \subseteq C \cup Q$, where $Q$ is the external/commonsense knowledge base. However, we omit this for simplicity.
\par
As mentioned in the introduction, many recent works focus on \textbf{explainable MHQA} to ensure the accountability and interpretability of the models while answering multi-hop questions. Formally, explainable MHQA is the setting of MHQA that requires $f$ to output the reasoning chain $P'_q$ (as an explanation for the answer) along with the answer, $a$. The set of `facts' in the reasoning chain are often referred to as supporting facts \citep{DP1}.
% and $p_{qi}$'s need to be in the order in which they are required to conclude $a$. \par

The given definition is generic and can be extended to accommodate multiple variations of MHQA, some of which are listed below. Note that the given list is not exhaustive and the proposed definition of MHQA may be extended with new variations.
\begin{itemize}
    \item \textbf{MHQA via fact composition:} $C_i$ represents independent facts.
    \item \textbf{MHQA over long documents:} QA over long documents can be considered multi-hop if the question requires the model to aggregate information across different sections, passages or sentences of the same document \citep{DP5, SP22}. Here, each $C_i$ is a section/passage in the same document.
    \item \textbf{MHQA over multiple text documents:} Each $C_i$ is an independent document.
    \item \textbf{Multiple choice MHQA:} For each question, there is a small set of possible answers given beforehand. Thus, $\mathcal{A}$ in Eq. \ref{eq:def} is dependent on $q$, $\mathcal{A} = \mathcal{A}(q)$.
    \item \textbf{Open Domain vs Closed Domain MHQA:} In the open domain setting, the set of contexts relevant to the question, $C$ spans the entire corpus i.e., $C=\mathbb{C}$ whereas in the closed domain setting, $C$, the input to the model along with the question $q$, is a small subset of $\mathbb{C}$ and may be different for each question i.e., $C=C_q\subset\mathbb{C}$. The closed domain setting guarantees that $C_q$ is sufficient to answer $q$ and might also contain noisy irrelevant passages. It is important to note that the distinction between open-domain and closed-domain is regardless of the sizes of $C, \mathbb{C}$, but is determined by the input of the task - whether each question is provided with a specific subset of sufficient contexts or not. As we will see in Chapter \ref{sec:method}, the open-domain setting can be reduced to closed-domain by performing a preliminary retrieval over $\mathbb{C}$.
    \item \textbf{MHQA over Knowledge Bases/Knowledge Graphs:}\footnote{For the scope of this book, we do not consider QA over KB or KG since these have been extensively covered by \citet{complex-survey-1, complex-survey-2,DBLP:series/synthesis/2021Roy}.} $\mathbb{C}$ is a Knowledge Base (KB) or a Knowledge Graph (KG) with $C_i$ representing a triplet or a graph node. 
    \item \textbf{Visual MHQA:}\footnote{For the scope of this book, we only consider text based QA.} $\mathbb{C}$ is a set of images and/or videos (or equivalently, sequence of images). 
    \item \textbf{Conversational QA with clarifying questions:} Conversational QA can be regarded as an MHQA problem if answering the question requires the model to ask follow-up questions to the user. In this task, the user asks a question and the model tries to answer it based on some context. If the model cannot find the information necessary to answer the question, it generates a follow-up question for the user and enquires the missing information. Here, each follow-up question can be considered as a retrieval hop and figuring out the missing information can be regarded as a reasoning hop \citep{SP22}.
    \item \textbf{Temporal MHQA:} $C_i$ represents here diverse temporal contexts which could be defined in several different ways. These could be simply documents published over different time points. However, on a more general level, contexts could be in the form of relevant time periods. Such time periods could mark bursts in the temporal distribution of relevant documents returned for an input question, or they could be formed by the focus time \citep{focus-time} of relevant documents estimated either based on the embedded temporal expressions in the documents \citep{wangIRJ21,WangJ0Y20}, or inferred from past events and entities mentioned in text \citep{wangSIGIR21}.

\end{itemize}
We conclude this chapter by acknowledging some of the limitations of the proposed definition. Firstly, there may be multiple ways of mathematically denoting the same task. We do not claim the presented notation to be better than other possible notations in any way. We use the definition and notation that works best for us and allows us to clearly distinguish the scope of this book. We also prefer this notation as it allows us to naturally derive the commonly used concepts in MHQA such as reasoning chains, contexts and hops. It also allows us to capture the two-step process of retrieval and reasoning that is commonly adopted in question answering. We welcome the community to progress towards a more accurate and widely acceptable definition.
\chapter{A Comprehensive Study of Datasets: Analysis and Guidelines} \label{sec:data}

The previous chapter highlighted the different forms of the task which implies the existence of multiple datasets that are unique and equally significant, and are suited to different variants of the problem. In this chapter we aim to briefly summarize existing datasets and provide their comparison. This chapter details on the process of dataset creation, statistics, comparison among the existing datasets, and is followed by our critique of these datasets. We start the discussion with existing datasets as this will provide necessary context and intuition of the data for the following chapters devoted to methods which operate on  top of these datasets.
\par
We split our discussion on datasets into three broad sections. In the first section, we dive into details of how the datasets are created, followed by a section devoted to some statistics and comparisons among the existing datasets. We end the discussion with a section explaining some critiques and observed shortcomings of the current datasets. 
\section{Dataset Creation}\label{subsec:data_create}
We begin by describing how MHQA datasets are generated. In this section, we include major challenges for creating these datasets, followed by different steps involved in the process.
\subsection{\textbf{Challenges}}
Creating a dataset for multi-hop question answering is more challenging than for a traditional (i.e., single hop) QA setup. The major challenges include:
\begin{itemize}
    \item Contexts used for questions should form a valid and unambiguous reasoning chain. This implies that a context used for a particular question should have some information overlap or some kind of entailment relation with at least one of the other contexts present in the expected reasoning chain \citep{DP1}.
    \item Since the questions should require reasoning over multiple contexts, the process of question generation itself needs to somehow encapsulate information across those particular contexts.
    \item The creation process needs to ensure that the question is not answerable by using any single context. For instance, the question ``\textit{Who was the president of United States in the year in which World War II began?}'' requires two contexts containing ``\textit{World War II began in 1939.}'' and ``\textit{Franklin D. Roosevelt was the president of United States in 1939.}'' However, there may be a separate context containing ``\textit{Franklin D. Roosevelt, president of United States at the start of World War II, was unwilling to...}''. This challenge is more prevalent in the setting of Open Domain QA because of the incomplete prior knowledge about the possible contexts.
\end{itemize}  
Therefore, it becomes crucial to look at the methods adopted during creation of existing datasets, along with their advantages and drawbacks and possible ways to mitigate these. In general, the dataset creation task can involve three major steps, which we discuss next.
\begin{enumerate}
    \item Generating reasoning chains
    \item Question generation by crowd-sourcing
    \item Automatic/manual filtering
\end{enumerate}
\subsection{\textbf{Reasoning Chain Candidates}} Methods for coming up with reasoning chains are highly specific to the task and domain and thus, differ significantly for each dataset. Thus, we look at a few datasets individually.
\begin{enumerate}
    \item \textbf{HotpotQA \citep{DP1}:} HotpotQA contains only 2-hop questions formed using the first passages of documents from the English Wikipedia dump\footnote{\url{https://en.wikipedia.org/}}. Two passages are chosen as a reasoning chain (termed \textit{candidate passage pair}) if they satisfy either of the two conditions:
    \begin{itemize}
        \item There exists a hyperlink from the first document to the second. The entity which forms the hyperlink is termed as the \textit{bridge entity} and the questions are termed as \textit{bridge} questions.
        \item The entities for those passages belong to the same category (e.g. Michael Jordan and Kobe Bryant). These are specifically sampled from 42 manually created lists. Such pairs are used for creating \textit{comparison} questions.
    \end{itemize}

    \item \textbf{MultiHop-RAG \citep{LL23}:} MultiHop-RAG uses a set of diverse news articles and prompts GPT-4 to extract factual sentences from each article. These articles are again passed to GPT-4 for paraphrasing the factual sentence into a natural language claim. A topic and an entity are also generated in this step which are used as bridge entities for the reasoning chains. 

    \item \textbf{HybridQA \citep{DP6}:} In HybridQA, the definition of a context can be either a passage or a table. It is argued that using a table as context avoids the ambiguity of the questions that could arise when using texts. To ensure at least two hops, the questions are restricted to have reasoning chains containing at least one table and one text document. The tables are filtered from the set of tables released in WikiTables \citep{bhagavatula2013methods}, and the Wikipedia hyperlinks present in the cells of the table are used to retrieve relevant passages. At most 12 sentences of the first passage of each Wikipedia page are treated as the passages. 
    % Depending on the reasoning required, the questions are categorized as T$\rightarrow$P (table to passage), P$\rightarrow$T, P$\rightarrow$T$\rightarrow$P, T\&P (jointly using table and passage), T-compare (comparison using table), P-compare T-superlative and P-superlative.
    \item \textbf{NarrativeQA \citep{DP3}:} The dataset is for answering multi-hop questions on long stories. To ensure that the questions are indeed multi-hop and answering requires non localized reasoning, the annotators are asked to form questions using human-generated summaries of the stories. The stories are collected from books from Project Gutenberg\footnote{\url{https://www.gutenberg.org/}}, while movie scripts are scraped from the web.
    \item \textbf{OpenBookQA \citep{DP4}:} OpenBookQA requires question answering on the scientific domain using a \textit{book} (in simpler words, a collection) of scientific facts along-with a \textit{broad common knowledge} (large open-domain scientific sentences). The \textit{book} and \textit{common knowledge} are the two contexts required to answer the questions in OpenBookQA. \textit{Book} is collected by filtering a subset of the WorldTree corpus identified by \citet{worldtree} where the \textit{common knowledge} is collected from a collection of 14M scientific facts across Wikipedia, ConceptNet and other scientific corpora. 
    \item \textbf{QASC \citep{DP6}:} QASC is also a dataset for two-hop QA using scientific facts. The process of generating reasoning chains is very similar to that of OpenBookQA. The two contexts of the reasoning chain are chosen from a set of good quality seed facts and a large corpus of auxiliary facts, respectively.
    \item \textbf{MultiRC \citep{DP5}:} The purpose of this dataset is to create multi-domain multi-hop questions. Documents across various domains are selected from multiple corpora. Here, the multiple contexts are part of the same passage and the task of candidate reasoning chain generation is left to the annotators; each document is given to the annotators and they are asked to form valid multi-hop reasoning questions.
    \end{enumerate}

\subsection{\textbf{Generating Questions}} Generating multi-hop QA pairs requires accumulation of information across different contexts which itself is an unsolved problem \citep{SP14, AP1, DP7, QG1, QG2, QG3, QG4, QG5, QG6}. 

\paragraph{Automatic generation:} \citet{DP2} automatically generate questions using existing KBs where WikiData and DrugBank \citep{wishart2008drugbank} are used as the knowledge bases, and Wikipedia and {\sc Medline}\footnote{\url{https://www.nlm.nih.gov/medline/medline_overview.html}} are used as the document corpus. However, the question and answer types in the two datasets are highly constrained, owing to the creation technique. For instance, the only question type in the MedHop dataset is of the kind $(drug_1, interacts\_with, ?)$, where in the WikiHop dataset we have $(item_1, property, ?)$. Therefore, automatic generation using KBs is argued to result in datasets limited by the incompleteness of entity relations and schema of the KB used \citep{SP16, DP1}. Furthermore, automating the generation of free-form text questions is known to be a challenging task and requires substantial training data \citep{gatt2018survey, novikova-etal-2017-need, SP14}. However, with the latest research in LLMs, automatic question generation followed by semi-automatic verification is commonly adopted for creating small scale datasets \citep{LL23}.

\par
\paragraph{Crowd sourcing:} Consequently, traditional datasets propose that the question creation step should be done using human-intelligence. Since the task is not very straight-forward even for humans, existing works have added comprehensive guidelines for creating questions. These annotation guidelines follow a common pattern, with slight nuances, irrespective of the task. We aim to summarize the annotation guidelines used by various datasets and hope that this could serve as a reference for the relevant future endeavours. Based on the techniques adopted by the existing datasets, the common annotation instructions and best-practices include:
\begin{enumerate}
    \item To ensure that the questions \textbf{require multi-hop reasoning}:
    \begin{itemize}
        \item Give examples of both positive and negative instances of what the task requires.
        \item Break down the question generation task into simpler steps to prevent mistakes.
        \item Use rule-based techniques to provide in-the-loop friendly hints and prevent the annotators from submitting trivial incorrect samples.
        \item Use simple single-step IR or PLM (Pre-trained Language Models) based techniques to check if the question submitted is answerable using a single context only.
        \item Ask annotators to submit the reasoning chain along with the question and answer.
        \item Ask a different set of annotators to answer the questions using only a single context.
        \item Ask a different set of annotators if the question requires multiple contexts to be answered.
    \end{itemize}
    \item To ensure that the questions \textbf{are answerable}:
    \begin{itemize}
        \item Ask the annotator to also provide the answer to the question.
        \item Prohibit the use of negation words in answer choices that can trivially fool baselines.
        \item Ask for questions with very specific answers.
        \item Ask a different set of annotators to answer the question and discard questions with significant error rates.
    \end{itemize}
    \item To ensure that the questions and answers \textbf{are formed well}:
    \begin{itemize}
        \item Restrict the annotators to experts/native speakers of the intended language of the dataset.
        \item Ask a different set of annotators to verify if the questions are grammatically well-formed.
        \item Place a limit on the length of the answer.
        \item For multiple choice questions, randomly shuffling answer choices can avoid annotator biases, such as choice A being the correct answer more often than choice D.
    \end{itemize}
    \item To ensure that the questions \textbf{are challenging to answer}:
    \begin{itemize}
        \item Prohibit the annotator from copying spans of text from the input contexts.
        \item Ask the annotators to make the questions challenging.
        \item Ask the annotators to create confusing and irrelevant incorrect answer choices.
        \item Ask the annotators to consider high-level relations between entities and events rather than localized relations.
    \end{itemize}
\end{enumerate}

However, crowd-sourced datasets also have severe limitations. \citet{SP23} argue that these datasets usually only present a partial picture of the underlying data distribution and are marred by numerous biases, such as annotator bias \citep{geva-etal-2019-modeling}, label bias \citep{dua-etal-2020-benefits, gururangan-etal-2018-annotation}, survivorship bias \citep{AP4, SP32}, and ascertainment bias \citep{jia2017adversarial}. Furthermore, the absence of adversarial contexts during training time allows the models to learn shortcuts in reasoning and perform the task without doing multi-hop reasoning.
    
\paragraph{\textbf{Post-processing:}} The above-mentioned practices and instructions might not preclude all human errors. Thus, a further manual or rule-based automatic screening is generally done to get rid of these errors. For instance, \citet{DP6} remove a question 1) if the answer cannot be found from either table or passage, 2) if the answer is longer than 20 tokens, or 3) if the answer passage is easily retrieved using a single hop of TF-IDF retrieval. \citet{DP4, DP7} remove questions that the annotators are unable to answer correctly. \citet{DP5} remove a question if the annotators are able to answer it using only a single context.
\citet{DP2} aim to resolve the candidate frequency imbalance by sub-sampling the dataset to ensure that questions having any particular answer candidate constitute not more than 0.1\% of the dataset, and also by omitting articles that are about the United States. They tackle the issue of document-answer correlation by discarding the questions having any commonly occurring pair of a document and candidate answer.
Finally, \citet{LL23} use GPT-4 to evaluate examples in the dataset against the following criteria: 1) The generated query must utilize all provided evidence in formulating the response; 2) The query should be answerable solely based on the provided evidence; 3) The response to the generated query should be either a single word or a specific entity; 4) The query must conform to its designated query type.

{
\scriptsize
\begin{landscape}
\begin{table}
\centering
\caption{\textbf{Comparison of MHQA datasets.} $\mathbb{C}$ represents the set of all contexts present in the dataset whereas $C$ represents the set of contexts for a single question. In OD (Open Domain setting), $C=\mathbb{C}$. }
\label{tab:dataset}
% \begin{minipage}{15cm}
\scriptsize
\resizebox{1.5\textwidth}{!}
{%
\begin{tabular}{|c|c|c|c|c|c|c|c|c|c|}
\hline
 & \textbf{\begin{tabular}[c]{@{}c@{}}Context \\ granularity\end{tabular}} & $\mathbf{\vert \mathbb{C} \vert}$ & \textbf{Size} & \textbf{\#hops} & \textbf{| {C} |} & \textbf{Question Source} & \textbf{\begin{tabular}[c]{@{}c@{}}Context\\ source\end{tabular}} & \textbf{Domain} & \textbf{Ans. type} \\ \hline
\textbf{HotpotQA} & Passage & |Wikipedia| & 112,779 & 1/2/3 & 10 / OD$^a$ & Wikipedia & Wikipedia & Generic & Span \\ \hline
\multirow{2}{*}{\textbf{HybridQA}} & Table, & Tables: 13k & \multirow{2}{*}{69611} & \multirow{2}{*}{2/3} & 1 table  & Wikitables, & Wikitables, & \multirow{2}{*}{Generic} & \multirow{2}{*}{Span} \\
 & Passage & Passages: 293k &  &  & passages & Wikipedia & Wikipedia &  &  \\ \hline
\textbf{NarrativeQA} & Sentence & \begin{tabular}[c]{@{}c@{}}Books: 783\\ Movies: 789\end{tabular} & 46765 & - & 1 story & \begin{tabular}[c]{@{}c@{}}Multiple$^b$\end{tabular} & Multiple & Fiction & Generative \\ \hline
\textbf{MultiRC} & Sentence & 871 & 9872 & 2.37 & 1 passage & \begin{tabular}[c]{@{}c@{}}Multiple$^c$\end{tabular} & Multiple & Generic & \begin{tabular}[c]{@{}c@{}}MCQ\\ $\vert \mathcal{A} \vert$: 5.44\end{tabular} \\ \hline
\textbf{Medhop} & Passage & Medline & 2508 & - & OD & Drugbank & Medline & Medicine & \begin{tabular}[c]{@{}c@{}}MCQ\\ $\vert \mathcal{A} \vert$: 8.9\end{tabular} \\ \hline
\textbf{Wikihop} & Passage & |Wikipedia| & 51318 & - & OD & Wikidata & Wikipedia & Generic & \begin{tabular}[c]{@{}c@{}}MCQ\\ $\vert \mathcal{A} \vert$:19.8\end{tabular} \\ \hline
\textbf{QASC} & Sentence & \begin{tabular}[c]{@{}c@{}}Core: 928\\ Aux: 7672\end{tabular} & 9980 & 2 & OD & \begin{tabular}[c]{@{}c@{}}Core: WorldTree\\ Aux: \citep{facts}\end{tabular} & WorldTree & Science & \begin{tabular}[c]{@{}c@{}}MCQ\\ $\vert \mathcal{A} \vert$: 8\end{tabular} \\ \hline
\textbf{OpenBookQA} & Sentence & \begin{tabular}[c]{@{}c@{}}Core: 1326\\ Aux: 6000\end{tabular} & 5947 & 2 & OD & WorldTree & WorldTree & Science & \begin{tabular}[c]{@{}c@{}}MCQ\\ $\vert \mathcal{A} \vert$: 4\end{tabular} \\ \hline
\end{tabular}%
}
\newline
\begin{minipage}{12cm}
\vspace{0.5em}
$^a$\footnotesize{10 for distractor setting and OD for Full-wiki setting} \\
$^b$ \footnotesize{Summary: Wikipedia; Stories: Gutenberg; Movies: \url{http://www.imsdb.com/}, \url{http://www.dailyscript.com/}, \url{http://www.awesomefilm.com/}} \\
$^c$ \footnotesize{News: CNN, WSJ, NYT; Wikipedia; Articles on society, law and justice; Articles on history and anthropology; Elementary school science books; Stories: Gutenberg project, MCTest; Movies: CMU movie summary corpus}
\end{minipage}
\end{table}
\end{landscape}
}

\section{Existing Datasets: Statistics, Comparisons and Examples}
A comparison of the popular datasets used for MHQA is shown in Table \ref{tab:dataset}. Some of these datasets have multiple settings, and different statistics for different settings. We describe these settings along with some additional details for three such datasets below:
\begin{enumerate}
    \item \textbf{HotpotQA} test set has two settings i) \textit{Full-wiki setting} which is open domain with the set of contexts being the first passages of all Wikipedia pages, and ii) \textit{Distractor setting} which is closed domain with a set of 10 passages (2 gold + 8 distractors) provided with each question. The 8 distractors are collected by using a TF-IDF retriever with the question as the query. The dataset also evaluates an auxiliary task of predicting \textit{supporting facts}. These are the sentences in the gold passages which were used by the annotator to create the question. Since many of the \textit{comparison} questions in the dataset are of yes/no type, many models also use the answer type prediction as an auxiliary task. This involves a 3-way classification of the answer being 'yes'/'no'/the extracted span. 
    \\
    The train set of HotpotQA is split into easy/medium/hard. The easy subset predominantly consists of single-hop answerable questions (over 70\%) whereas distinction between medium and hard questions is determined by training multiple baselines and testing the answerability of the questions. Although the dev and test set contain only the hard questions, authors show that the easy questions are also useful while training the model.
    
    \item \textbf{QAngaroo} datasets \textbf{(MedHop and WikiHop)} contain a masked version along with the original unmasked dataset for avoiding the candidate frequency imbalance. For instance, in the MedHop dataset, some drugs (such as Aspirin) interact with more drugs than others (such as Isotretinoin), which can lead to candidate frequency imbalance. To mitigate this, any candidate expression is randomly replaced by a unique placeholder token (e.g. “Mumbai is the most populous city in <MASK7>”). It is argued that doing this removes the answer frequency cues and also removes statistical correlations between frequent answer strings and relevant contexts.
    
    \item \textbf{NarrativeQA} question-answer pairs are created using the summaries of movies or books whereas the model is asked to answer the question based on the original story (referred to as the \textit{story version}). Another somewhat less challenging task of answering questions by directly using the summaries, referred to as the \textit{summary version} is also provided.
\end{enumerate}

\section{Critiques and Challenges}
Even with carefully designed datasets it remains to be doubtful whether the datasets in fact require the model to perform multi-step reasoning to conclude the answer. \citet{AP4} show that the questions formed by compositions of two contexts (as used by many datasets) are not the same as they do not generalize well to multi-hop questions generated in typical use cases. Consequently, there have been multiple insightful methodologies proposed to test this aspect.  
\par
\citet{AP3} design two baseline models to predict the sentence containing the answer and constrain them to score each sentence from each context independently without looking at other sentences. Thus, the models should ideally perform poorly when trying to answer multi-hop questions. However, the performance of these baselines is overwhelmingly closer to the state-of-the-art than to a random classifier. This is observed more so for multiple choice MHQA datasets (WikiHop and a multi-choice modification of HotpotQA) than for a span-based MHQA dataset (HotpotQA). Furthermore, a no-context baseline that predicts the answer only by looking at answer candidates performs almost as good as the baselines defined above even when the number of candidates is increased significantly (refer to Figure 3 in \citet{AP3}). Thus, it is argued that multiple choice questions are easier to hack while testing and less helpful while training when compared to the span based questions.

\citet{AP4} train a single-paragraph BERT model \citep{bert} that achieves comparable performance to the state-of-the-art on the distractor setting, while it lags behind on the open-domain setting indicating that the open-domain setting is more challenging for a single-hop model and is worthy of future study. To further verify the hypothesis that the distractor setting is majorly single-hop solvable, human evaluation is performed. Humans achieve an accuracy of $87.37$ F1 when using all the ten input paragraphs and that of $82.06$ when one of the gold paragraphs is missing. In order to improve the quality of these distractors, an adversarial set of distractors is collected by training a single-hop BERT model and picking the top scoring incorrect paragraphs. Although, the performance of the single-hop model drops significantly, it is recovered after fine-tuning the model on these distractors.

Another possible way to improve the quality of distractors is to add a large number of such distractors. However, it is observed that even with 500 distractors, the single hop model performance is significant \citep{AP4}. For the open domain setting, the performance goes down significantly indicating the challenging nature of open-domain MHQA. \\
On manual analysis of single-hop solvable questions, it is found that $35\%$ of the questions in the HotpotQA dataset are solvable only by matching the entity type. These questions would require multi-hop reasoning in the open-domain setting. However, other $26\%$ of the questions are single-hop solvable even in the open-domain setting. These are questions whose answers can be derived by finding an entity that uniquely satisfies two properties (intersection-type questions) but a unique answer can also be found by using only one of these properties. Another $8\%$ of the questions are non-compositional single-hop and only the remaining $27\%$ require multi-hop reasoning.
\par
Similarly, \citet{SP14} split the HotpotQA dev set into single-hop solvable (3426) and single-hop non-solvable (3979) questions.
\par
\citet{SP3} make a similar conclusion by observing that 1184 (20\%) of the questions have the answer span mentioned in both of the supporting passages and the answer can be extracted by only considering one of them.
\par
\citet{SP5} conduct an experiment that compares a QA model that has access to all the supporting passages in HotpotQA, to another model that has access only to the answer (or the second hop) passage and observe that the model with full access only gives a marginal improvement from 49.43 EM to 50.96 EM. 
\par
\citet{SP28} find that more than 60\% of the dev questions in MultiRC have at least one adjacent relevant sentence pair making the multi-hop reasoning very easy.
\par

In summary, there seems to be enough evidence that the existing datasets have some flaws and a model should be able to exploit these flaws and show an impressive performance without being actually good at multi-hop reasoning. At the same time, these experiments point out the potential issues and challenges relating to MHQA datasets, and motivate development of higher quality datasets.  
\chapter{Existing Approaches for MHQA} \label{sec:method}

In this chapter, we will be taking a look at the existing approaches used for solving multi-hop question answering. Since there is a very large number of works we aim to cover, we try to categorize them on many levels into categories and sub-categories. We hope this leads to a more structured study culminating into a taxonomy proposed in the Chapter \ref{sec:tax}. For each level of categorization, we briefly describe the category/sub-category followed by a deep dive into some of the representative methods for the category and finally, contrast the sub-categories and list some known pros and cons for each. To avoid over technicality, the readers can skip the deep dive into particular methods.

\begin{figure}[ht]
     \centering
     \begin{tabular}{cc}
        \includegraphics[width=0.45\textwidth]{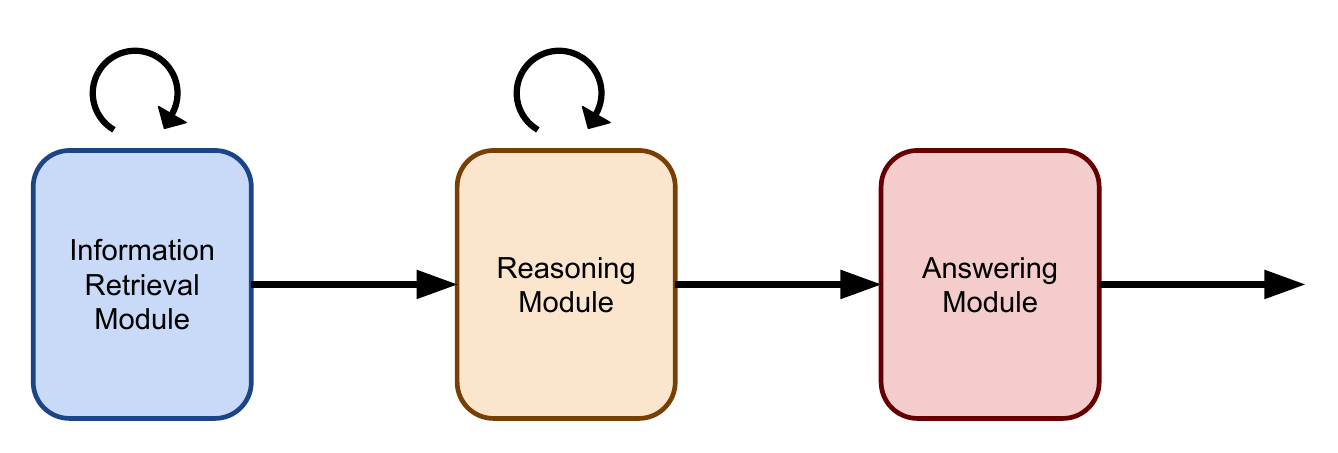} & \includegraphics[width=0.45\textwidth]{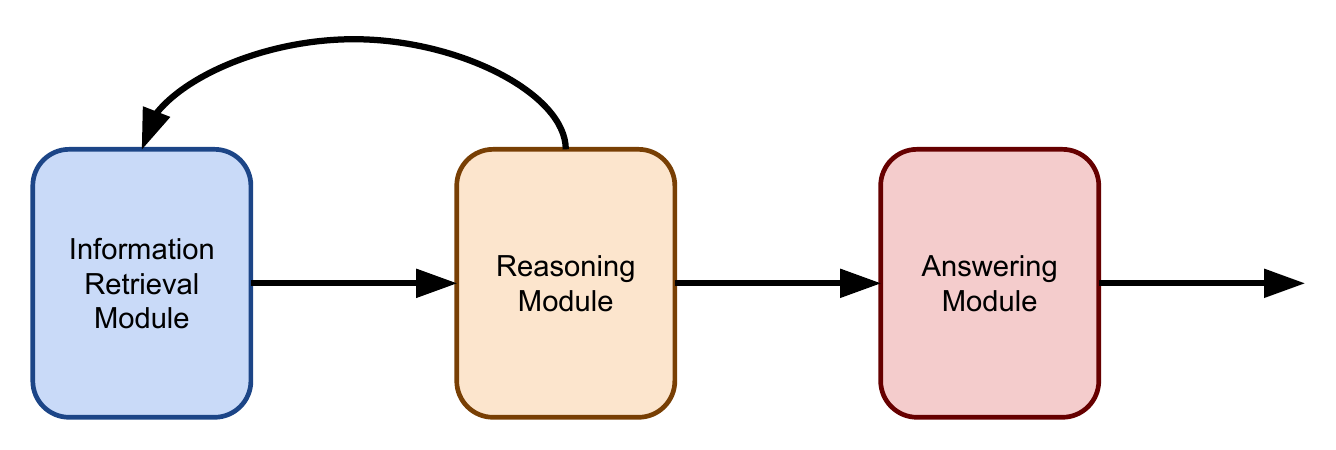} \\
        \small (a) Type-I: $Retr^{Itr}$->$Reas^{Itr}$->$Ans$
        & \small (b) Type-II: $(Retr$->$Reas)^{Itr}$->$Ans$
        \\
        
        \includegraphics[width=0.45\textwidth]{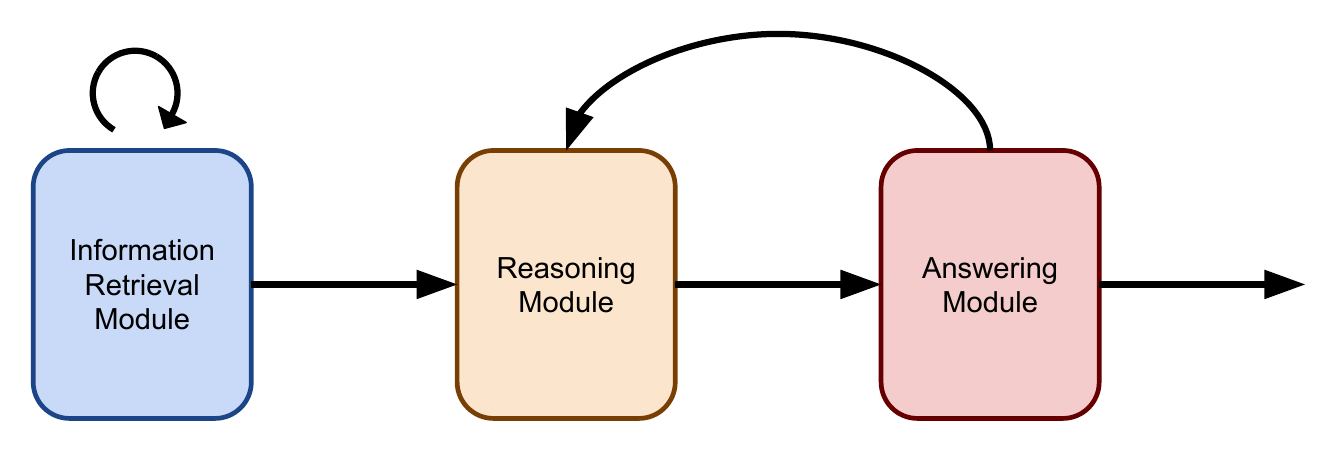} & \includegraphics[width=0.45\textwidth]{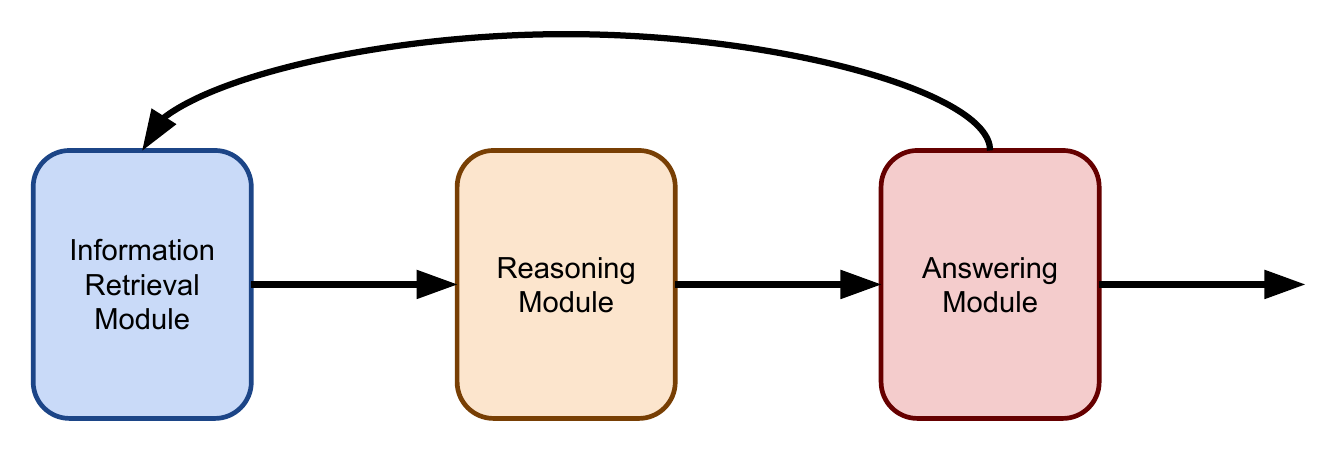} \\
        \small (c) 
       Type-III: $Retr^{Itr}$->$(Reas$->$Ans)^{Itr}$ 
        & \small (d) Type-IV: $(Retr$->$Reas$->$Ans)^{Itr}$
     \end{tabular}
     
     \caption{\textbf{The four types of architectures used for MHQA.} A self loop at a module indicates that that module is independent of the succeeding modules whereas an incoming connection from a succeeding module represents some kind of feedback from that block. a) 
     %Type-I: 
     The IR and reasoning modules perform multiple hops independent of each other as well as of the answering module. b) %Type-II: 
     At each hop, the IR module sends its output to the reasoning module which then gives feedback to the IR module. Answering module then predicts the answer in a single step. c) %Type-III: 
     IR module first iteratively retrieves all the relevant documents for all the hops. The reasoning module performs a hop and sends the output to the answering module. Answering module either answers the question or sends feedback to the reasoning module asking for another hop of reasoning. d) %Type-IV: 
     At each hop, the IR module sends its output to the reasoning module which further sends its output to the answering module. The answering module either predicts a final answer or provides feedback to the IR module indicating that some required information is missing.}
     \label{fig:model-types}
\end{figure}

As discussed in Chapter \ref{sec:prob_def}, a large number of works divide the task of MHQA into two steps, a retrieval (IR) step that extracts all the relevant contexts from the corpus and a reading comprehension (MRC) step that reads the resulting contexts to find the answer. In general, the existing works have three basic units - Retriever, Reasoner (or Reader) and Answer Predictor\footnote{Although initial works have a single module for reasoning as well as answer prediction, more recent works recommend further segregating these two as different modules. Therefore, we consider these as separate modules in our discussion.}. These units are often employed iteratively to be able to perform multiple hops. A coarse level classification of methods can be how these units interact with each other. One way is to complete all the iterations of one unit before moving to the next one, and the other is to perform two or three of the tasks in a single iteration and repeat for the corresponding hops. Thus, there are four possibilities that arise by considering the multi-step nature of the retrieval and reasoning modules. These four types of models are shown in Figure \ref{fig:model-types}. 
\par 

A few models retrieve important sentences or entities from the contexts as an intermediate step for reasoning. Since the granularity of the context is sentences for some datasets like MultiRC and OpenBookQA, it can be difficult to determine what constitutes reasoning and what constitutes retrieving. To avoid such confusion, we refer back to our definition of the task in Section \ref{sec:prob_def}. If the granularity of the output of a step is the same as the granularity of the context, we refer to that step as part of the retrieval process. If the granularity is lower, we take it as part of the reasoning.
We begin by describing the techniques used for these three units followed by some auxiliary tasks in practice.

\begin{figure}[ht!]
    \includegraphics[width=\textwidth]{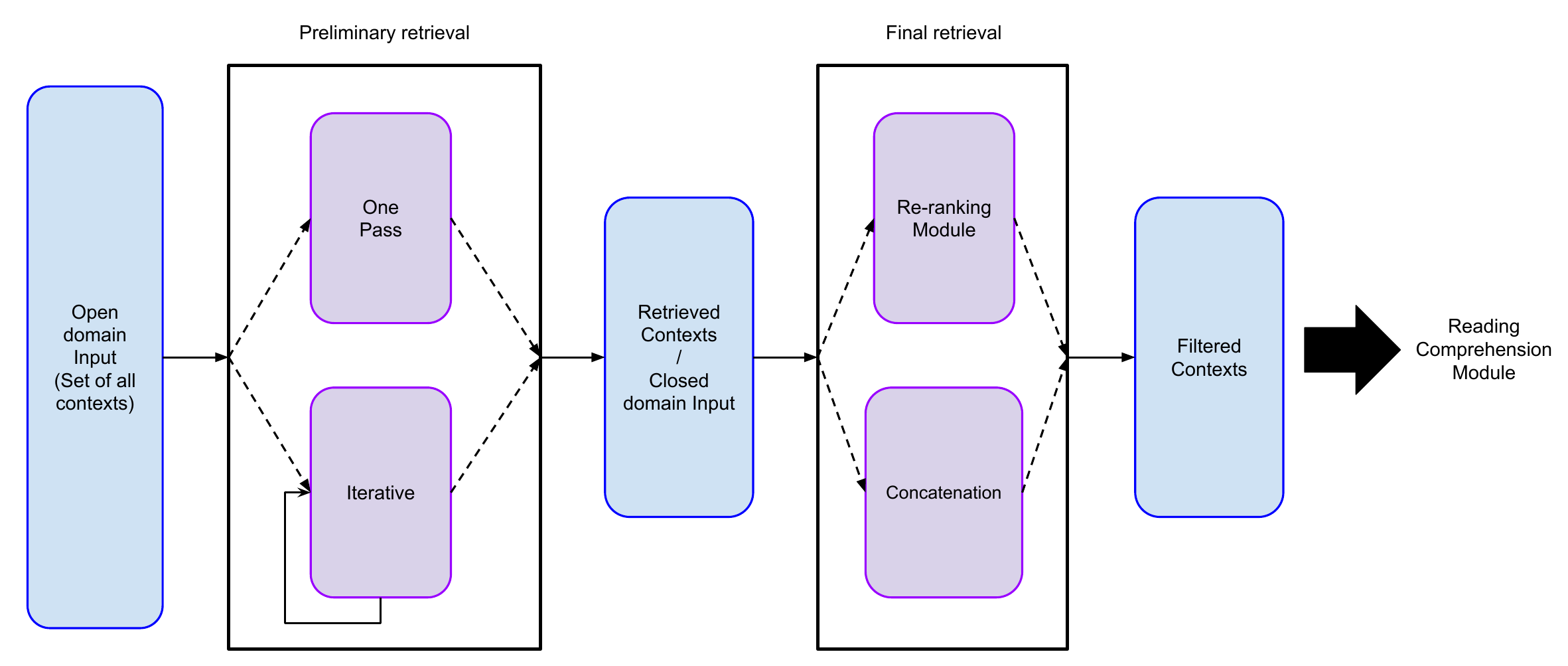}
    \centering
    \caption{\textbf{Retrieval Module.} The process of retrieval can be decomposed into two steps: a) Preliminary retrieval which is a quick and high recall step to retrieve all the relevant context from the set of all contexts in the open domain setting. It can be performed either as a single pass or iteratively. 
    %Preliminary retrieval can either be performed in a single pass or by iteratively querying the input. 
    b) Final retrieval which is a more thorough step to filter out all the irrelevant contexts. Because of a smaller number of input contexts, more complex re-ranking models can be used at this step. Some models directly pass the concatenation of the input documents to the reasoning module. Note that the closed domain setting of MHQA eliminates the need for preliminary retrieval.}
    \label{fig:retrieval}
\end{figure}

\section{Retrieval}
The retrieval step can be a bottleneck when the set of available contexts for a question is large and becomes a particularly challenging task in case of Open Domain QA. Intuitively, multi-hop retrieval is significantly more challenging than a single-step retrieval since any retrieval error in a hop is accumulated with each subsequent hop of retrieval and leads to a well known problem of semantic drift \citep{SP7}.
\citet{SP3} verify the hypothesis by running a simple BM25 on `easy' and `hard' subsets of HotpotQA (which contain predominantly single-hop and multi-hop questions respectively) to find that the accuracy drops from 53.7\% on the easy subset to 25.9\% on the hard subset. Similarly, \citet{SP2} run a TF-IDF retriever and find that although it succeeds in retrieving at least one of the gold passages among the top 32 passages for more than 90\% questions, it often fails to retrieve both the gold passages. 
% Some of the methods also utilise the output of the retrieval step to get the reasoning chain (or supporting facts) for reaching the answer.
\par
The existing methods for retrieval can be broadly classified into two categories as shown in Figure \ref{fig:retrieval}: open domain retrieval where the model has to search through the entire context set $\mathcal{C}$, and closed domain retrieval where the model is provided with a smaller set of relevant (and noisy) contexts $C \subset \mathcal{C}$ along with the question $q$. Many of the techniques working in the open domain setting use a two step strategy: a) the first step (referred in this text as \textit{preliminary retrieval}) is a fast and high recall filtering to get an initial set of relevant contexts. b) the second step (referred in this text as \textit{final retrieval}) is then reduced to the closed domain setting where more complex techniques can be used to further remove the noisy contexts. This two step approach combines the best of both worlds, getting the good time efficiency of coarse retrievers and the good performance of fine-grained retrievers.
% Many of the existing works use a reranking module for the second step.
\subsection{\textbf{Preliminary Retrieval}}
The first step of the open domain retrieval (referred to preliminary retrieval from hereon) can be a single shot retrieval as done in single hop question answering i.e., the model attempts to extract all the relevant contexts in a single retrieval step. We call this strategy \textit{single-pass}. Another solution would be to retrieve the passages iteratively, where each iteration of retrieval can correspond to each hop in the reasoning. We describe in detail some of these approaches below.
\paragraph{Single-pass approach}
As mentioned above, the single-pass approach performs the retrieval in one step. We list some of the representative methods of this approach below.
\begin{itemize}
    \item \citet{SP24} and \citet{DP6} follow a simple setting of the single-pass retrieval approach, calling the retrieval only once for each query. 
    \item For the multi-choice questions setting, \citet{SP4} append each candidate answer with the question to get multiple queries and BM25 is used on these queries to extract top $n(=20)$ contexts (sentences in case of the MultiRC dataset).
    \item For the text and table hybrid setting, \citet{DP6} parse the passages linked to each cell in the given table and retrieve all the cells relevant to the question in one go.
\end{itemize}

\textbf{Critique:} This strategy works well for a few-hop case where the number of contexts to be retrieved is limited and the candidate answers are provided for the multi-choice scenario. However, for more than two hops, the retrieval or the ranking algorithm could fail to capture the terms relevant to the intermediate hops. For instance, in a three-hop question, the information in the second hop could be unrelated to both the question and the candidate answers. Furthermore, if the candidate answers are not provided, this strategy could also fail to capture the second hop well \citep{SP1}. To validate this hypothesis, \citet{SP1, SP2} use a simple BM25 with the original question as the query to evaluate the second hop paragraph recall on HotpotQA and find the results to be indeed underwhelming.

\paragraph{Iterative approach}

Many of the existing techniques, therefore, propose a multi-step retrieval. Two main classes of iterative retrieval are query reformulation and entity-linking, described below.
\par
\textit{Query reformulation:}
 Deriving a query for the current hop ($q_t$) from the previous hop query ($q_{t-1}$) is commonly known as query reformulation (sometimes also referred to as pseudo-relevance feedback). Based on the strategy used for query reformulation, these methods can be further classified into two subcategories: text space reformulation and hidden space reformulation.
 \par
 \textit{Hidden space query reformulation: } In this case, the query is reformulated in the embedding space. 
     As a representative example for hidden space query reformulation, we take a detailed look at \citet{SP2}. They first encode the question using a Bi-GRU layer on contextualized ELMo \citep{elmo} embeddings and retrieve all the relevant passages using MIPS. To reduce the search space for MIPS, a TF-IDF based retriever \citep{chen-etal-2017-reading} is employed to get top $n_i$ passages. A supervised re-ranker scores the paragraphs and each of the top k paragraphs is used to modify the question hidden representation to give k new search vectors for the next step. The reformulation module uses a bi-directional attention \citep{bidaf} on the paragraph and question encodings followed by a linear layer with ReLU activation. A residual connection is added with this output being passed to a Bi-GRU layer followed by another linear layer with ReLU activation. Max pooling is applied to the residual output to give the updated query vector. At every step, top k paragraphs are selected by the re-ranker. The paragraph encoding being independent of the question allows the encodings to be pre-computed for an efficient retrieval during training and inference.
\par
\textit{Text space query reformulation: } Some methods reformulate the question by adding to, modifying or re-weighting the text of the question. Changing the questions in the text space allows the intermediate questions to be interpretable. We dive a little deeper into some of the methods following this approach:
\begin{itemize}
    \item \citet{SP7} concatenate the question with each answer candidate to obtain initial queries. Justification sentences are retrieved using the unsupervised alignment method proposed by \citet{yadav-etal-2019-alignment} which uses GloVe embeddings \citep{glove}. 
    {They compute a matrix storing the cosine similarity of embeddings of the query tokens with the sentence tokens. Max pooling across sentence tokens is applied to get the most similar tokens for each query token. Dot product between this vector and a vector containing the IDF values of the query tokens is calculated to produce the overall alignment score.}
    For MultiRC, these are selected from the sentences in all the relevant paragraphs. For QASC, relevant sentences are retrieved using heuristic IR approaches. The reformulation process only keeps the uncovered tokens and if the number of such tokens is $<T(=2-4)$, new tokens are added from the previously retrieved sentences. The process is repeated until either a) no new query tokens are retrieved or b) all tokens are discovered. 10.7\% improvement is observed when the tokens are identified using soft matching over GloVe embeddings. 
    \item \citet{SP8} use TF-IDF with the question to get the first hop paragraphs. For each subsequent hop, the ALBERT based reader module is used as a span extractor on the previously retrieved documents to extract the text relevant to the question. The extracted span is concatenated with the query of the previous step for performing the next hop of retrieval. This is repeated until the reader module finds an answer or a maximum number of hops is reached.
    \item \citet{SP24} follow a similar approach to \citet{SP8} with DrQA's Document Reader model used as the span extractor. During training, heuristics are used to find the oracle query.
    \item \citet{SP9} retrieve k justification sentences using an alignment technique similar to \citet{SP7}. The question $Q$, is concatenated with each retrieved justification $q_k$, and the token weights are assigned as: For each token $t$ in original question, if $q_k$ contains $t$, weight for $t$ is 1, else it is 2. All terms in $q_k$ have a weight of 1. This is expected to result a in higher coverage. The queries are used for second hop retrieval to get a final set of $N$ contexts. Then, $\binom{N}{p}$ evidence chains are generated and each evidence set is ranked by how many query terms are included (coverage) and top $n$ are picked for the supervised reranking.
    \item \citet{SP17} propose generating text based free-form follow-up questions for performing iterative retrieval. An IR model (BM25) retrieves the set of relevant contexts using the question. A BERT based three-way controller module predicts whether each context contains the final answer, contains some intermediate information or is irrelevant. A QG model is used for generating a follow up question for each passage that contains intermediate information. BERT is used for getting the answers from contexts containing the final answer. Irrelevant passages are ignored. The QG model by \citet{zhao-etal-2018-paragraph} is trained on the reverse SQuAD. The controller model is trained with cross-entropy loss for ternary classification on the ground truth triplets.
\end{itemize}
\textbf{Critique:} \citet{SP3} argue that query reformulation methods do not necessarily use the information about entities present in the evidence as they might not be the most frequent/ salient terms in it. Therefore, many works have proposed using entity mentions in the retrieved contexts for performing the second hop retrieval.
\par
\textit{Using entity links/hyperlinks}: In this approach, entity mentions in the first hop retrieval are used to find Wikipedia passages with titles containing any of these entities. This is an efficient technique since the entities can be extracted as part of the pre-processing step. It can be noted that this technique is particularly effective for HotpotQA since the creation process of the HotpotQA uses the Wikipedia paragraph for a bridge entity as the second hop context. However, this also indicates a bias of dataset design when building models \citep{SP6}. Trying to imitate (or benefit from the knowledge of) the creation process of some dataset while building the models for this dataset can improve accuracy on that dataset but is not guaranteed to perform well on the other datasets. On similar lines, \citet{SP3} even suggest not to use off-the-shelf entity linker as they are usually trained on Wikipedia and can lead to data leakage. 
\par
Below, we list some of the representative methods for this class of methods.
\begin{itemize}
    \item \citet{SP3} use BM25 for the first hop retrieval and then retrieve Wikipedia paragraphs for each mentioned entity. Since HotpotQA is known to contain both single-hop and multi-hop questions, a self-link is added for each entity retrieved in the first hop to deal with single hop questions.
    \item \citet{SP5} use a Hybrid TF-IDF + BM25 for the first hop retrieval to get 10 documents. Span prediction and external entity linking is used to get the bridge entities from the first hop passages. A supervised re-ranker gives top 10 entities and Wikipedia passages for these entities are used as the second hop passages. A span loss is added for predicting answer passage title entities as an auxiliary task.
    \item \citet{SP1, SP6} retrieve Wikipedia passages whose titles contain an entity mentioned in the question as the first hop passages. Hyperlinks from these passages are used to get the second hop passages.
    \item \citet{SP10} extract key words and match them with the paragraphs title and select top $N_1$ paragraph with best TF-IDF scores. Apart from these, top $N_2$ paragraphs with best TF-IDF scores are added. For the second hop, add all the paragraphs having hyperlinks from and hyperlinks to the first hop paragraphs.
\end{itemize}
% \\
%\textit{Query reformulation} (aka pseudo relevance feedback (PRF)) \\

\textbf{Critique:} \citet{SP11} evaluate the performance of the two sub-categories of iterative retrieval approaches by evaluating the performance on each of the two hops on HotpotQA dataset. It is argued that entity based retrieval is limited by the availability of such hyperlinks and passages whereas the reformulation approach is limited by the performance of lexical term-based retrieval. They observe that while BM25 followed by BERT based reranking performs well on the first hop, it fails on the second hop retrieval (evaluated as the ability to retrieve second passage given the first passage). Similarly, although the model by \citet{SP51} has the best overall performance, it has some room for improvement when evaluated for the second hop retrieval. Hence, a hybrid technique is proposed where the re-rank model is used for the first hop and a single-hop dense passage retrieval model \citep{SP30} is used for the second hop. Results show that the hybrid technique outperforms the existing techniques which calls for further study in this direction.

\subsection{\textbf{Final Retrieval}} \par
After getting an initial pool of retrieved contexts, many works suggest using a more fine-grained and more sophisticated retrieval on the retrieved contexts to get a better quality of contexts. Based on how the initial pool of retrieval output is processed, we can categorize the approaches into concatenation, supervised re-ranking and unsupervised re-ranking.

\textbf{Concatenating/union} Some of the methods do not filter the input contexts and directly pass a union or concatenation of the contexts to the reasoning model. For example:
\begin{itemize}
    \item \citet{SP7} maintain N-parallel reasoning chains in the preliminary retrieval step and a union of these chains is passed to the answer prediction module.
    \item \citet{SP1, SP4} do not filter the passages and directly perform reasoning at a lower granularity.
\end{itemize}

\textbf{Supervised re-ranking} Majority of the methods use a supervised module to score and rank the input contexts, and then use some criterion to filter the less relevant documents before passing to the answer prediction module: 
\begin{itemize}
    \item \citet{SP2} use a linear layer with sigmoid activation to get a relevance score of each sentence in a paragraph with the question. Max pooling across all sentences is applied to give a relevance score and top-k most relevant paragraphs are picked. Max pooling allows only one of the sentences in a paragraph to be relevant to the question with a high score.
    \item \citet{SP3} use BERT to compute query-aware embeddings for every pair of the first hop and second hop paragraph. The two paragraphs are concatenated and fed to 2-layer Neural Network to get a score. Top $k$ pairs are passed to the reader module.
    \item \citet{SP5} use a bi-LSTM layer \citep{lstm} to predict relevance of each second hop passage with the question.
    \item \citet{SP6} use a RoBERTa encoder \citep{roberta} followed by a fine tuning layer to get the top N paragraphs.
    \item \citet{SP8} pass the node (document) representations to a binary classifier for predicting their relevance and keep the $k$ highest scoring documents.
    \item \citet{SP9} use RoBERTa \citep{roberta} for reranking trained to predict the F-1 score of evidence chain.
    \item \citet{SP16, SP20} concatenate paragraphs with the question and feed it to a BERT followed by a binary classifier and keep N(=3) paragraphs with the highest scores. \citet{SP25} follows a similar approach but set a threshold to retrieve a variable number of paragraphs.
    \item \citet{SP10} use the gated memory flow network inspired by the Neural Turing Machine \citep{graves2014neural} to model the probability of a paragraph being the next context in the reasoning chain, conditioned on the question and the previous paragraphs in the reasoning chain. At every time step $t$ BERT is used to compute the question aware embeddings of the paragraphs, $x_t$. A KVMemNN architecture models the memory as a set of key-value pairs. 
    The model passes the key vectors and $x_t$ to linear layers and applies Softmax to the output of $W_xx_t\cdot W_k k_i$. The output after Softmax multiplied with another matrix $W_v$ is then used as weights while summing the value vectors $v_i$ to give the readout vector $o_t$. 
    The memory reading process is similar to computing self-attention \citep{transformer} and the authors concatenate the outputs after computing $o_t$ for the $h$ attention heads. $o_t$ and $x_t$ are passed to another linear layer with tanh and sigmoid activation to give the relevance score $s_t$. $x_t$ is written to the memory if $s_t>gate$ where $gate$ is a hyper-parameter. 
    Hard negative examples are generated by training a BERT based model to predict the relevance score $s_t$ and choosing the top-8 of the non-evidence paragraphs.
    \item \citet{SP23} propose generative context selection that learns to predict how the question would have been formed. Mathematically, the model tries to learn $p(a, q\vert C)$ instead of $p(a\vert q, C)$ (as in discriminative models). This probability is modeled as
    \begin{equation}
        p(a, q\vert C) = \sum_{c_{ij}} p(a\vert q, c_{ij})\cdot p(q\vert c_{ij})\cdot p(c_{ij}\vert C)
    \end{equation}
    where $p(c_{ij}\vert C)$ is the \textit{prior} that computes compatibility between any two contexts, $p(q\vert c_{ij})$ is the question generation model that predicts the probability of $q$ being formed from the given contexts, and $p(a\vert q, c_{ij})$ is the standard answering model. During inference, the model retrieves the contexts as $c^\ast _{ij} = argmax_{c_{ij}} p(q\vert c_{ij})\cdot p(c_{ij} \vert C).$ To model these probabilities, a pre-trained T5 is used for obtaining the contextual embeddings. The prior and generative models are trained together. Concatenation of every pair of contexts is passed to an encoder and a Pointer Generator Network (PGN) \citep{pgn} decoder is used to predict the question. The training objective is to increase the likelihood of the question for gold context pairs and the unlikelihood \citep{welleck2019neural} for a sample set of negative context pairs. T5 is used for answering using the best pair of contexts. The advantage of using a generative model is that it avoids the annotator biases in the dataset. This hypothesis is tested by running the model on the adversarial set of questions formed by \citet{SP32}. A multi-label sentence classifier $p(s|q, C)$ that selects relevant sentences is found to have a better performance but at the same time, is more biased.
    \item \citet{DP6} feed each cell along with its neighboring cells to the cell encoder to obtain their representations. The representations are aggregated and further fed to a feed-forward neural network to obtain a score.
    \item \citet{SP22} perform the retrieval in two steps where the first step retrieves a paragraph, and the second step retrieves a sentence from these paragraphs. A weighted sum of paragraph and sentence retrieval scores is computed to find the best retrieved sentence. The retrieval score of the paragraph retrieved from the first hop is broadcast to the sentences in the paragraphs. The best retrieved sentence fed to BERT-large for extractive QA.

\end{itemize}
\textbf{Unsupervised re-ranking} 
Some of the methods for supervised re-ranking include:
\begin{itemize}
    \item \citet{SP4} consider all $\binom{n}{k} (k\in[2, 5])$ reasoning chains formed by the $n$ retrieved passages and use a simple formula to compute the scores. They define \textbf{R}elevance as the average BM25 scores of the chain, \textbf{O}verlap as the word overlap between each sentence pair in the chain and \textbf{C}overage as the product of word overlap of sentence with the question and with the answer. The final score is given by $\frac{R\cdot C}{O}$. By experiments, $k=3, 4$ are found to be the best values with the justification that getting two correct retrieved passages in a chain of size two is tough while a chain of size 5 will suffer from noise.
    \item \citet{DP6} retrieve cells from the table in the HybridQA dataset in an unsupervised manner. A cell is selected if its value is either mentioned in the question, or is min/max of the corresponding column. A cell is also added if it hyperlinks to one of the passages that are retrieved by a TF-IDF retriever.
\end{itemize}
\textbf{Critique:} Supervised re-ranking can lead to a model that is more suited to the task at hand but requires some training signals which may not be available for certain tasks. On the other hand, unsupervised methods are more flexible but might not be optimized for a given task. In particular, \citet{SP4, SP7} experimentally verify that using a supervised re-ranking method has an implied shortcoming of resulting in a domain dependent performance. The performance for each domain depends on the portion of training data belonging to that domain. Thus, it is attempted to propose\textbf{ unsupervised re-ranking} techniques that can lead to similar level of quality as supervised re-ranking.

\par
Another parallel classification of the re-ranking approaches is based on whether the module scores each retrieved context independently or together as constituents of a possible reasoning chain. We call the two categories of methods \textbf{Context re-ranking} and \textbf{Chain re-ranking} respectively. Table \ref{tab:taxonomy} lists the methods belonging to the two categories. \citet{SP4} show that evaluating the reasoning chain as a whole leads to better performance than evaluating each passage independently.

\section{Reading Comprehension}
A reading comprehension module is responsible for reading the final set of retrieved contexts, combining the information across contexts and performing the reasoning steps or hops. The output of this module can either be an answer or some representation of the reasoning performed (for example, a reasoning chain, a semantic graph or some latent representation). A vast majority of approaches have used graphs or question decomposition to perform the reasoning, therefore we classify the reading comprehension techniques into graph-based, question decomposition based and miscellaneous categories.
\subsection{Graph-based Techniques}
The general flow of the graph based methods can be summarized in a three-step process: 1) A graph of one or more types of nodes (entity, sentence, paragraph, document nodes, etc.) is constructed and edges are added based on some lexical heuristics. 2) Contextual encodings of the nodes are passed to one or more layers of a Graph Neural Network \citep{gnn} (or a Graph Attention network \citep{gat} or a Graph Convolutional Network \citep{gcn}). These layers update the representation of each node using the representation of each of its neighbouring nodes. In this way, after $n$ such layers, nodes separated by a path of $\leq n$ edges have shared information with each other. Therefore, each layer is expected to perform one hop of the reasoning process.
3) The updated embeddings are passed to the answering module.
\par As can be expected, some methods slightly deviate from the proposed general flow. We describe these differences below along with some details on graph building and reasoning steps.
\begin{itemize}
    \item \citet{SP1} build an entity graph by starting with the entities extracted during the retrieval process. For each node, a BERT model takes its representation along with the question and a retrieved passage and extracts entities from the paragraph to be added to the graph. The process is repeated until either a) no new nodes can be added or b) a maximum number of nodes are extracted. Sentences containing the extracted entities are considered as `clues' for the respective entities. Clues are then used for updating the node representations using a Graph Neural Network (GNN). Clues are also used later for the auxiliary task of supporting facts prediction. 
    \item \citet{SP6}  propose constructing a hierarchical graph network with four types of nodes that represent question text, relevant paragraphs, sentences in these paragraphs and entities in these sentences. Bidirectional edges are introduced between the first hop source sentences and second hop target paragraphs. Edges are also introduced between paragraph and its sentences; sentence and the entities it contains; question and the entities it contains; between all paragraphs; between each sentence and its previous and next sentences. RoBERTa is used for encoding and a bi-attention and a Bi-LSTM layer are used for getting initial contextualized representations of the nodes. Graph Attention Network (GAT) \citep{gat} is applied over the hierarchical graph and the updated representations are merged with the original contextual representations using a gated attention mechanism. The merged representations are passed to the answer prediction module. The proposed gated merging is supposed to be effective for dealing with smaller number of hops.
    \item \citet{LL14} build a use abstract meaning representations (AMR) \citep{amr} to build a semantic graph from a passage. Graphs across passages are connected by adding edges between the same concepts in different graphs. A non-parametrized method is employed for generating reasoning chains. A depth first search (DFS) is used to find all paths from the question nodes to the answer nodes and all the edges on the path are used to recover the facts from the context that form the reasoning chain. The reader reads these reasoning chains and the question to predict the answer. A reinforcement learning based chain-aware loss is proposed to boost the performance of the system.
    \item Similar to \citet{LL14} ll13 and \citet{LL19} build a semantic graph. However, the entities and relations are extracted using GPT-3.5.
    \item \citet{SP8} construct a graph of documents and add edges if the documents have a shared entity. Question aware embeddings of the document are generated by Albert and used as initial node representations. GAT is used to update the representations of each shared entity. To update the representations of non-entity tokens, they are fed to a transformer along with the updated representations for multi-document fusion. The updated representations are passed to the answer prediction module.
    \item \citet{SP12} make entity graph of all entity mentions of the answer candidate in supporting documents. Cross-document edges are added between the mentions about the same entity and within-document edges are added between every node pair belonging to the same document. For initial representations, GloVe and ELMo are used for token-level and contextualized embeddings respectively. Contextualized embeddings are crucial since graph nodes only contain entity tokens. To deal with entities containing multiple words, average is taken over their embeddings. The two embeddings are passed to a 1 layer NN (replaced by Bi-LSTM for encoding the question) for getting the initial node representations which are concatenated with the NER and POS embeddings before feeding to a GCN layer. 
    \item \citet{SP13} compute similarity between GloVe embeddings of each word of the sentence with each word of the query and take the mean among m(=5) closest sentence words to get a sentence score and select top k(=25 or 30) sentences. A graph containing sentence nodes and document nodes is created with only the selected sentences. Edges are added between documents if they have a shared entity and between a paragraph and all its sentences. Adding edges among sentences adds complexity without much improvement. Initial representation of a node is computed by passing the matrix of GloVe embeddings of the entities contained in the sentence or paragraph to a bi-linear attention \citep{kim2018bilinear} layer on nodes and query. To compress the representations into fixed size, self-attention is used. T(=3) layers of Gated Graph Neural Network is used to update the representations.
    \item \citet{SP15} create an entity graph similar to \citet{SP12} but have some additional edge types: co-reference edges: between co-reference mentions of the same entity (separate type since these are less reliable) and complement edges which signify no connection between the two nodes. Co-reference edges is a separate edge type since these are less reliable owing to the error in the co-reference system. These are not required for the masked version.
    The query representations are formed by passing ELMo \citep{elmo} to a Bidirectional RNN \citep{rumelhart1985learning}. Initial query dependent node representations are formed by passing the ELMo contextual embeddings to a feed forward network. L layers of a gated relational-GCN \citep{schlichtkrull2017modeling} R-GCN, a version of GCN are applied to get the final representations which are passed to the prediction module. Relational-GCN is able to accommodate different edge types by using different weight matrices for the neighbouring nodes connected by different edge types. An ablation experiment verifies that, although useful, co-reference edges have the least contribution to performance. Another experiment tries to predict the edge type by training a model but the performance is poor.
    \item \citet{DP6} solve the text-table hybrid MHQA by using the retrieved cells from the previous stage to then decide which neighboring cell to hop to.
    \item \citet{SP25} construct an entity graph and add edges between entities with the same mention text, between entities belonging to the same sentence, and between each entity in a paragraph and each entity in its title. Question and contexts are concatenated and passed to a BERT model followed by a bi-attention layer to get the contextual node representations. The following four steps are performed iteratively: 1) Mean-max pooling over tokens in an entity mention to update the entity embeddings. 2) Attention between query and entity is used to compute the mask weights which are multiplied to the entity representations before feeding to a GAT. 3) Updated entity representations are concatenated to entity tokens and representations of all tokens are updated using an LSTM layer. 4) Bi-directional attention between query and entity representations is used to update the query representation.
    Updating embeddings of each token is necessary since the answer might not be an entity.
    \item \citet{SP16} create two different graphs: a) sentence graph with edges between sentences belonging to the same paragraph and between sentences that share an entity. b) entity graph with edges between entities appearing in the same sentence, between different mentions of same entities and between each entity in the paragraph to each entity in its title. The intuition is that humans first focus on question-related paragraphs, then sentences, and then the important words. BERT word embeddings are passed to a bidirectional attention layer to get the contextual embedding. They also propose a novel similarity matrix for the layer. Self attention is applied to the query and query aware node embeddings are passed to a GAT. Self attention is applied on the sentence node representations and the output is appended by the representation of each word in the sentence. An LSTM is applied to fuse the output of 2 graphs.
    \item \citet{SP20} form a sentence graph and add an edge between sentences $s_i$ and $s_j$ with the weight $w_{ij}$ given by: 
    \begin{equation}
        w_{ij} = \begin{cases}
            \frac{1}{1+e^{-n+K_1}} & \text{$s_i$ and $s_j$ have $n>0$ shared entities} \\
            \frac{1}{1+e^{d+K_2}} & \text{$s_i$ and $s_j$ belong to the same paragraph} \\
                & \text{separated by $d$ sentences.}
        \end{cases}
    \end{equation}
    These weights allow the model to deal with the edge types in a novel way. The concatenated paragraphs along with the question are fed to a BERT followed by a bi-attention layer. Sentence representations are obtained by extracting token level embeddings from the paragraph encoding and a weighted addition of token embeddings where the weights are calculated using a two layer MLP (Multi Layer Perceptron).
    Most methods using GNN perform message passing for each node in parallel. This requires the representation of nodes to be updated exactly $L$ times where $L$ needs to be specified as a hyper-parameter. If $L$ is large, it leads to over smoothing. If it is too small, it inhibits long-path reasoning. Moreover, this algorithm performs unnecessary updates leading to inefficiency. Thus, a novel message passing algorithm is proposed which performs a BFS starting from the question and passing the messages through every edge that is visited in the process.
\end{itemize}
\textbf{Critique:} \citet{SP13} argue that the advantage of using graph-structured representations lies in reducing the inference steps necessary to combine multiple pieces of information related in a path-like manner.
The fact that the required graphs (or some parts of it) can be created offline gives it a computational advantage. However, graph structure also has certain limitation \citep{SP10}, particularly for comparison type questions in HotpotQA where the evidence paragraphs about the two entities are independent. Many of the techniques assume that the relation between nodes is directional which may not always be the case. For gated GNNs, computational efficiency as well as the learning ability of the model degrades with the increasing number of nodes and edge types in the graph used.
\par
While the graph structure is prevalent for multi-hop reasoning, \citet{SP26} argue that it is not necessary and that graph attention can be considered as a special case of self-attention. Treating \citet{SP25} as the baseline, results are compared after removing the graph fusion module. It is observed that the performance gained by using the graph structure can be easily compensated for by fine-tuning the BERT based encoder. While the graph structure enables the model to focus only on adjacent nodes, a model without any prior knowledge can still learn this behaviour. This is verified by further experiments. When the graph fusion module is replaced by self-attention layers the results are very similar whereas replacing it by a transformer leads to a significant improvement. Graph attention reduces to self-attention for a fully connected graph making it a special case of self-attention. 
Attention patterns are visualized and observed similar to \citet{kovaleva-etal-2019-revealing}.
% for each attention head in the pre-trained model, we sum the absolute attention weights among those tokens belong to an entity and tokens not belong to an entity. The score of an attention head is the difference between the sum of weights from entities and non-entities tokens. We then average the derived scores over all the examples. Finally, the attention head with the maximum score is the desired head that contains entity-centered attention patterns.
It is found that the pre-trained transformers are able to capture several types of attention patterns: a) between entities, b) between co-referenced entities, c) between an entity and its attribute, and d) between an entity and a sentence. Depending on the structure of the graph, these patterns may or may not be covered by graph attention. Therefore, self-attention can be said to be more general and flexible than graph attention.

\subsection{Question Decomposition Techniques}
The key idea here is to decompose the multi-hop question into multiple single hop questions and use a single-hop QA model to answer each of the questions. This approach is particularly effective for questions like the bridge questions in HotpotQA that are constructed using a bridge entity. These questions can be easily decomposed into two sub-questions by finding the bridge entity. \citet{LL2} manually label the decomposition for questions across multiple datasets and use GPT-3 and RoBERTa to answer the questions and observe that question decomposition can help fix $60\%$ of the errors made on the original multi-hop questions. However, this technique exploits the knowledge about the structure of question and thus, is difficult to adapt when the structure of the questions can be flexible. Below we list some representatives methods for this approach:
\begin{itemize}
    \item \citet{SP14} use the hypothesis that each sub-question can be formed from a multi-hop question by copying and lightly editing a key span from it. Questions in HotpotQA are categorized into four categories: bridge (47\%), comparison (22\%), intersection (23\%)(questions asking for an entity that satisfy multiple properties) and others (8\%). Editing methods are proposed to be dependent on the question type for the first three categories.
    Question text is segmented into several spans by training a pointer network that predicts $p_{ij}$, the probability of the $i^{th}$ word to be the $j^{th}$ index to split the question. 400 annotations are manually generated for training. For each question, three such indices for bridge questions, two indices for intersection and four indices for comparison questions are predicted by maximizing the joint probability of the decomposition. These spans are modified slightly depending on the reasoning type. 
    Any single-hop QA model can be used for answering each single hop question. Concatenation of the question, the reasoning type, the answer, and the evidence is encoded using BERT and scored using 1 layer feed forward NN with sigmoid activation. The reasoning type is decided as the one resulting in the maximum score. 
    An alternative way where the reasoning type is predicted before decomposing and answering is tried and found to be giving poor results. To verify the original hypothesis, same technique is tested with span-based questions replaced by human written questions. There is a little difference in model performance indicating that the span-based sub-questions are as effective as free-form sub-questions.
    \item \citet{SP21} break down the task of MHQA into two components: Coarse grained decomposition and Fine grained interaction. The decomposition module is responsible for making the high dimensional vector distribution of entity nouns or pronouns more inclined to the intermediate answer to the question. The fine grained interaction module is a modified bidirectional attention module. The resulting context representations are passed to a self-attention layer and then used for supporting facts prediction by passing to a Bi-GRU layer.
    \item \citet{SP22} propose MHQA methods for four different datasets: HybridQA, QASPER, HotpotQA-Long and ShARC-Long. QASPER is originally proposed as single hop QA over long documents whereas ShARC is proposed as a conversational QA task. The paper utilizes the fact that long documents are often structured into sections and subsections which can be used as separate contexts with limited dependence among them. A pre-trained ETC model \citep{ainslie-etal-2020-etc} is used as the question encoder and the context encoder.
    ETC is a pre-trained Mask Language Model that employs a global-local attention mechanism. ETC assigns to each sentence a special global token that only attends to local tokens in the sentence, and its embedding is trained to summarize the information of local tokens in the sentence. ETC additionally adopts Contrastive Predictive Coding (CPC) \citep{van2018representation} to train the embedding of global tokens to make them aware of other sentences in the context. 
    It takes in a sequence of sentences and outputs the contextualized representations of each sentence. For conversational QA, the question is formed by concatenating the original question with a sentence formed by each pair of follow-up question and answer. ETC is again used to encode the question paragraph. For the 2-hop questions (HotpotQA, HybridQA), a null sentence is passed to ETC along with the question to get two question encodings. The paragraph embeddings are computed by a weighted sum of sentence embeddings, where weights are the attention score of the query vector to that sentence. At every step, each sentence and every paragraph in the document is attended to the query and the output is used to update the query representations. The updated query representation are then combined with the embedding of next query sentence to give the query vector for the next hop. 

\end{itemize}

\textbf{Critique:} \citet{SP14} also find that some of the questions are not composed of two single-hop questions but require implicit multi-hop reasoning, hence cannot be decomposed. Secondly, for some questions, answer for each sub-question does not exist explicitly in the text and has to be inferred with commonsense reasoning. 

\subsection{Miscellaneous Techniques}
In this section, we list some interesting approaches that do not fall into the above defined categories.
\begin{itemize}
    \item \textbf{Question generation:} \citet{LL18} propose multi-task training of a model to perform both multi-hop question answering and sub-questions generation. The motivation is that a model capable of asking good sub-questions corresponding to a complex question should perform well at MHQA. The system consists of a GNN based QA component and an LM based QG component that are trained together with a shared encoder. 
    \item \textbf{Entailment based MHQA:} Equation \ref{eq:def} presents MHQA roughly as an entailment task, where the premise consists of multiple contexts and the hypothesis is "a answers q" (referred as $H_{aq}$ henceforth). This indicates the utility of entailment models for the task. However, modelling MHQA as entailment faces three major challenges: a) A larger set of possible answers makes this method difficult to scale. b) MHQA requires aggregation of multiple contexts and single sentence based entailment models can not be used directly. c) Entailment models are usually not trained to filter irrelevant information. 
    \citet{SP28} aim to tackle the two latter challenges on Multiple choice datasets, OpenBookQA and MultiRC. Two simple baselines are proposed: \textit{Concatenate} that concatenates all sentences as the input to an entailment task, and \textit{Max of local decisions} that uses entailment on each sentence independently and then aggregate the results with a max operation. The two baselines perform poorly validating the challenges mentioned above. 
    Therefore, a novel method `Multee' comprising of a sentence relevance module and a multi-level aggregation is proposed. The sentence relevance module uses a pre-trained entailment model to produce hypothesis aware representation of each sentence which are passed to a Bi-LSTM layer to give contextual representation of each sentence which is fed to a feed forward layer to get the relevance scores $\alpha_i$. The multi-level aggregation module pass each sentence along with the hypothesis to $k$ ESIM \citep{chen-etal-2017-enhanced} entailment stacks to generate $k$ paragraph level vectors which are concatenated and passed to a feed forward layer to predict the entailment. 
    Each entailment stack has $m$ layers, $l$ of these layers process each sentence independently and the outputs are aggregated by using $\alpha_i$'s. Rest of the layers process this aggregated output to result in paragraph level embeddings. The aggregation is also done in the cross attention layer to produce a cross attention matrix containing attention between each hypothesis token and each paragraph token. The sentence relevance weights are used here as well. Each entailment stack used is pre-trained on SNLI \citep{bowman2015large} and MultiNLI \citep{williams-etal-2018-broad}.
    \item \textbf{Commonsense knowledge for MHQA:} \citet{SP27} argue that a pre-trained model may not be able to capture every grounded common-sense knowledge even with large corpora and propose using ConceptNet \citep{speer2016conceptnet} for extracting the same and using it, form reasoning paths leading to the answer. A tree of various reasoning paths, all starting from the question concepts ($c_1's$) is formed by following four steps: a) Selecting relations $r_1's$ from ConceptNet that link $c_1$ to another concept $c_2$ from the context. b) selecting relations $r_2's$ that link $c_2$ to another concept $c_3$ from the context. c) Selecting all neighbouring concepts $c_4$ of $c_3$ connected by $r_3$. d) Selecting relations $r_4$ that link $c_4$ to another concept $c_5$ from the context. This results in a large number of reasoning paths that are scored in two steps: a) $n-score$ is computed by term frequency of each of $c_2, c_3, c_5$ in the context $C$. $c_4$ is scored with its Point-wise Mutual Information (PMI) \citep{church-hanks-1990-word} with $c_{1-3}$. Each node's score is normalized across all its siblings in the tree using Softmax. b) $c-score$ The node scores are accumulated starting from the leaf node and updating each non-leaf node recursively. \begin{equation}
        c\_score(c_i) = \begin{cases}
            n\_score(c_i) & c_i\text{ is a leaf node} \\
            n\_score(c_i) + \frac{c\_score(c_{i+1}') + c\_score(c_{i+1}'')}{2} & otherwise
        \end{cases}
    \end{equation}
    where $c_{i+1}'$ and $c_{i+1}''$ are the top two scoring children of $c_i$. At each level of the tree, for every node, only the two top scoring children are maintained and the rest are pruned resulting in a total of $2^4$ reasoning paths. Context representation is attended to the commonsense representations formed by embedding the tokens in each of the reasoning paths. Updated context representation is combined with the original one using a sigmoid gate. The gate incorporates the fact that the commonsense might be optional. Hence, the unit is called NOIC (Necessary and Optional Information Cell). 
    \item \textbf{Reasoning over tables:} \citet{DP6} use BERT to encode a cell in a table where a cell is represented by its value, position, hyperlinks etc. Encoding of a cell along with the encodings of its neighbouring cells is fed to a feed forward model and a Softmax layer to find the cell for the next hop. The model can also hop to the same cell. The cell value is prepended to the hyperlinked passage and passed to the answer prediction module.

    \item \textbf{Pointer network for reasoning chain prediction:} \citet{SP19} encode the question and obtain query dependent paragraph encoding by using BERT. Sentence encodings are extracted from the paragraph encoding similar to \citet{SP20}. Alternative baselines are a) BERT-Sent that obtains query aware encoding of each sentence by using BERT and b) BiDAF-Para that uses BiDAF \citep{bidaf} to encode paragraphs. Results show that BERT-para performs the best. An LSTM Pointer Network is trained to predict probability of each sentence being the $t-{th}$ sentence in the reasoning chain i.e., $P(s_i = rc_t)$. The ground truth reasoning chain is determined using a chain extractor model discussed in Section \ref{subsec:aux}. The model is trained using both Negative Log Likelihood (NLL) and Reinforcement Learning (RL). However, RL does not improve the performance significantly. During test time, a beam of possible chains is maintained since the best chain may not be evident until it is entirely retrieved.
    \item \textbf{Bi-directional attention baseline:} \citet{DP1}  propose a baseline by modifying the architecture of \citet{SP45}. Question and the concatenation of paragraphs is passed to an RNN for combining the character and word level embeddings. Bi-directional attention is applied across the question and context embeddings to get query aware context representations. A residual connection is added to the output of another RNN on these representations before passing to a self-attention layer.
\end{itemize}

\section{Answer Prediction Module} 
After the reasoning module outputs a reasoning chain or some latent space representations of the reasoning, an answer prediction module predicts the final answer. This module can be directly categorized based on the type of answer required by the task at hand:

\subsection{Candidate Answering} 
This category corresponds to the multi-choice setting of the MHQA task.
Methods dealing with candidate answer prediction usually start the retrieval/reasoning process by using a candidate-aware embedding of the question, and output the representations for each candidate independently. We look at some of these methods below:
\begin{itemize}
    \item A Bi-directional attention layer on query and context \citep{SP12}, or a self-attention layer on the embeddings may also be employed.
    \item Some of the methods use these representations to perform an independent binary classification for each candidate (for multiple-correct type questions such as MultiRC) \citep{SP4, SP7, SP9, SP28}.
    \item On the other hand, some methods perform a multi-way classification for all candidates (for single-correct type questions such as OpenBookQA, WikiHop) \citep{SP7, SP9, SP12, SP15, SP19}.
    \item \citet{SP15} use an ensemble of five models, each trained with a different weight initialization. 
    \item \citet{SP7} use both the answering methods for QASC and find that the multi-way classification results in an improvement of ~5\% accuracy.

\end{itemize}

\subsection{Span Answering} 
For tasks requiring the answer as a text span from the contexts, the following methods have been proposed:
\begin{itemize}
    \item \citet{DP1} suggest a span answering approach where the output of the reasoning module is passed to another RNN for supporting facts prediction. Another RNN is used to predict the start and end token of the answer span. Finally, a three-way classifier layer is used to predict the answer type among 'yes', 'no' and the extracted span. Many methods focusing on the first two units use this baseline as the answer prediction unit \citep{SP2, SP3, SP10, SP16, SP21, SP25, SP26}.
    \item \citet{SP24} make two changes to the above baseline: a) Concatenating passages before encoding makes the representations depend on the order while concatenating. Thus, a shared RNN encoder first encodes each passage independently and then concatenates the representations. b) All the attention layers are replaced by self-attention on the concatenated question and context representation.
    \item \citet{SP1, SP6} replace the RNNs in the above baseline by MLPs and directly feed the node representations for answering. \citet{SP5} deal only with bridge question and hence use a single MLP for span prediction on entities. \citet{SP13} only deal with supporting facts prediction and use a single MLP for the same.
    \item \citet{SP8} use an AlBERT model to predict the answer span among the $K$ identified passages. The model then predicts if the answer was found in the given passage. If not, the predicted span is used for iterative retrieval for the next hop.
    \item \citet{SP14} use an off-the-shelf single-hop answering module to answer the sub-questions formed by the reasoner.
    \item \citet{SP17} use a simple BERT model to answer the question from the paragraph identified as the answer paragraph. \citet{SP23} do the same from the concatenation of the two identified passages.
    \item \citet{SP19} concatenate question and all sentences of all the retrieved reasoning chains and feed to BERT to perform the 4 tasks.
    \item \citet{SP20} propose to use the output of the GNN model. Sentence scores are computed using MLPs on the sentence node representations. Paragraph scores are computed using MLPs on max pooling over the sentence representations. The two scores are combined with the span extraction score to predict the final answer. Separate MLPs are used for predicting the answer type and the supporting facts.
    \item \citet{SP22} Concatenate the embeddings at all retrieved steps $\{k^0, \cdots, k^\ast, \cdots, k^0, \cdots, k^\ast \}$ and perform a weighted sum to get $\mathcal{K}$ that is used to make the final prediction. The Softmax weight $\mathcal{Y}_j$ is computed across the retrieved sentences from all steps.
    % ( small section to Long docs as MHQA and the datasets?)
    % - HybridQA: each cell converted to a paragraph using column name + text + hyperlink para
    % - QASPER: NLP papers as docs (55\% multi-hop) treat each subsection as a paragraph.

\end{itemize}
\subsection{Generative Answering} 
Previously, generative answering models had a low prevalence because of the unsatisfactory performance of models to generate relevant texts as well as the lack of reliable evaluation techniques. As a representative of the earlier approaches, \citet{SP27} use a self-attention layer followed by a PGN decoder to get the final answer. 
\par
However, with the recent advancements relating to large language models (LLMs), generative answering approaches have become a lot more popular. We describe these methods in detail in Chapter \ref{sec:llm} designated to LLMs for MHQA.

\section{Auxiliary Tasks} \label{subsec:aux}
A lot of existing approaches suggest using an auxiliary training task that can help the training of the model by providing an extra signal. Here are two of the commonly used auxiliary tasks:
\subsection{Reasoning Chain Prediction} Reasoning chains are an integral part of explainable MHQA. HotpotQA contains the supporting facts that are required to answer the question, however these are not ordered. Several works have aimed to predict the order among these supporting facts by using simple lexical heuristics \citep{AP6, AP5, AP2} so that the reasoning chain formed can be used to further train or evaluate a model. However, it is also crucial for the models to be able to predict the reasoning chains only by using the question and contexts.
\begin{itemize}
    \item \citet{SP18} propose a semi supervised reinforcement learning for two modules to recover the reasoning chain in a \textit{cooperative-game} approach. Apart from predicting the reasoning chain, the model is also able to predict the relations among the sentences. The two modules are: 
    a) Ranker module that, given $k$ passages and a question, selects a reasoning chain. Question and passages are encoded by bi-GRU \citep{cho2014learning} and a Match-LSTM \citep{wang2016machine} model is used to get a probability for each passage containing a part of the reasoning chain. A passage is then sampled and used to update the question by passing to an MLP. The process is repeated with the updated question. The ranker module is rewarded for selecting the correct passage at the correct reasoning step.
    b) Reasoner module that predicts the entity from the current passage to the next passage. Given the first passage (called head) selected by the trained Ranker, the Reasoner uses a Math-LSTM model to predict the probability of each entity appearing in the second passage (called tail).
    While the ordering of the supporting facts is generated similar to \citet{DP1} for HotpotQA, the reasoning chains need to be manually annotated for MedHop. The manual annotation is done by extracting all valid paths such that the first sentence contains an entity in the second sentence and the second contains the answer. A chain is manually labelled as positive if the corresponding passage describes the drug-protein interaction. Reasoning chains extracted in this way may not be unique.
    \item \citet{SP19} derive pseudo gold chains using NER and co-reference resolution. A sentence (only) graph is constructed with edges between sentences that belong to the same paragraph and between those that have a shared entity. Reasoning chains are collected by starting from the question node and finding all possible paths in the graph that lead to an answer containing sentence. These chains are then ranked by two heuristics: a) shorter chains are given higher scores. b) chains whose sentences have a high F-1 ROUGE overlap with the question are given higher scores. Experimental results indicate that b) is a good criteria for scoring chains. Human evaluation shows that the reasoning chains produced are of similar quality compared to the supporting facts present in the HotpotQA. 
    A pointer network is then trained to predict each sentence in the reasoning chains.
\end{itemize}
\subsection{QA with Partial Knowledge} \citet{SP29}  propose a new sub task of MHQA where the model has to figure out and fill the knowledge gap for question answering. It is assumed that the first hop retrieval has been performed ideally and the task is to use the retrieved context to answer the question. A modified version of OpenBookQA is released where the core fact is part of the input and several relations have been modified.

\section{Conclusion}
In this chapter, we covered a large number of pre-LLM machine learning methods proposed to solve MHQA. While doing so, we provided some structural patters that many methods follow. We also provided various levels of classifications of the methods, with technical details for some representative methods pf each category or subcategory. In chapter \ref{sec:tax}, we build on this classification to propose a taxonomy covering most of the works discussed in this chapter. In the next chapter, we provide details for how LLMs have been incorporated for different stages of the task along with some challenges and their proposed solutions.
\chapter{LLMs for MHQA} \label{sec:llm}
Large Language Models (LLMs), including the multiple variants of GPT, BERT and T5 models, have achieved remarkable success in various natural language tasks. We present a comprehensive background of LLMs and various prompting techniques in the Appendix \ref{sec:llm-back}. We also recommend \citet{llm-survey} as an additional reading for a comprehensive survey on LLMs.
\par 
Since language models like T5, BERT have been used for multiple tasks in most methods we have discussed so far, we use the following distinction for LLM-based methods: A method is LLM-based if it makes use of the emergent abilities of the LLMs \citep{llm-survey} which include in-context learning and instruction following\footnote{This distinction is made for the sole purpose of better structuring the discussion.}.
We categorize LLM based methods based on the sub-task the LLM performs.

\section{Retrieval}
\citet{LL20} explore using LLMs for retrieving relevant evidences from long documents. Major challenge in using LLMs for retrieval is their limited context window size. The performance decreases drastically as the input context becomes larger than the context window size. To deal with this challenge, each section in the document is independently summarized by an LLM. 
These summaries are concatenated together in another LLM prompt, where the instruction is to get a list of relevant sections. Once a set of relevant sections is obtained, the next step is to retrieve the relevant paragraphs from each of these sections. 
For this, each paragraph is summarized and passed to the LLM to get the set of relevant paragraphs. A bart-large \citep{bart} trained over the CNN/Daily-Mail Corpus \citep{cnn} is used for summarization, and GPT-3.5 is used for retrieval and answering.

\section{Reasoning Chain Generation}
Existing works have extensively explored GPT prompting for various QA tasks. Although few-shot prompting on LLMs for reasoning showed limited success \citep{folio}, chain of thought prompting (CoT) has shown better reasoning abilities. CoT is particularly useful for explainable MHQA since the generated chain of thought can be used as the reasoning chain \citep{gpt-qa5, gpt-qa6}. \cite{saparov} use synthetic data to evaluate GPT-3's reasoning ability and measure the coherence of generated reasoning chains. Results indicate that GPT-3 can perform multiple steps of reasoning but may rely on background knowledge rather than explicit reasoning over the given context.
\par
Below, we list a few works and describe in detail how the LLM was used for generating reasoning chains.
\begin{itemize}
    \item \cite{LL13} evaluate T5 and Flan-T5 \citep{flant5} models with and without fine-tuning on chain-of-thought explanations and in zero-shot, few-shot and CoT prompting settings on complex multi-hop queries. The results indicate the need for improvement in the existing methods.
    \item \cite{gpt-qa5, gpt-qa6} get promising results by directly using the chain-of-though explanations as reasoning chains. 
    \item PathFiD \cite{LL15} was one of the first to use a large language model for modeling the task as a sequence generation problem. However, the task was not modelled as a free-text natural language generation. A pre-trained T5 model was fine-tuned to generate a `reasoning path'. A reasoning path is defined as a sequence of sentences from the context passage along with the corresponding passage title, followed by the answer to the question. The encoder independently encodes the `blocks' (sentence and corresponding paragraph title) in the retrieved context and the decoder decodes the reasoning path by selecting the relevant blocks and deriving the answer.
    \item \citet{LL16} extend the idea of PathFiD by proposing the LLM to generate parallel exploratory inference chains (EICs). They follow a multi-step generation process, prompting an LLM multiple times with different prompts in the process. During the first step, a LLM is prompted to generate some keywords from the question and the paragraph titles. These keywords are then processed independently in parallel. A keyword is matched lexically with available paragraph titles to retrieve the paragraphs for the next step, and the LLM is prompted with the retrieved passage to predict some structured facts about the keyword. The object in these generated fact is used as a keyword for the next reasoning step. Two independent LLMs are prompted to aggregate the facts for each keyword and to aggregate facts across keywords to generate a final answer. Impressively, the resulting system is able to significantly outperform a single chain-of-thought prompt.
    \item \cite{LL12} propose IRCoT which uses interleaving-retrieval with CoT reasoning to iteratively perform retrieval and reasoning together. For each iteration, 1) the CoT-guided retrieval step (“Retrieve”) uses the last generated CoT sentence as a query to retrieve more paragraphs and adds them to the accumulating set of the collected paragraphs, and 2)the retrieval-guided reasoning step generates the next sentence of the CoT using the previously generated sentences along with the retrieved contexts. Finally, another LLM derives the answer using the generated CoT and the retrieved paragraphs.
\end{itemize}

\section{Hybrid Text-table Reasoning}
The following methods for utilizing LLMs for solving hybrid MHQA have been proposed:
\begin{itemize}
    \item \citet{LL17} were one of the first to use LLMs for solving table-text hybrid MHQA. They train a sequence-to-sequence LLM similar to \citet{LL15}, to encode the retrieved contexts and question. The decoder, however, performs auto-regressive generation to predict the answer sentence. They also compare a prompting based approach where the retrieved contexts are provided to GPT-3.5 along with few-shot and CoT prompts. The results showed promising performance of CoT but still the fine-tuned model outperformed all the prompting techniques. 
    \item \citet{finqa} propose converting a table into text using template based transformation and then use a RoBERTa to retrieve sentences from the combined set of original text and converted text.
    \item \citet{mavi} test the GPT-3.5 and Llama2's ability to parse tabular data by representing a table in text by using separators such as $\vert$ and $\vert\vert$ to separate cells in a row and multiple rows respectively and observe promising performance on tabular reasoning.
    \item \citet{LL25} explore using program generation and execution based framework for hybrid MHQA and get promising results.
\end{itemize}

\section{Question Decomposition}
Since GPT has shown remarkable language understanding, motivating its use as a decomposition module for decomposition based MHQA (\cite{gpt-qa2, gpt-qa4}).
\begin{itemize}
    \item \citet{LL1} explore large-scale intermediate pre-training of a T5 model to perform question decomposition. For constructing a large enough dataset for pre-training, distant supervision from comparable texts is used. In particular, parallel news articles describing the same events are used. The hypothesis is that seeing multiple descriptions of the same facts will help the model make better educated guesses. The proposed model, called DecompEntail, first generates explicit question decomposition, then makes factual corrections on the decomposed statements with GPT-3. As a final step, an entailment model is used to derive the final answer with the generated decomposition as the premise and the question and candidate answer as the hypothesis. 
    \item \citet{LL35} perform question decomposition in the semantic space by using an AMR graph. They build an AMR graph from the question and add an additional `amr-unknown' node denoting a concept that represents the answer. The AMR is segmented into sub-graphs using some heuristics and each sub-graph is passed to a pre-trained BART model that converts it into a sub-question. Finally, a single-hop QA model is used for answering the sub-questions.
    \item \citet{LL26} fine-tune a sequence-to-sequence LLMs to generate sub-questions using the given multi-hop question. A set of retrieved context is also given as input during this step. The sub-questions are used for performing a second round of retrieval using a DeBERTa model \citep{deberta}. Another DeBERTa model is used on the concatenation of the retrieved paragraphs and sub-questions to generate the final answer.
\end{itemize}

\section{Graph Construction}
\citet{LL19} prompt an LLM with a Wikipedia passage to build a semantic graph. This is done by prompting the LLM to extract all the entities in the document and then prompt it again to get all the relations among these entities. Finally, each passage and the corresponding semantic graphs are concatenated into a prompt for generating reasoning chain and the final answer.

\section{Multi-hop Retrieval Augmented Generation}
\textbf{Retrieval Augmented Generation (RAG)} is the term given to the general framework where a generative model (usually a LLM) performs generation by using a set of facts extracted by a retriever. In RAG, an external corpus containing multiple documents serves as the knowledge base. Each document within this corpus is segmented into a set of chunks. These chunks are then transformed into vector representations using an embedding model and stored in an embedding database. Given a user query, the system typically retrieves the top-K chunks that best match the query. The retrieved chunks, combined with the query and an optional prompt, are then fed into an LLM to generate a final answer. 
\par
\cite{LL23} build a novel dataset for the development and benchmarking of multi-hop RAG models. The dataset contains a knowledge base for retrieval, a set of multi-hop queries along with their ground truth answers and corresponding supporting facts. The knowledge base is built from a set of news articles. GPT-4 is prompted to extract factual sentences from the news articles

\section{Critique and Limitations}
Despite the impressive linguistic and reasoning abilities of LLMs, they face certain limitations. In this section, we list these limitations and some attempts at resolving these.
\subsection{Hallucinations}
The biggest challenge that the LLMs currently face is that they are known to hallucinate at times, i.e., generating text that is factually incorrect or not derivable from the given information. This poses a threat to accountability of the resulting MHQA system and can produce incorrect results. A large number of works have tried to propose sophisticated and clever ways of overcoming this challenge. We describe few such interesting works in detail below:
\begin{itemize}
    \item \textbf{Self-consistency:} Self-consistency \citep{LL33} is proposed as an alternative for the naive greedy decoding used in chain-of-thought prompting. Instead of only taking the greedy decoded reasoning path, a diverse set of reasoning paths are sampled, and then the most consistent answer across the reasoning paths is predicted by marginalizing out the sampled reasoning paths. Self-consistency is motivated by the intuition that a complex reasoning problem typically admits multiple different ways of thinking leading to its unique correct answer. Assuming that the hallucinations in the LLM output are random, self-consistency is more likely to lead to the correct answer since it is unlikely that majority of the reasoning paths with hallucinations lead to a common answer. Therefore, self-consistency has been adopted to mitigate hallucinations.
    \item \textbf{Self-refine:} In self-refine \citep{LL34}, the LLM generates an initial response to the prompt, and then iteratively provides feedback for the output and refines it. The same LLM serves as the generator, feedback provider as well as the refiner. During self-refine, the LLM can also evaluate the response on factual correctness, thus fixing the hallucinations.
    \item \cite{LL21} extend self-consistency and self-refine by fine-tuning external LLMs for providing feedback on the generated reasoning chains. The authors collect labelled data for correctness, error types and descriptions, and the required corrections for each reasoning chain and answer pairs for 2361 examples. They use a weighted self-consistency approach where a fine-tuned Llama2 predicts a score for whether each answer and its corresponding reasoning path are correct or not. This score is used as the weight while voting for the correct answer. They also explore extending self-refine by using a fine-tuned Llama2 model to identify the errors in a reasoning path and refining the generated output. 
    \item \cite{LL6} propose `Verify and Edit' that combines self-consistency and self-refine. It follows a three-step process: finding uncertain predictions using self-consistency, editing their rationales by searching for supporting facts, and using the edited rationales to generate final answers. An answer is considered as uncertain, if none of the generated answers gets the majority agreement. For each of the predictions corresponding to the uncertain answer, each sentence in the reasoning chain is verified by generating corresponding questions. These questions are answered after retrieval and the answer is used to update the particular sentence in the original reasoning chain. Finally, the updated reasoning chain is used to generate the final corrected answer.
    \item \cite{LL3} propose a claim decomposition based technique for self-evaluation of the LLM generated answers. The complex question is broken down into a set of claims that a correct answer to the question must satisfy. The LLM then identifies which claims mention the same entities and include extra tags for them. For evaluating a generated answer, the answer is replaced in each of the claims and the LLM is prompted to answer true/false based on whether the answer satisfies the claim or not. The answer is then scored on the ratio of claims satisfied by it. Multiple answers are predicted by sampling based generation of the reasoning chains and the one with the highest score is predicted.
\end{itemize}
\subsection{Knowledge Gaps}
\cite{LL24} argue that the challenge of hallucinations is difficult to overcome since the LLMs might always suffer from knowledge gaps. A knowledge gap here refers to some missing or outdated information. Since LLMs heavily rely on knowledge gained during the pre-training phase, outdated or incomplete information in the pre-training corpus can confuse the model, leading to generating factually incorrect text. Further, pre-training LLMs is extremely expensive in terms of time and compute. It is therefore ideal that the LLM is able to detect a knowledge gap and refrain from generating information in such a case. For identifying knowledge gaps, the following approaches can be followed:
\begin{itemize}
    \item \textbf{Calibration-based:} getting a probability score for an answer by either using token probabilities and temperature during generation \citep{mtlearner, holisticeval} or directly asking the model to generate its confidence in its output \citep{tian2023just}. 
    \item \textbf{Training based:} training an extra layer \citep{hidden}, or an external module \citep{cobbe2021training} to predict whether the answer produced has a high enough confidence. Another interesting way is to use identifying knowledge gaps as one of the instruction tuning tasks \citep{instr}.
    \item \textbf{Prompting based:} techniques such as self-reflect \citep{selfreflect}, self-consistency \citep{LL33} predicting none-of-the-above \citep{selfreflect} or more information needed \citep{moreinfo} instead of predicting the answer.
    \item \textbf{Multi-LLM collaboration:} using two independently trained LLMs for either providing feedback to each other, or complementing each other to fill the knowledge gaps in both \citep{LL24}.
\end{itemize}

% \subsection{Context window size}
% Another limitation is the limited context window size which restricts the amount of information that can be passed to the LLMs in one go. Following directions have been explored as possible solutions to the problem:
% \begin{itemize}
%     \item \textbf{Context reduction:} These methods tackle the problem by reducing the effective size of the contexts such that it fits within the context window. Possible methods for this include fine-grained retrieval \citep{mavi}, 
%     \item \textbf{Fine-tuning:}
%     \item \textbf{Position embeddings}
% \end{itemize}

\chapter{MHQA Taxonomy} \label{sec:tax}

\begin{figure}[h]
\includegraphics[width=\textwidth]{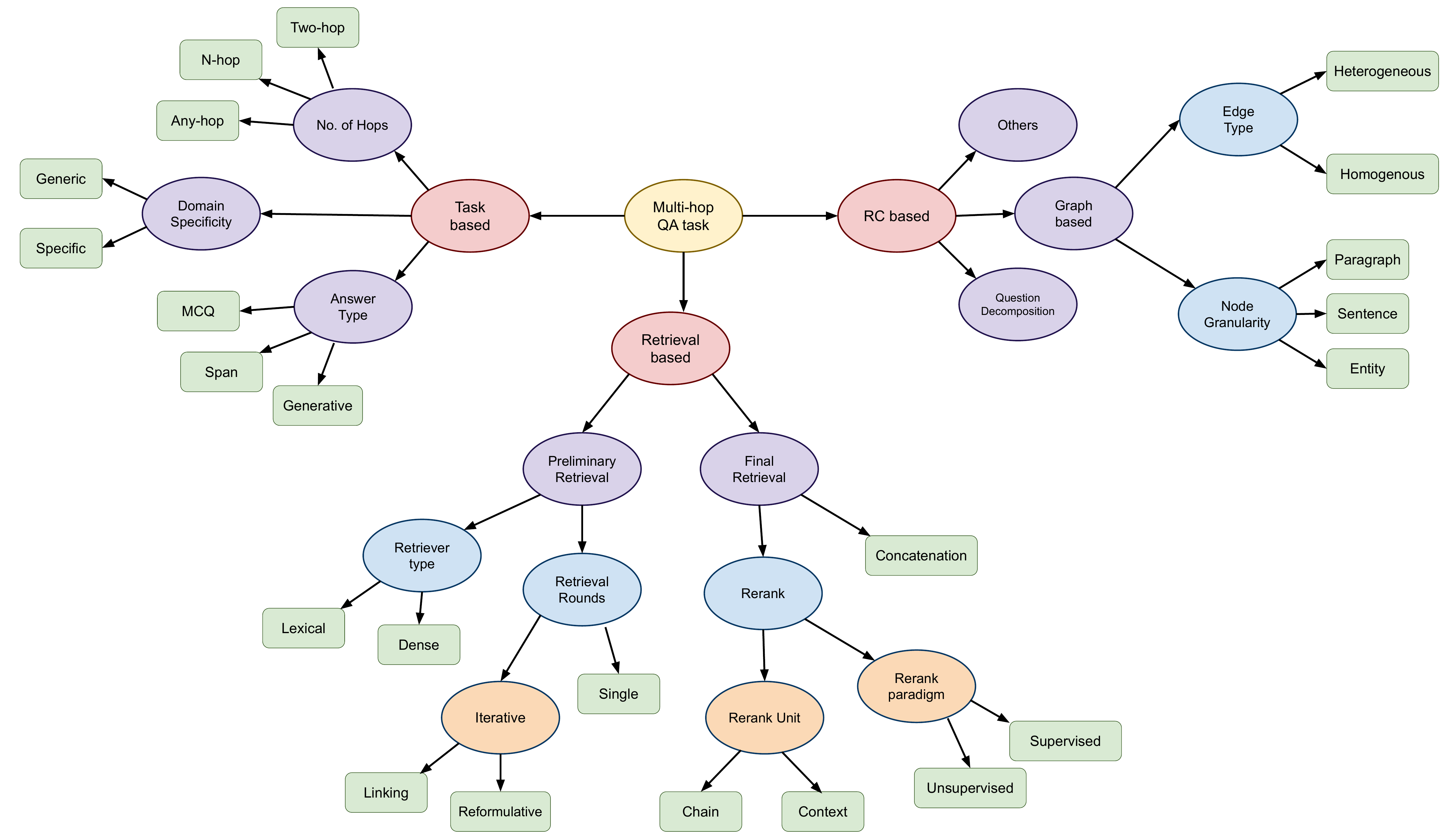}\centering
\caption{Overview of Our Taxonomy.} \label{fig-taxonomy}
\end{figure}

This chapter builds on the distinctions and classifications drawn in Chapter \ref{sec:method} and our taxonomy that covers the existing methods in a structured way. Creating a taxonomy of existing methods for multi-hop QA is key for sorting through research and making comparisons. It also helps spotting where the community needs to pay more attention.
We summarize these in Table \ref{tab:taxonomy}. The titles and acronyms used in the table are described below, and illustrated in Figure \ref{fig-taxonomy}.

\begin{itemize}
    \item \textbf{General:}
    \begin{itemize}
        \item \textbf{Datasets used\footnote{Note that we do not include the datasets used in the proposed Taxonomy, but describe it in this section for the sake of completionstill add it in the Table \ref{tab:taxonomy} for the sake of completeness.}:} The datasets used for MHQA along with their acronyms in Table \ref{tab:taxonomy} are: HotpotQA \citep{DP1} in the full-wiki (\textbf{HP-F}) and distractor (\textbf{HP-D}) settings. \citet{SP22} provide the modified versions of the ShARC \citep{saeidi2018interpretation} and the HotpotQA datasets as ShARC-Long (\textbf{Sh-L}) and HotpotQA-Long (\textbf{HP-L}). Other datasets are referenced as: MultiRC \citep{DP5} (\textbf{MRC}), ARC \citep{clark2018think} (\textbf{ARC}), QAngaroo datasets \citep{DP2} WikiHop (\textbf{WH}) and MedHop (\textbf{MH}), HybridQA \citep{DP6} (\textbf{Hy}), QASPER \citep{dasigi-etal-2021-dataset} (\textbf{QSP}), OpenBookQA \citep{DP4} (\textbf{OB}), QASC\citep{DP7} (\textbf{QSC}) and \textbf{Nr} \citep{DP3}.
        
        \item \textbf{Hop Constraints:} Some of the methods require the type of questions to be exactly two-hop questions (\textbf{Two}) or require the number of hops in the question to be provided as an input (\textbf{N}). Other methods are flexible for answering any-hop questions without any additional input (\textbf{Any}).
        
        \item \textbf{Answer types:} Whether the methods proposed work for MCQ type questions (\textbf{MCQ}), Span based questions (\textbf{Span}) and generative answers (\textbf{Gen}). Note that some methods focus on more than one of these answer types.
        
        \item \textbf{Domain Specificity:} Whether the works focus on domain specific (\textbf{Specific}) or domain generic techniques (\textbf{Generic}).
    \end{itemize}
    
    \item \textbf{Retrieval:}
    \begin{itemize}
        \item \textbf{Retriever Type:} Whether the preliminary retrieval step uses a dense (\textbf{Dense}) or a lexical (\textbf{Lexical}) retriever.
        \item \textbf{No. of Retrieval Passes:} Whether the preliminary step is single pass (\textbf{Single}) or iterative (\textbf{Iter}). In case it is iterative, whether it uses query reformulation (\textbf{Iter-QR}) or Entity-links/Hyperlinks (\textbf{Iter-EL/H}).
        \item \textbf{Final Retrieval Strategy} Whether the final retrieval step uses a simple concatenation/union (\textbf{Concat}) of the contexts or it is uses a re-ranking approach (\textbf{RR}). In case, it uses re-ranking, whether the re-ranking is supervised (\textbf{RR-S-?}) or unsupervised (\textbf{RR-U-?}), and whether the individual contexts (\textbf{RR-?-X}) or the entire chain candidates (\textbf{RR-?-C}) are scored during the re-ranking.
    \end{itemize}
    
    \item \textbf{RC based:}
    \begin{itemize}
        \item \textbf{Node Granularity:} In case the reasoner is a graph based model, what kinds of nodes it has: entity nodes (\textbf{Ent}), sentence nodes (\textbf{Snt}) or passage nodes (\textbf{Psg}). Note that the graph may have more than one type of nodes. (\textbf{-}) in case the reasoner is not graph based.
        \item \textbf{Relational Edge:} Whether the graph is a relational graph (\textbf{Yes}) (i.e., has multiple types of edges) or not (\textbf{No}) (i.e., has single edge type). (-) in case the reasoner is not graph based.
        \item \textbf{Question Decomposition:} Whether the reasoning module uses question decomposition (\textbf{QuesD}) or not (\textbf{Other}). The rows where both graph-based and decomposition-based columns are (\textbf{-}) are for the models that use other techniques like \citep{bidaf}.
    \end{itemize}
\end{itemize}

By proposing the taxonomy, we hope to help the readers dive into these techniques, see what suits best to their needs or what needs more attention in the existing setup. This should help the research community keep pushing the boundaries for solving MHQA.

\begin{landscape}
\tiny
\centering
\label{tab:taxonomy}
\begin{table}[h]
\caption{Comprehensive study of existing work using the proposed taxonomy.}
\resizebox{18cm}{!}{
\begin{tabular}{|l|l|l|l|l|lll|llll}
\hline
\multicolumn{1}{|c|}{\multirow{2}{*}{}} & \multicolumn{4}{c|}{General} & \multicolumn{3}{c|}{Retrieval} & \multicolumn{3}{c|}{Reasoning} \\ \cline{2-11} 

\multicolumn{1}{|c|}{} & 
% \multicolumn{1}{c|}{\multirow{2}{*}{\begin{tabular}[c]{@{}c@{}}\end{tabular}}} &
\multicolumn{1}{c|}{Datasets} & \multicolumn{1}{c|}{\begin{tabular}[c]{@{}c@{}}Hop\\  Constraints\end{tabular}} & \multicolumn{1}{c|}{\begin{tabular}[c]{@{}c@{}}Answer\\  type\end{tabular}} & \multicolumn{1}{c|}{\begin{tabular}[c]{@{}c@{}}Domain\\  Specificity\end{tabular}} & \multicolumn{2}{c|}{Preliminary} & \multicolumn{1}{c|}{Final} & \multicolumn{2}{c|}{Graph based} & \multicolumn{1}{c|}{\multirow{2}{*}{\begin{tabular}[c]{@{}c}Ques\\ Decomp\end{tabular}}}  \\ \cline{6-10}

\multicolumn{1}{|c|}{} & \multicolumn{1}{c|}{} & \multicolumn{1}{c|}{
% \begin{tabular}[c]{@{}c@{}}Two-hop\\ /N-hop\\ /Any-hop\end{tabular}
} & \multicolumn{1}{c|}{
% \begin{tabular}[c]{@{}c@{}}Mcq/\\ Span/\\ Gen\end{tabular}
} & \multicolumn{1}{c|}{
% \begin{tabular}[c]{@{}c@{}}Specific\\ /Generic\end{tabular}
} & \multicolumn{1}{c|}{ Type } & \multicolumn{1}{c|}{ Technique} & \multicolumn{1}{c|}{ Technique} & \multicolumn{1}{c|}{Node types} & \multicolumn{1}{c|}{ \begin{tabular}[c]{@{}c@{}}Relational \\ graph?\end{tabular}} & \multicolumn{1}{c|}{}  \\ \hline

\citet{SP1} & HP-F & N & Span & Generic & \multicolumn{1}{l|}{Lexical} & \multicolumn{1}{l|}{Iter-EL/H} & Concat & \multicolumn{1}{l|}{Ent} & \multicolumn{1}{l|}{Yes} & \multicolumn{1}{l|}{Other} \\ \hline
\citet{SP2} & HP-F & N & Span & Generic & \multicolumn{1}{l|}{Dense} & \multicolumn{1}{l|}{Iter-QR} & RR-S-X & \multicolumn{1}{l|}{-} & \multicolumn{1}{l|}{-} & \multicolumn{1}{l|}{Other} \\ \hline
\citet{SP3} & HP-F & Two & MCQ & Generic & \multicolumn{1}{l|}{Lexical} & \multicolumn{1}{l|}{Iter-EL/H} & RR-S-C & \multicolumn{1}{l|}{-} & \multicolumn{1}{l|}{-} & \multicolumn{1}{l|}{Other} \\ \hline
\citet{SP4} & ARC, MRC & Any & MCQ & Generic & \multicolumn{1}{l|}{Lexical} & \multicolumn{1}{l|}{Single} & RR-U-C & \multicolumn{1}{l|}{-} & \multicolumn{1}{l|}{-} & \multicolumn{1}{l|}{Other} \\ \hline
\citet{SP5} & HP-F & Two & Span & Specific & \multicolumn{1}{l|}{Lexical} & \multicolumn{1}{l|}{Iter-EL/H} & RR-S-X & \multicolumn{1}{l|}{-} & \multicolumn{1}{l|}{-} & \multicolumn{1}{l|}{Other} \\ \hline
\citet{SP6} & HP-F & Two & MCQ & Generic & \multicolumn{1}{l|}{Lexical} & \multicolumn{1}{l|}{Iter-EL/H} & RR-S-C & \multicolumn{1}{l|}{Ent-Snt-Psg} & \multicolumn{1}{l|}{No} & \multicolumn{1}{l|}{Other} \\ \hline
\citet{SP7} & MRC, QSC & Any & MCQ & Specific & \multicolumn{1}{l|}{Lexical} & \multicolumn{1}{l|}{Iter-QR} & Concat & \multicolumn{1}{l|}{-} & \multicolumn{1}{l|}{-} & \multicolumn{1}{l|}{Other} \\ \hline
\citet{SP8} & HP-F & Any & Span & Generic & \multicolumn{1}{l|}{Lexical} & \multicolumn{1}{l|}{Iter-QR} & RR-S-X & \multicolumn{1}{l|}{Psg} & \multicolumn{1}{l|}{Yes} & \multicolumn{1}{l|}{Other} \\ \hline
\citet{SP9} & MRC, QSC & Two & MCQ & Generic & \multicolumn{1}{l|}{Dense} & \multicolumn{1}{l|}{Iter-QR} & RR-S-C & \multicolumn{1}{l|}{-} & \multicolumn{1}{l|}{-} & \multicolumn{1}{l|}{Other} \\ \hline
\citet{SP10} & HP-F & Two & Span & Generic & \multicolumn{1}{l|}{Lexical} & \multicolumn{1}{l|}{Iter-EL/H} & RR-S-X & \multicolumn{1}{l|}{-} & \multicolumn{1}{l|}{-} & \multicolumn{1}{l|}{Other} \\ \hline
\citet{SP11} & HP-F & Two & Span & Generic & \multicolumn{1}{l|}{Dense} & \multicolumn{1}{l|}{Single} & RR-S-X & \multicolumn{1}{l|}{-} & \multicolumn{1}{l|}{-} & \multicolumn{1}{l|}{Other} \\ \hline
\citet{SP12} & WH & N & Span & Generic & \multicolumn{1}{l|}{-} & \multicolumn{1}{l|}{-} & Concat & \multicolumn{1}{l|}{Ent-Snt-Psg} & \multicolumn{1}{l|}{No} & \multicolumn{1}{l|}{Other} \\ \hline
\citet{SP13} & HP-D & N & Span & Generic & \multicolumn{1}{l|}{-} & \multicolumn{1}{l|}{-} & Concat & \multicolumn{1}{l|}{Snt-Psg} & \multicolumn{1}{l|}{No} & \multicolumn{1}{l|}{Other} \\ \hline
\citet{SP14} & HP-D & Two & Span & Generic & \multicolumn{1}{l|}{-} & \multicolumn{1}{l|}{-} & Concat & \multicolumn{1}{l|}{-} & \multicolumn{1}{l|}{-} & \multicolumn{1}{l|}{QuesD} \\ \hline
\citet{SP15} & WH & N & Span & Generic & \multicolumn{1}{l|}{-} & \multicolumn{1}{l|}{-} & Concat & \multicolumn{1}{l|}{Ent} & \multicolumn{1}{l|}{No} & \multicolumn{1}{l|}{Other} \\ \hline
\citet{SP16} & HP-D & N & Span & Generic & \multicolumn{1}{l|}{-} & \multicolumn{1}{l|}{-} & RR-S-X & \multicolumn{1}{l|}{Ent-Snt} & \multicolumn{1}{l|}{No} & \multicolumn{1}{l|}{Other} \\ \hline
\citet{SP17} & HP-D & Any & Span & Generic & \multicolumn{1}{l|}{Lexical} & \multicolumn{1}{l|}{Iter-QR} & RR-S-X & \multicolumn{1}{l|}{-} & \multicolumn{1}{l|}{-} & \multicolumn{1}{l|}{QuesD} \\ \hline
\citet{SP18} & MH, HP-D & N & Span & Specific & \multicolumn{1}{l|}{Dense} & \multicolumn{1}{l|}{Single} & RR-S-C & \multicolumn{1}{l|}{-} & \multicolumn{1}{l|}{-} & \multicolumn{1}{l|}{Other} \\ \hline
\citet{SP19} & WH, HP-D & N & Span & Generic & \multicolumn{1}{l|}{-} & \multicolumn{1}{l|}{-} & Concat & \multicolumn{1}{l|}{Snt} & \multicolumn{1}{l|}{No} & \multicolumn{1}{l|}{Other} \\ \hline
\citet{SP20} & HP-D & Any & Span & Generic & \multicolumn{1}{l|}{-} & \multicolumn{1}{l|}{-} & Concat & \multicolumn{1}{l|}{Snt} & \multicolumn{1}{l|}{No} & \multicolumn{1}{l|}{Other} \\ \hline
\citet{SP21} & HP-D & Any & Span & Generic & \multicolumn{1}{l|}{-} & \multicolumn{1}{l|}{-} & Concat & \multicolumn{1}{l|}{-} & \multicolumn{1}{l|}{-} & \multicolumn{1}{l|}{QuesD} \\ \hline
\citet{SP22} & {\begin{tabular}[l]{@{}l@{}}HP-L, Sh-L\\ QSP, Hy\end{tabular}} & N & Span & Specific & \multicolumn{1}{l|}{-} & \multicolumn{1}{l|}{-} & Concat & \multicolumn{1}{l|}{-} & \multicolumn{1}{l|}{-} & \multicolumn{1}{l|}{QuesD} \\ \hline
\citet{SP23} & WH, HP-D & Two & Span & Generic & \multicolumn{1}{l|}{-} & \multicolumn{1}{l|}{-} & RR-S-C & \multicolumn{1}{l|}{-} & \multicolumn{1}{l|}{-} & \multicolumn{1}{l|}{Other} \\ \hline
\citet{SP24} & HP-D & N & Span & Generic & \multicolumn{1}{l|}{Lexical} & \multicolumn{1}{l|}{Iter-QR} & Concat & \multicolumn{1}{l|}{-} & \multicolumn{1}{l|}{-} & \multicolumn{1}{l|}{Other} \\ \hline
\citet{SP25} & HP-D & N & Span & Generic & \multicolumn{1}{l|}{-} & \multicolumn{1}{l|}{-} & RR-S-X & \multicolumn{1}{l|}{Ent} & \multicolumn{1}{l|}{No} & \multicolumn{1}{l|}{Other} \\ \hline
\citet{SP26} & HP-D & N & Span & Generic & \multicolumn{1}{l|}{-} & \multicolumn{1}{l|}{-} & RR-S-X & \multicolumn{1}{l|}{-} & \multicolumn{1}{l|}{-} & \multicolumn{1}{l|}{Other} \\ \hline
\citet{SP27} & Nr & N & Gen & Specific & \multicolumn{1}{l|}{-} & \multicolumn{1}{l|}{-} & RR-S-X & \multicolumn{1}{l|}{Ent} & \multicolumn{1}{l|}{Yes} & \multicolumn{1}{l|}{Other} \\ \hline
\citet{SP28} & MRC, OB & Any & MCQ & Generic & \multicolumn{1}{l|}{-} & \multicolumn{1}{l|}{-} & Concat & \multicolumn{1}{l|}{-} & \multicolumn{1}{l|}{-} & \multicolumn{1}{l|}{Other} \\ \hline
\citet{SP29} & OB & - & MCQ & Specific & \multicolumn{1}{l|}{Lexical} & \multicolumn{1}{l|}{Single} & Concat & \multicolumn{1}{l|}{-} & \multicolumn{1}{l|}{-} & \multicolumn{1}{l|}{Other} \\ \hline
\end{tabular}
}
\end{table}
\end{landscape}

\chapter{How to Evaluate MHQA Systems?} \label{sec:eval}

Having evaluation metrics suited to a task is crucial for grasping the task's nuances and gauging the performance of the existing or proposed methods accurately. It ensures researchers assess systems effectively and compare results meaningfully. Future research should benefit clear metrics guide development, revealing where improvements are needed and facilitating advancements in techniques tailored to the intricacies of the task. In this chapter, we look at some commonly used evaluation metrics and potential limitations of using them. We follow our discussion with various sophisticated and clever experiments conducted by the community which shed further light on the intricacies of how the existing systems perform and suggest some directions where further research and development is required.
\section{Evaluation Metrics}
Diversity in multi-hop QA tasks and datasets engenders the need for different evaluation metrics. The general trend in multi-hop QA methods is to split the task into a retrieval (IR) component that finds the relevant contexts, and a reading (RC) component that produces the answer. Therefore, it makes sense to evaluate the two together as well as separately. We describe the methods used for the two evaluations below:
\subsection{\textbf{Retrieval Evaluation}} 
As mentioned in Chapter \ref{sec:method}, retrieval is often broken down into two steps. Different metrics can be used to evaluate the two steps:
\paragraph{Preliminary retrieval:} \citet{SP37} propose three evaluation metrics: \textbf{P EM} (Paragraph exact match) that measures the ability of the retriever to retrieve all the gold paragraphs in the reasoning chain; \textbf{PR} (Paragraph recall) that computes the recall of gold paragraphs among the retrieved paragraphs; and \textbf{AR} (answer recall) that checks if any of the retrieved paragraph contains the answer. \citet{SP3} propose \textbf{acc@$k$} which measures the fraction of cases where the model was able to retrieve \textit{all} the supporting facts within top $k$ retrieved documents. \citet{SP11} define the per-hop retrieval evaluation that treats each hop of retrieval independently. First hop retrieval performance is measured by fraction of cases where the first gold context is retrieved in the first hop of retrieval. For the second hop, the gold paragraph is added to the set of contexts retrieved in the first hop and the ability to retrieve the second hop gold paragraph is evaluated. Since the paper only deals with 2-hop questions, the definition is originally limited to 2 hops. However, we note that this might be extended to $n$ hops where the $n^{th}$ retrieval step is evaluated by the ability to retrieve the $n^{th}$ gold paragraph $p_n$ given $\cup_{i=1}^{n-1} (\{p_i\}\cup R_i)$ where $R_i$ is the set of contexts retrieved during $i^{th}$ hop.
\paragraph{Re-ranking:} \citet{AP6} require the models to classify or rank several reasoning chains as valid explanations of the answer. Therefore, they use \textbf{AUC-ROC} \citep{Melo2013} (area under the Receiver Operating Characteristics (ROC) curve) and \textbf{F1} scores for classification, and \textbf{P@1} and \textbf{Normalized Discounted Cumulative Gain (NDCG)} \citep{10.1145/582415.582418} for ranking. P@1 measures the fraction of cases where the top ranked chain is valid, whereas NDCG is a commonly used metric for evaluating rankings. \textbf{MRR} \citep{10.1145/582415.582418} (Mean Reciprocal Rank) is another ranking based metric used for evaluating ranking of reasoning chains \citep{AP1, DP3}. \citet{SP3} also use Mean Average Precision (MAP) \citep{liu2009mean} that considers the relative position of the relevant document in the ranked list \citep{kadlec-etal-2017-knowledge}.

\subsection{\textbf{Answer Evaluation}}
Based on the type of the answer, the following evaluation metrics have been used for large scale evaluation for the task:
\paragraph{Multi-choice questions:} MCQ questions are straight-forward to evaluate when there is a single correct answer, and classification accuracy over the answer choices is used. To deal with multiple correct answer candidates, \citet{DP5} propose \textbf{F1a} and \textbf{F1m}. Precision and Recall are computed by evaluating each predicted answer candidate. Macro harmonic mean of average precision and average recall is the F1m score. F1a uses micro harmonic mean of precision and recall values for all answer candidates.
\paragraph{Span based answers:} The most commonly adopted metrics are \textbf{Exact Match (EM)} and \textbf{F1} scores on the predicted string tokens. \citet{AP8} argue that EM can often be too strict and propose an alternative \textbf{Partial Match (PM)} where a predicted answer $a_p$ is said to be a partial match with the ground truth answer $a_g$ if either (a) $F1(a_p, a_g) > 0.8$ or (b) $F1(a_p, a_g) > 0.6$ and one of $a_p, a_g$ is a substring of the other. These evaluation metrics are known to work well when the answer spans are small ($<10$ tokens). 
\paragraph{Auxiliary task evaluation: supporting facts:} \citet{DP1} propose to evaluate the reasoning chains by reporting EM and F1 on the supporting facts (note that this is different from reasoning chain as supporting facts are sentences and the contexts are passages). They also propose \textbf{ Joint-EM} and \textbf{ Joint-F1} where the precision is defined as $P_{joint} = P_{ans}\cdot P_{sup}$ and the recall as $R_{joint} = R_{ans}\cdot R_{sup}$.
\citet{SP25} compute the score of a reasoning path in the entity graph by multiplying the corresponding soft masks and attention scores along the path and selecting the top-k scoring paths. If any entity in a supporting fact is reached by any of the $k$ paths, that fact is said to be a hit. Entity-level Supporting fact Prediction (ESP) scores are reported as Exact Match (EM) and Recall values over these supporting facts.
\paragraph{Generative answers:} For longer sequences of texts, directly matching strings to give a binary score fails to tell which answers are closer to the gold answers. Thus, natural language generation (NLG) evaluation metrics are required. \citet{DP3} propose using \textbf{Bleu-1, Bleu-4, Meteor} \citep{papineni2002bleu, banerjee-lavie-2005-meteor} and \textbf{ROUGE-L} \citep{lin-2004-rouge} to evaluate predictions on their dataset. \citet{SP27} also use \textbf{CIDer} \citep{vedantam2014cider} for evaluating long answers on NarrativeQA which emphasizes on annotator consensus.

The above discussed evaluation metrics serve the purpose of providing representative scores for large scale evaluation of methods and facilitating comparison among different techniques. However, multi-hop QA is a complex task and further evaluation experiments designed particularly for the task might be needed to capture nuances of the model. We discuss these in the next section.

\section{Adversarial Evaluation}
Adversarial evaluation is a commonly used evaluation technique for various types of models. It involves testing the models against intentionally crafted difficult examples to assess their robustness and expose potential weaknesses or vulnerabilities. 
\par 
As indicated in Section \ref{subsec:data_create}, the particular choice of available contexts ($C$) for a question in the dataset can lead to `reasoning shortcuts' where a model can correctly answer the question by using only a single context. To avoid such shortcuts in the HotpotQA's distractor setting, the authors used TF-IDF for retrieving confusing contexts. \citet{SP14} collect a different set of distractor paragraphs for the HotpotQA dataset, to evaluate if the models are robust to this change. Same strategy as \citet{DP1} is used while making sure that there is no overlapping distractor paragraph with the original set. An adversarial set of comparison questions is also created by altering the original question so that the correct answer is inverted (for instance, replacing 'which is higher' by 'which is lower').
\par
\citet{SP32} use a clever technique to add fake distractors that can fool a model which uses single hop reasoning shortcuts to answer the questions. A word in the final answer is replaced by another word having a similar GloVe embedding to create a fake answer. For instance, `Mumbai' is replaced by `Delhi'. All occurrences of the word are replaced in the answer passage to get a confusing distractor passage. Since the bridge entity is mentioned in the title of the answer passage, all mentions of a word from the title are also replaced with a similar entity. This is done to break the connection between the fake distractor and the first gold context. This ensures that there is only a single reasoning chain and that a model cannot answer the question by only looking at the answer's paragraph. Evaluation using the adversarial distractors shows a significant drop in the accuracy of a baseline model. Moreover, training using adversarial distractors leads to better performance on the original distractors as well. Therefore, more confusing distractors would lead to better training as well as testing of the MHQA models.

\section{Verifying the Extent of Multi-Hop Reasoning}

Despite the improved scores of models indicated by multiple evaluation metrics on various datasets, it remains doubtful whether the models are actually performing the multi-hop reasoning and following the expected reasoning path for reaching the correct answer. Therefore, different evaluation techniques and modifications of existing datasets are proposed as benchmarks for testing the multi-hop reasoning capabilities of a model. In this section, we list various experiments and benchmarks proposed for the same along with their outcomes and conclusions:
\begin{itemize}
    \item In order to evaluate the interpretability of a model, \citet{SP1} define the \textit{Logical rigor} of a model as Joint EM/Ans EM. Intuitively, it tries to measure among the questions that were answered correctly, what fraction also had correct supporting facts prediction. Surprisingly, baselines have scores of only 30.3\% and 7.9\%. 

    \item \citet{AP2} modify and further annotate HotpotQA to provide three settings, where the models are provided (1) only the passage containing the answer, (2) both supporting passages in random order and (3) both supporting passages in the order of their occurrence in the reasoning chain, with the intuition that a model that employs multi-step reasoning to answer multi-hop questions should benefit from the supporting passages whereas a model that tries to guess the answer directly would instead be confused by the extra information given. Two common techniques, BERT and HotpotReader were tested after employing both query-reformulating and co-matching approaches (see Section \ref{sec:tax} on taxonomy). It was observed that the models could gain very little performance ($ \sim 1\% $ and $4\% $ accuracy with query reformulation and co-matching respectively) by using the reasoning chains provided. This highlights the inability of the existing techniques to incorporate multi-hop reasoning to perform MHQA. Further, it is found that BERT and co-matching show slightly higher improvements than their respective counter-parts.

    \item \citet{AP8} use BERT and DecompRC \citep{SP14} to generate single hop sub-questions comprising the 2-hop questions in the HotpotQA dataset and the answers to these questions. The claim is that if a model employs multi-hop reasoning to answer a question, it should trivially be able to answer the individual sub-questions. Surprisingly, for $\sim 50-60\% $ of the questions correctly answered, at least 1 of their corresponding sub-questions could not be answered correctly. Further, of the questions where both the sub-questions were answered correctly, $\sim 10\% $ were incorrectly answered. This indicates that the models tend to jump directly to the answer instead of breaking down the questions into simpler questions. 
    \item \citet{AP6} propose three modifications of the QASC dataset that require the model to explicitly predict the reasoning chains along with the final answers (explainable MHQA). i) eQASC: For each question in QASC, up to 10 candidate reasoning chains are automatically generated and each candidate chain is annotated to be valid (if the chain can imply the answer) or invalid (otherwise). ii) eQASC-perturbed: In the candidate chains of QASC, one word/phrase that is likely to be a bridge entity among two facts, is replaced by a similar meaning word ensuring that the chain remains to be valid. This is done by crowd sourcing where workers were asked to replace one occurrence of the word that appears in different sentences of a candidate chain. iii) eOBQA: a small number of questions in OpenBookQA are used for generating candidate reasoning chains using sentences present in QASC and are annotated by crowd-sourcing. This is done to test the generalization of the model on an unseen dataset.
    \item \citet{AP5} use the term Disconnected Reasoning (DiRe) for when the model is able to arrive at the correct answer using (possibly multiple independent) incomplete reasoning chains. To measure disconnected reasoning, a DiRe probe is created that checks if the output of $h(q, C\setminus\{p_1\})$ and $h(q,C\setminus\{p_2\})$ (refer to Chapter \ref{sec:prob_def} for notations) can be trivially combined to answer $q, C$ (where '$\setminus$' denotes set difference). To discourage disconnected reasoning, the dataset is modified to include negative samples where the given $C$ is not sufficient to answer the question i.e., $C \cap P_q \neq \phi$ and the model is required to identify these questions as unanswerable. When running the DiRe probe, disconnected reasoning is found to be reduced significantly after training using this modification. 
    \item \citet{AP7} argue that requiring the model to only output the supporting facts might not be enough to ensure the explainability of the model and the model should be required to also output the derivation steps. A derivation step is formalized as a triplet of the form $\langle d^h, d^r, d^t \rangle, \text{ where }d^h, d^t$ are entities (noun phrases), and $d^r$ is a verb phrase representing a relationship between the two entities. A small subset of the HotpotQA dataset is annotated by crowd-sourcing and released. Evaluating models on this set indicates the scope of improvement on this benchmark.

\end{itemize}

Results of these works are significant as they suggest that improved accuracy on existing datasets may not correlate well with the models' ability to perform multi-hop reasoning. Furthermore, they highlight the inefficacy of existing models to perform multi-hop reasoning as well as the inefficacy of the datasets to evaluate the same. This implies a need for more carefully created datasets and challenging benchmarks that do not allow the models to score well without accurately following the required reasoning paths. Additionally, it is encouraged to formulate better and more such tests/probes that check and prevent the models from using loop-holes instead of doing multi-hop reasoning. Above all, it is fair to say that the task of MHQA is far from solved.

% \citet{SP18} insight: It has been show that $\leq$ 3-hops can cover most real-world cases, such KB reasoning (Xiong et al., 2017; Das et al., 2018) 
% Include GRC from eQASC somewhere.

\chapter{Multi-Hop Question Generation} \label{sec:mhqg}

The task of Multi-Hop Question Generation (MHQG) is very closely related to MHQA and shares some of the required reasoning and natural language understanding abilities. The task has many applications and has also received a growing attention in the recent years. Therefore, we add a brief discussion of MHQG in this chapter.

The goal is to generate a multi-hop question given a set of contexts and optionally, an answer. In chapter \ref{sec:data}, we already discussed the challenges of question generation while creating MHQA datasets. A lot of those challenges directly apply to the task of MHQG as well.
MHQG has widespread applications in multiple domains including education, where generating questions that require multiple steps of reasoning can be very useful for inspiring critical thinking in students \citep{qg4online}. QG also has a direct application for chat-bots e.g., in initiating conversations, asking and providing detailed information to the user by considering multiple sources of information. MHQG will enhance the ability of these chat-bots to ask useful questions \citep{teachingqg}. It can also combine with question answering (QA) models as dual tasks to boost QA systems with reasoning ability \citep{qaqgdual}.
\par
The task of traditional question generation (QG) has gained a lot of interest recently \citep{qg4rc, maxout-qg, ansagnosticQG}. However, MHQG is a more challenging task than simple QG. It requires the model to first identify scattered pieces of information that can be aptly combined to form a valid reasoning path from the answer to the question, and then reason over these pieces of information to generate a factual and coherent question. 

\section{Datasets}

The datasets for MHQA can also be used to train and evaluate MHQG models by modifying the input and the required output of the model \citep{QG2, QG3}. Since HotpotQA has annotated supporting facts, it is able to provide stronger training supervision and hence, it is the most commonly used dataset for MHQG. \citet{QG1} use the DecompRC model \citep{SP14} to decompose each question in HotpotQA into two sub questions and fine tune a  GPT2-small model to rewrite the first question into the second. \citet{QG6} use HotpotQA as the labelled dataset and ComplexWebQuestions \citep{complexwebques} and DROP \citep{dua-etal-2019-drop} as large corpora for multi-hop questions.

\section{Evaluation}

Language generation metrics such as BLEU (BLEU1-4), ROUGE-L, METEOR are usually adopted for MHQG \citep{QG1, QG2, QG3, QG4, QG6}. QBLEU4 \citep{qbleu}, a QG metric which was shown to correlate significantly better with human judgements, is also used for evaluating MHQG \citep{QG1, QG2}. The task also often requires human evaluation of fluency, semantics, answerability etc of the generated questions. \citet{QG4} also use GLEU \citep{gleu} for their experiments. Another method of evaluating MHQG is to measure the gain in performance of SOTA MHQA models when trained with data augmentation using the generated questions \citep{QG1,QG5}.

\section{Methods}

\citet{QG1} tackle the problem of difficulty controllable question generation (DQG)\citep{dqg} by generating questions which require a particular number of reasoning hops. The assumption is that the difficulty of a question directly correlates to the number of inference steps required to answer it. The first step of the proposed algorithm builds a context graph in the same way as \citep{localKGS2Smulti}. All sentences from the context are converted into triplets of the form $\{ s(subject), r(relation), o(object) \}$ and a relational edge of type $r$ is added from $s$ to $o$. Co-reference resolution is used to merge the nodes referring to the same entity.
Next, a node $N_0$ is sampled as the final answer and with $N_0$ as the root, a maximum spanning tree is extracted. For generating a question with difficulty (same as the number of hops) $=d$, the tree is pruned to have $d+1$ nodes. A GPT2-small model is fine-tuned on HotpotQA and used to generate an initial question $q_0$ using $N_0$, $N_1$ and the context sentence connecting the two nodes $S_1$. Another GPT-2 model is used to rewrite the question iteratively and successively increase the difficulty. In simpler words, for generating a question $q_d$ with difficulty $d$, the re-writer model is run on $q_0$ for $d$ iterations.
While the model performs well for up-to $3$ hops, the input to the re-writer model becomes too large for subsequent hops and results in questions with poor quality.
\par
\citet{QG2} leverage a graph constructed similar to \citet{SP25} to generate multi-hop questions. Pre-trained GloVe embeddings and answer tagging embeddings \citep{qg4online} are passed to two Bi-LSTM layers to get the initial contextual representations. An attention layer and another Bi-LSTM layer is used to get the answer aware context embeddings. An answer aware sub-graph is computed by masking the entities that are irrelevant to the answer and Graph Attention Network is applied to this sub-graph. The context encodings across the hops are combined via a gated fusion module. The Answer embeddings are updated using bi-attention and The Maxout Pointer \citep{maxout-qg} framework is used on top of a Uni-directional LSTM decoder to generate the question. An additional BFS loss proposed by \citep{SP25} is found to improve the performance.
\par
\citet{QG3} exploit the presence of supporting facts in HotpotQA while training by adopting an RL reward of the auxiliary task of supporting facts prediction (SFP). 
Similar to \citet{QG2}, answer tagging features are concatenated with the document word embeddings and fed to a Bi-LSTM encoder. The output of the encoder is shared by the MHQG and supporting fact prediction (SFP) models. The SFP model is a binary classifier trained on HotpotQA to output the probability of each sentence being a supporting fact. The F1 score between the predicted and ground truth supporting facts is added as a reward. The REINFORCE algorithm \citep{reinforce} is used with the self-critical sequence training \citep{scst} framework to avoid the high variance. In order to make the training more stable, a weight history similar to \citet{scst} is added. The output probabilities of the SFP model are also used to update the answer aware encodings by another Bi-LSTM model. For generating the question, a LSTM decoder with global attention mechanism \citep{attention4nmt} is used along with the copy mechanism \citep{pgn, pgn2}.
\par
\citet{QG4} argue that using standard transformers instead of Graph networks should be enough to reason about the relations between entities for forming multi-hop questions. A transformer is extended with sentence id embeddings and answer token indicator embeddings and trained with an additional contrastive loss as regularization. The contrastive learning setup assumes supporting fact sentences as positive samples and others as negative samples and a binary classifier consisting of a MLP. A significant mismatch in the distribution of question length over train and dev set of HotpotQA is observed and mitigated by filtering out all the questions that are more than 30 words long. Most of the pruned questions are from the train-easy subset. Both data filtering and contrastive training are found to boost the performance significantly.

A novel graph-augmented transformer encoder (GATE) is proposed which has two additional layers than the standard transformer encoder (TE): a) Graph-attention sub-layer computes the similarity scores for attention using only the nodes that are connected in a dynamically created graph. The graph is a multi-relational graph with three types of nodes: named-entity mentions, co-referent-entities and sentence-ids. b) Fused attention sub-layer which uses a MLP with ReLU activation to aggregate the graph-attention embeddings and the TE embeddings. Experiments show that the added layers alone do not improve the performance significantly whereas an ensemble of TE and GATE does.

\citet{QG5} propose a question generation technique for the task of unsupervised MHQA. Following HotpotQA, their method generates two types of questions:

\textit{Bridge questions:} Given two contexts as inputs, all entities common to the two contexts are considered as bridge entities. A Google T5 model \citep{t5} is fine-tuned on SQuAD to generate two single hop questions using the answer entity and the bridge entity respectively. The latter question is converted to a declarative form, $s$ following \citet{qa2nli}. The bridge entity in the bridge entity question is replaced with "The [MASK] that \{s\}" and BERT-Large is used to fill the [MASK].

\textit{Comparison questions:} Entities with NER types Nationality, Location, DateTime and Number are treated as potential comparative properties. Two single hop questions are generated on two entities of the same NER type and a pre-defined template is used to combine these into a multi-hop comparison question.
\\
For generating questions using a table, a GPT-TabGen model \citep{nlg-tablegen} is used to generate sentences that describe a given entity using information from the given table. These sentences are then used for generating either \textit{bridge} or \textit{comparison} type questions. We refer the reader to Figure 4 of the original paper \citep{QG5} for a better understanding of the different types of generated questions.
A pre-trained GPT-2 model is used to filter questions that are unnatural or dis-fluent. A BART model \citep{bart} is also used to paraphrase the generated questions. The generated questions are then used to train the model resulting in a zero shot algorithm.

\citet{QG6} aim to tackle MHQG in a low resource setting. Specifically, the model uses a small amount of labelled data $D_L$, in the form of $(context, answer, question)$ triplets, and a large set of multi-hop questions $D_U$. The idea is to first learn the semantics of multi-hop questions by training a neural hidden semi-Markov model \citep{semi-markov} on the unlabelled data $D_U$. The model uses two latent variables for parameterizing the similar segments in questions from $D_U$: a) a state variable $z_t$, indicating which segment the $t^{th}$ term belongs to, and b) a length variable $l_t$, specifying the length of the current segment. The term probabilities are computed using a GRU decoder followed by an attention layer.

The patterns learned in the first step are used as priors for the QG model for regularization. Prior is estimated by sampling a sequence of states $z_t$ having length $l_t$. The reasoning chain extraction is similar to as described in \citet{QG2}. For encoding the textual input, BERT embeddings are passed to a bi-GRU followed by an attention layer. The decoder is another GRU with a copy mechanism \citep{gu-incorporating-copy-s2s} which is regularized to fit the prior pattern. The training loss is the weighted sum of the cross entropy loss and RL policy gradient \citep{adversarial-dialogue}. The reward function evaluates a) fluency (following \citet{sentence-deeprl}), b) Answerability (using QBLEU4), and c) Semantics (using WMD). 

\paragraph{Conclusion:}
The task of multi-hop question generation has gained some attention of the community and has advanced at a rapid pace. Solving MHQA would go a long way in solving MHQA since we can use MHQG models to generate high quality large scale datasets for the development of powerful MHQA models. Further, by dynamically constructing intricate questions that traverse multiple pieces of information, we pave the way for more nuanced understanding and exploration of complex knowledge domains, ultimately driving the evolution of AI-driven information retrieval and comprehension.
\chapter{Future of MHQA} \label{sec:future}

Multi-hop QA has been researched quite extensively in the recent years with multiple diverse models proposed that aim to model the multi-step retrieval-reasoning process and achieve promising improvements on existing datasets and benchmarks. Such systems capable of performing multi-step reasoning have a variety of applications ranging from chat-bot assistants that are capable of interactive conversations, to search engines that are capable to retrieve results that may be relevant but not reachable directly from the query text. At the same time the task of MHQA is significantly more challenging than its single hop counterpart. Since paragraphs multiple hops away from the question could share few common words and little semantic relation with the question \citep{SP1}, the task to retrieve such contexts is challenging and suffers from semantic drift. The ability of pre-LLM models to combine multiple contexts for reasoning is also limited. Further challenges for solving MHQA is the difficult process of creating datasets that require the models to perform multi-hop reasoning, as well as the task of evaluating the models' abilities to do so without any hacks. Some challenging benchmarks and evaluation methods have been recently proposed that bring out some surprising and interesting observations. These results point out to several limitations of existing systems and call for further research.
\par
Below, we list and discuss some of these directions for future research (including ones originating from the currently recognized shortcomings) which we believe could be promising to be explored. 

\section{Flexible Any-Hop Models}
As discussed in Section \ref{sec:tax}, majority of the existing methods for MHQA are either limited to two hops or require the number of hops to be a hyper-parameter. Since natural questions can require any number of reasoning hops, this attribute of existing models is artificially limiting. QA systems should be flexible and robust to the number of hops in a question in order to be practically usable. In order to do so, methods following the type III  and IV in Figure \ref{fig:model-types} should be explored with a greater interest since feedback from the answering module can serve as a useful stopping criteria.

\section{Explainable Multi-Hop QA}

Despite the impressive gains in performance on various multi-hop datasets, it is not evident whether the models are performing the multi-step reasoning or just guessing the answers. Therefore, following the release of HotpotQA, a large number of works have focused on explainable MHQA. Apart from the standard evaluation metrics, various evaluation methods and dataset benchmarks have been proposed to test the explainability of the models which have revealed several significant results. Further such benchmarks and evaluation strategies are encouraged to measure and reflect the true progress of models in performing multi-hop reasoning. While LLM prompting strategies such as CoT prompting make the task naturally explainable, LLMs suffer from hallucinations which lowers their trust for such applications.

\section{Better Datasets}

Many works have highlighted the limitations of existing MHQA datasets. \citet{AP3} show experimentally that multi-choice questions are easier to hack by models, regardless of number of answer candidates being small or very large. Similarly, \citet{SP14} show that the distractor setting, even with as many as 500 distractors, is easier to hack for single-hop QA models. \citet{DP1} and \citet{SP16} argue that datasets that are created using KBs suffer from lack of diversity in question and answer types. Following these observations, it is encouraged that the future datasets are in the open-domain setting, have questions with either span-based or generative answers, and do not rely completely on the structure of existing KBs.
\par
\citet{SP3}, \citet{SP5}, \citet{SP14} and \citet{AP4} find that a significant portion of questions in the existing datasets are single-hop solvable due to a variety of reasons. One of these reasons is that the source of the questions is same as the set of contexts. Therefore datasets like \citep{DP2, DP3, archival, newsqa, quac} that use separate sources for generating and answering questions are encouraged. However, attention needs to be given to the cases where there is a discrepancy between what is mentioned in the two sources \citep{SP6}.
\par
Most of the existing works have focused on datasets with MCQ or questions with span answers and more focus on the more challenging problem of generative MHQA is desirable.

\section{Better Evaluation Metrics}

As discussed in Section \ref{sec:eval}, a variety of evaluation metrics have been used for evaluating MHQA models. However, existing metrics face some challenges and might not be sufficient for evaluating MHQA. Since MHQA is a more complex task compared to single-hop QA, more metrics specific to MHQA are encouraged. One promising direction is to perform per-hop evaluation and accumulate the per-hop scores to get a final score. This kind of evaluation would require the models to be explainable as well as interpretable.
\par
Some challenges are common to both single-hop and multi-hop QA evaluation metrics. For instance, while evaluating span based answers, metrics based on lexical matching would mark \textit{U.S.} as incorrect when the gold answer is \textit{United States} \citep{SP6}. Therefore, evaluation metrics should be able to deal with synonyms when matching the answers. Another issue might be a metric giving a score of 0 to the answer \textit{U.K.} when the gold answer is \textit{ London}. These cases are frequent in datasets like WikiHop where the sources of question and the contexts are different \citep{SP15} and might have answers mentioned with different granularity. Therefore, it might be useful to give some partial score for the answer being geographically close or for the answer having a coarser granularity. Similarly, answering \textit{January, 1989} should earn some score if the gold answer is \textit{December, 1988} due to the predicted answer being temporally close to the gold answer. Some partial score can also be rewarded to answers having a coarser granularity (\textit{December, 1988} vs \textit{1988}). On similar lines, using hypernym relations from ConceptNet or WordNet \citep{10.1145/219717.219748} for evaluating the answer can be a possible direction.
\par
Following a similar reasoning, the evaluation should match the semantics of the answer rather than the lexical overlap. Therefore, evaluation metrics like word mover similarity \citep{kusner2015word} or sentence mover similarity \citep{clark2019sentence} that perform soft matching over embeddings might be a promising direction. Since the evaluation of language generation tasks are widely known to have a scope of improvement, evaluation of generative MHQA 
% as well as MHQG 
is also an open problem and new evaluation techniques are encouraged.

\section{Methods to Incorporate Commonsense}

A hop in multi-hop reasoning can be performed using some retrieved context, where the context may either be retrieved from the corpus or from the commonsense knowledge. \citet{SP6} find that $16\%$ of the failures of their model were caused by missing commonsense background knowledge. \citet{SP27} propose a novel method for incorporating commonsense for MHQA that shows impressive results. More techniques for exploiting the rich commonsense knowledge bases to perform multi-step reasoning can be a promising direction to explore.

\section{Arithmetic Questions}

The inability of QA systems to perform arithmetic operations is well known \citep{patel2021nlp, hendrycks2021measuring, schubotz2018introducing} and this inability is exacerbated in the multi-hop setting. \citet{AP4} observe that $45\%$ of the comparison questions in HotpotQA are numerical questions. \citet{SP25, SP6, SP14} find that their model is unable to give a correct answer when the query is a comparison between two dates ``February 20, 1959'' and ``February 10, 1967''. Arithmetic calculation may also be required for non-comparison type questions. For example, answering ``Who was the president of USA in 1994'' from the context ``Bill Clinton: 1993-2001'' requires some arithmetic computation. \citet{wangIRJ21} approached this kind of temporal problems by computing the overlap of content time expressions that occur in text with the computed question's time scope using kernel density estimate. Another example is \citep{wangSIGIR21} that answers ``when'' type questions by predicting the event dates based on the analysis of multi-variate time series derived from underlying news collection. Even powerful LLMs have been shown to perform poorly with arithmetic reasoning \citep{llm-arith-3, llm-arith-4} and multiple methods are being proposed for the same \citep{llm-arith-1, llm-arith-2, llm-arith-4, llm-arith-3}

Multi-hop QA systems that are capable of solving arithmetic comparisons and computations would greatly enhance accuracy of MHQA.
\par
Similarly, it is observed that some particular types of questions (temporal, geographical, count) are more challenging than others \citep{SP1, SP8, SP14, SP15}. Figure 3 in \citet{SP8} shows the complexity and model performance on different types of questions in HotpotQA. Targeting the more challenging questions specifically could also lead to better MHQA systems.

\section{Better Incorporation of Powerful LLMs}

LLMs have been widely adopted due to their performance exceeding expectations for most tasks. Existing works employ LLMs for some components of the task to boost performance. The MHQA community would benefit from keeping up with the rapid developments of LLMs, incorporating the advancements in the models' abilities and efficiency. Consequently, developing more robust and powerful LLMs suited to multi-step reasoning would greatly help the development of MHQA systems

\section{Conclusion}
As we conclude, acknowledging both the strengths and weaknesses of existing data, models, and evaluation methods in multi-hop QA provides a solid foundation for charting future research paths. By leveraging insights gained from these limitations, we can propel advancements towards more robust, adaptable, and comprehensive question answering systems, shaping the landscape of AI-driven knowledge exploration for years to come.

% \begin{acknowledgements}
% We thank the editors and board members of the Foundations and Trends in Information Retrieval for their patience and assistance with preparation of our book.
% \end{acknowledgements}

\appendix

\chapter{Background} \label{app:background}

\section{BM25}

BM25 is a ranking function used to retrieve documents given a search query.
BM25 stands stands for \textit{Best Match 25}\footnote{BM25 is also known as Okapi BM25, which was used first by the Okapi information retrieval system implemented by London's City University
(\url{https://en.wikipedia.org/wiki/Okapi_BM25)}.}. 
It uses a bag-of-words mechanism to score proximity between the search query and the documents. 
Given a query 
$Q = {q_1, q_2, ... , q_n}$,
where $q_i$ denotes a keyword in the query 
$Q$,
the BM25 score of the document $D$ is defined as follows - 
\begin{equation}
    BM25(D, Q) = \sum^{n}_{i=1}IDF(q_i).\frac{freq(q_i, D).(k_1 + 1)}{freq(q_i, D) + k_1. (1 - b + b. \frac{|D|}{avg.doc.len.})}
\end{equation}
where $freq(q_{i}, D)$ is the number of times $q_i$ occurs in $D$, $|.|$ denotes the number of words in D, $avg.doc.len.$ denotes the average number of words in the document, $k_1$ and $b$ are free parameters\footnote{Typically $k_1 \in [1.2, 2.0]$ and $b=0.75$.}, and $IDF(q_i)$ denotes the inverse document frequency weight of query term usually computed as follows - 
\begin{equation}
    IDF(q_i) = ln(\frac{N-n(q_i)+0.5}{n(q_i)+0.5}) + 1
\end{equation}
where $N$ is the total number of documents in the collection, and $n(q_i)$ is the number of documents containing $q_i$.

Even though the technique was devised in 1970s-80s, BM25 and its variations are still widely adopted for document retrieval, especially when the document corpus is very large and using \textit{dense retrievers}\footnote{\textit{Dense retriever} is a general umbrella term used to refer to the neural network based retrieval systems.} has a big computational overhead. 

\section{Recurrent Neural Networks}
Recurrent Neural Networks (RNNs) are a class of artificial neural networks that have loop connections that allow information propagation across time through the same neurons. Prior to transformer networks \citep{vaswani2017attention}, RNNs were the most popular framework class to process sequential information, and are still widely adopted in real-world systems. Most practical RNN-based architectures have additional stored states that allow the vanilla RNN architecture to overcome its shortcoming of short-term memory loss. Gated recurrent units (GRU) cells \citep{cho2014learning} and long short term memory (LSTM) cells \citep{hochreiter1997long} are two of the most popular stateful RNN cells that use gated mechanism to handle long term memory. \citep{see2017get} proposed a pointer generator network to overcome the over-repetition of RNN generated output using coverage loss. We point the readers to the comprehensive survey of recurrent neural networks by \citep{lipton2015critical} for extensive explanation on the topic.

\section{Transformers for Language Modeling}
Even the advanced RNN models like LSTMs and GRUs have a tough time dealing with long sequences. \citet{attention4nmt} introduced the attention mechanism which allows the model to focus on certain parts of the input when predicting a particular output token. Doing so significantly helps with tasks like machine translation where certain words of the input sequence are directly related to a word in the output sequence. Many forms of attention have since been used effectively for various tasks.
\par
\citet{transformer} extended the idea of attention by removing the recurrent component of the model altogether and proposed the transformer model where both the encoder and the decoder consist of several self-attention and feed forward layers. The transformer model also introduced the multi-head attention. These components allows for very large models which can have a lot more parameters without comprising on the performance. Transformers are also proved to be very versatile, having great success in a large number of natural language applications. 
\par
While the original transformers model was trained using the next-token prediction task implying the unidirectionality of the encoder model, BERT \citep{bert} was a bidirectional encoder based transformer which was trained using the masked language modeling task. BERT has proved to be a versatile model and the word representations learned using BERT have been used as embeddings for almost all natural language tasks. 
\par
Success of transformer models including BERT led to their use as large pre-training models and several models like AlBERT \citep{albert}, RoBERTa \citep{roberta} and GPT were proposed. AlBERT uses parameter reduction techniques which allow for smaller and faster training of the BERT models while achieving a similar level of accuracy as BERT. RoBERTa is a much more robustly optimized version of BERT, trained with optimized design and hyperparameters choices, which could significantly outperform the originally trained BERT model.
\par
Pre-training of large language models (LLMs) has become increasingly popular leading to larger and larger models trained on huge corpora of natural language. The different versions of the model follow the same principle, with GPT-1 having 117 million parameters and GPT-4 having about a 100 trillion parameters. GPTs are trained on huge corpora using the next token prediction task.
An extensively detailed explanation of different architectures and training techniques for transformer based models is neither feasible nor in the scope for this work. Therefore, we point the readers to the comprehensive survey of transformers by \citep{LIN2022111} for further details on the topic.

\section{Graph Neural Networks}
Graphs are a very simple and versatile method of representing data and its inherent structure. Neural Networks could be adapted to incorporate this structure leading to Graph Neural Networks (GNNs). GNNs can be adopted for various different types of data and tasks, leading to several improvements increasing their capabilities. The integral part of all these models is the message passing algorithm briefly explained below.
\par
Given a graph $G = (V, E)$ having $n=\vert V \vert $ nodes, the representation of each node is updated following the given steps:
\begin{itemize}
    \item \textbf{Initialization:} The representation of every node $v$ is initialized as $h_v^0 = X_v$, where $X_v$ is the feature vector.
    \item \textbf{Update:} For each layer $i$, the representations of each node $v$ is updated as:
    \begin{equation}
        h_v^i = \sigma_{u\epsilon N(v)}(W_i \Sigma\frac{h_u^{i-1}}{N(v)} + U_ih_v^{i-1})
    \end{equation}
    where $\sigma$ is the activation function, $W_i$ and $U_i$ are the weight matrices corresponding to the layer $i$ and $N(v)$ is the set of neighbouring nodes of the node $v$.
    \item \textbf{Prediction:} The representations after layer $K$ are passed to a linear network for the eventual prediction task.
\end{itemize}
At every layer, the representation of node $v$ is updated with an activation applied to the weighted average of representations of the nodes directly connected to $v$. Therefore, after $k$ layers, the node $v$ is supposed to receive the `message' from all nodes having a path to $v$ of length $\leq k$. The weighted average also ensures that the nodes that are closer to $v$ in the graph end up affecting its representation more.
\par
A layer of a Graph Convolutional Network (GCN) \citep{gcn} consists of a GNN layer followed by a Linear layer. Relation GCN (R-GCN) \citep{schlichtkrull2017modeling} allow for different kinds of edges by having different weight matrices for nodes connected to $v$ via different kind of edges. Graph Attention Networks (GAN) \citep{gat} incorporate self attention into GNNs by using the attention weights while performing the message passing algorithm. Several other modifications of GNNs are proposed for different tasks.
\par
We point the readers to the comprehensive survey of graph neural networks by \citep{wu2020comprehensive} for further reading on the topic.

\section{Large Language models} \label{sec:llm-back}
\textbf{Language models} refer to a class of self-supervised NLP models that are trained on large unlabeled datasets to learn to predict the likelihood of a word or sequence of words occurring based on the context provided by the preceding words. This ability to estimate the probability of a word given its context forms the foundation of language modeling. These models undergo training on various tasks, such as next-word prediction \citep{gpt}, masked language modeling (the task of predicting randomly missing tokens), and next-sentence prediction \citep{bert}, without the need for labeled data. Due to their reliance on extensive training data, language models develop a strong grasp of underlying language patterns and concepts. Generally, language models are not designed for specific tasks and can be fine-tuned with minimal data for various downstream applications. Extensive research has shown that utilizing large language models (LLMs) pre-trained on vast amounts of data yields impressive results in language understanding and generation tasks \citep{gpt-success-qa, gpt-success-entail, gpt-success-mt, gptsara}. The advent of transformer models has made it possible to train such highly advanced language models, resulting in popular models like BERT, T5, and GPT-3 \citep{bert, t5, gpt}.

\subsection{Generative Pre-trained Transformer (GPT)}
GPT, a series of generative pre-trained large language models \citep{gpt}, is characterized by its decoder-only transformer architecture. Unlike other transformer models that have both encoder and decoder blocks, GPT models consist solely of decoder blocks, eliminating the encoder-decoder cross-attention layer from each block. The different versions of GPT, namely GPT, GPT-2, GPT-3, and GPT-4, vary in terms of model size and training data. For example, GPT-3 has 175 billion model parameters and is trained on a massive corpus of 499 billion tokens, while GPT-2 has 1.5 billion parameters and is trained on a dataset of 10 billion tokens. 

\begin{figure}
    \centering
    \includegraphics[width=0.5\columnwidth]{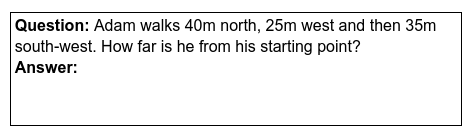}
    \includegraphics[width=0.5\columnwidth]{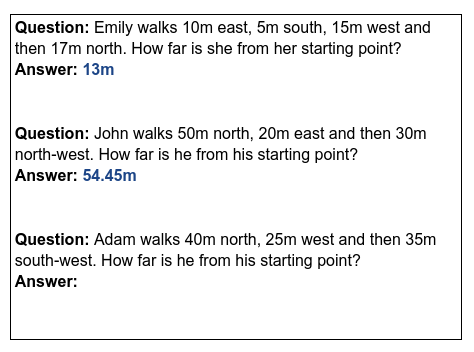}
    \includegraphics[width=0.5\columnwidth]{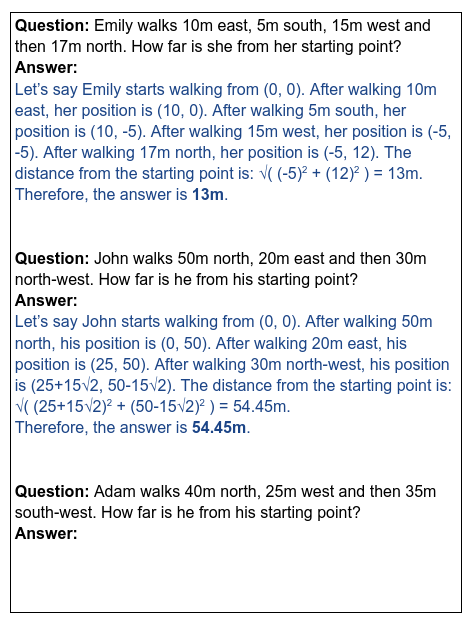}
    \caption{Example of zero-shot (top), few-shot (middle), and CoT (bottom) prompting for the same question. }
    \label{fig:prompt_egs}
\end{figure}

\subsection{Prompting GPT-3}
GPT-3 has achieved remarkable success in various downstream natural language tasks, including question answering \citep{gpt-success-qa}, Machine Translation \citep{gpt-success-mt} and Entailment prediction \citep{gpt-success-entail}, with minimal supervision required. During a typical run of the model, an incomplete piece of text is provided as a `prompt', and the model iteratively generates the most likely tokens to complete the text. This prompting technique has demonstrated impressive performance in the zero-shot setting, where the model is not provided with any in-context examples and is expected to predict the correct output for the given question in the prompt (Figure \ref{fig:prompt_egs}).
\\
On the other hand, few-shot prompting \citep{fewshot} involves including a small number of sample input-output pairs within the prompt as references for the model (Figure \ref{fig:prompt_egs}). The inclusion of a few reference examples provides valuable guidance to the model, allowing it to generate more accurate and relevant responses.
\\
In their work, \citet{cot} introduced the concept of chain-of-thought (CoT) prompting, which goes a step beyond simply providing input-sample output pairs. CoT prompting includes a coherent sequence of reasoning steps that gradually build up to the correct answer. By presenting the model with a step-by-step thought process, CoT prompting offers explicit examples of how to arrive at the correct answer based on the given input facts. This method is particularly valuable for tackling complex tasks that demand multiple layers of reasoning including the task that this study focuses on.
Figure \ref{fig:prompt_egs} shows examples of zero-shot, few-shot, and CoT prompts for an arithmetic question. Here, the prompt consists of 2 in-context examples is 2.
\par
For further background and details, we refer the readers to the comprehensive survey on LLMs by \citet{llm-survey}

%BACKMATTER SEE DOCUMENTATION
\backmatter  % references, restarts sample

\printbibliography

@inproceedings{SP1,
    title = "Cognitive Graph for Multi-Hop Reading Comprehension at Scale",
    author = "Ding, Ming  and
      Zhou, Chang  and
      Chen, Qibin  and
      Yang, Hongxia  and
      Tang, Jie",
    booktitle = "Proceedings of the 57th Annual Meeting of the Association for Computational Linguistics",
    month = jul,
    year = "2019",
    address = "Florence, Italy",
    publisher = "Association for Computational Linguistics",
    url = "https://aclanthology.org/P19-1259",
    doi = "10.18653/v1/P19-1259",
    pages = "2694--2703",
}

@inproceedings{focus-time,
  author       = {Adam Jatowt and
                  Ching{-}man Au Yeung and
                  Katsumi Tanaka},
  editor       = {Qi He and
                  Arun Iyengar and
                  Wolfgang Nejdl and
                  Jian Pei and
                  Rajeev Rastogi},
  title        = {Estimating document focus time},
  booktitle    = {22nd {ACM} International Conference on Information and Knowledge Management,
                  CIKM'13, San Francisco, CA, USA, October 27 - November 1, 2013},
  pages        = {2273--2278},
  publisher    = {{ACM}},
  year         = {2013},
  url          = {https://doi.org/10.1145/2505515.2505655},
  doi          = {10.1145/2505515.2505655},
  bibsource    = {dblp computer science bibliography, https://dblp.org}
}

@book{DBLP:series/synthesis/2021Roy,
  author       = {Rishiraj Saha Roy and
                  Avishek Anand},
  title        = {Question Answering for the Curated Web: Tasks and Methods in {QA}
                  over Knowledge Bases and Text Collections},
  series       = {Synthesis Lectures on Information Concepts, Retrieval, and Services},
  publisher    = {Morgan {\&} Claypool Publishers},
  year         = {2021},
  url          = {https://doi.org/10.2200/S0113ED1V01Y202109ICR076},
  doi          = {10.2200/S0113ED1V01Y202109ICR076},
  isbn         = {978-3-031-79511-4},
  bibsource    = {dblp computer science bibliography, https://dblp.org}
}

@inproceedings{SP2,
    title = "Multi-Hop Paragraph Retrieval for Open-Domain Question Answering",
    author = "Feldman, Yair  and
      El-Yaniv, Ran",
    booktitle = "Proceedings of the 57th Annual Meeting of the Association for Computational Linguistics",
    month = jul,
    year = "2019",
    address = "Florence, Italy",
    publisher = "Association for Computational Linguistics",
    url = "https://aclanthology.org/P19-1222",
    doi = "10.18653/v1/P19-1222",
    pages = "2296--2309",
}

@inproceedings{SP3,
  title={Multi-step entity-centric information retrieval for multi-hop question answering},
  author={Das, Rajarshi and Godbole, Ameya and Kavarthapu, Dilip and Gong, Zhiyu and Singhal, Abhishek and Yu, Mo and Guo, Xiaoxiao and Gao, Tian and Zamani, Hamed and Zaheer, Manzil and others},
  booktitle={Proceedings of the 2nd Workshop on Machine Reading for Question Answering},
  pages={113--118},
  year={2019}
}

@inproceedings{SP4,
    title = "Quick and (not so) Dirty: Unsupervised Selection of Justification Sentences for Multi-hop Question Answering",
    author = "Yadav, Vikas  and
      Bethard, Steven  and
      Surdeanu, Mihai",
    booktitle = "Proceedings of the 2019 Conference on Empirical Methods in Natural Language Processing and the 9th International Joint Conference on Natural Language Processing (EMNLP-IJCNLP)",
    month = nov,
    year = "2019",
    address = "Hong Kong, China",
    publisher = "Association for Computational Linguistics",
    url = "https://aclanthology.org/D19-1260",
    doi = "10.18653/v1/D19-1260",
    pages = "2578--2589",
}

@inproceedings{SP5,
  title={Simple yet Effective Bridge Reasoning for Open-Domain Multi-Hop Question Answering},
  author={Xiong, Wenhan and Yu, Mo and Guo, Xiaoxiao and Wang, Hong and Chang, Shiyu and Campbell, Murray and Wang, William Yang},
  booktitle={Proceedings of the 2nd Workshop on Machine Reading for Question Answering},
  pages={48--52},
  year={2019}
}

@inproceedings{SP6,
  title={Hierarchical Graph Network for Multi-hop Question Answering},
  author={Fang, Yuwei and Sun, Siqi and Gan, Zhe and Pillai, Rohit and Wang, Shuohang and Liu, Jingjing},
  booktitle={Proceedings of the 2020 Conference on Empirical Methods in Natural Language Processing (EMNLP)},
  pages={8823--8838},
  year={2020}
}

@inproceedings{SP7,
  title={Unsupervised Alignment-based Iterative Evidence Retrieval for Multi-hop Question Answering},
  author={Yadav, Vikas and Bethard, Steven and Surdeanu, Mihai},
  booktitle={Proceedings of the 58th Annual Meeting of the Association for Computational Linguistics},
  pages={4514--4525},
  year={2020}
}

@inproceedings{SP8,
  title={Answering Any-hop Open-domain Questions with Iterative Document Reranking},
  author={Zhang, Yuyu and Nie, Ping and Ramamurthy, Arun and Song, Le},
  booktitle={Proceedings of the 44th International ACM SIGIR Conference on Research and Development in Information Retrieval},
  pages={481--490},
  year={2021}
}

@inproceedings{SP9,
  title={If You Want to Go Far Go Together: Unsupervised Joint Candidate Evidence Retrieval for Multi-hop Question Answering},
  author={Yadav, Vikas and Bethard, Steven and Surdeanu, Mihai},
  booktitle={Proceedings of the 2021 Conference of the North American Chapter of the Association for Computational Linguistics: Human Language Technologies},
  pages={4571--4581},
  year={2021}
}

@article{SP10,
  title={Memory augmented sequential paragraph retrieval for multi-hop question answering},
  author={Shao, Nan and Cui, Yiming and Liu, Ting and Wang, Shijin and Hu, Guoping},
  journal={arXiv preprint arXiv:2102.03741},
  year={2021}
}

@inproceedings{SP11,
  title={Combining Lexical and Dense Retrieval for Computationally Efficient Multi-hop Question Answering},
  author={Sidiropoulos, Georgios and Voskarides, Nikos and Vakulenko, Svitlana and Kanoulas, Evangelos},
  booktitle={Proceedings of the Second Workshop on Simple and Efficient Natural Language Processing},
  pages={58--63},
  year={2021}
}

@inproceedings{SP12,
  title={BAG: Bi-directional Attention Entity Graph Convolutional Network for Multi-hop Reasoning Question Answering},
  author={Cao, Yu and Fang, Meng and Tao, Dacheng},
  booktitle={Proceedings of the 2019 Conference of the North American Chapter of the Association for Computational Linguistics: Human Language Technologies, Volume 1 (Long and Short Papers)},
  pages={357--362},
  year={2019}
}

@inproceedings{SP13,
  title={Identifying Supporting Facts for Multi-hop Question Answering with Document Graph Networks},
  author={Thayaparan, Mokanarangan and Valentino, Marco and Schlegel, Viktor and Freitas, Andr{\'e}},
  booktitle={Proceedings of the Thirteenth Workshop on Graph-Based Methods for Natural Language Processing (TextGraphs-13)},
  pages={42--51},
  year={2019}
}

@inproceedings{SP14,
    title = "Multi-hop Reading Comprehension through Question Decomposition and Rescoring",
    author = "Min, Sewon  and
      Zhong, Victor  and
      Zettlemoyer, Luke  and
      Hajishirzi, Hannaneh",
    booktitle = "Proceedings of the 57th Annual Meeting of the Association for Computational Linguistics",
    month = jul,
    year = "2019",
    address = "Florence, Italy",
    publisher = "Association for Computational Linguistics",
    url = "https://aclanthology.org/P19-1613",
    doi = "10.18653/v1/P19-1613",
    pages = "6097--6109",
}

@inproceedings{SP15,
  title={Question Answering by Reasoning Across Documents with Graph Convolutional Networks},
  author={De Cao, Nicola and Aziz, Wilker and Titov, Ivan},
  booktitle={Proceedings of the 2019 Conference of the North American Chapter of the Association for Computational Linguistics: Human Language Technologies, Volume 1 (Long and Short Papers)},
  pages={2306--2317},
  year={2019}
}

@ARTICLE{SP16,
  author={Zhang, Min and Li, Feng and Wang, Yang and Zhang, Zequn and Zhou, Yanhai and Li, Xiaoyu},
  journal={IEEE Access}, 
  title={Coarse and Fine Granularity Graph Reasoning for Interpretable Multi-Hop Question Answering}, 
  year={2020},
  volume={8},
  number={},
  pages={56755-56765},
  doi={10.1109/ACCESS.2020.2981134}
}

@article{SP17,
  title={Generating followup questions for interpretable multi-hop question answering},
  author={Malon, Christopher and Bai, Bing},
  journal={arXiv preprint arXiv:2002.12344},
  year={2020}
}

@article{SP18,
  title={Learning to recover reasoning chains for multi-hop question answering via cooperative games},
  author={Feng, Yufei and Yu, Mo and Xiong, Wenhan and Guo, Xiaoxiao and Huang, Junjie and Chang, Shiyu and Campbell, Murray and Greenspan, Michael and Zhu, Xiaodan},
  journal={arXiv preprint arXiv:2004.02393},
  year={2020}
}

@article{SP19,
  title={Multi-hop question answering via reasoning chains},
  author={Chen, Jifan and Lin, Shih-ting and Durrett, Greg},
  journal={arXiv preprint arXiv:1910.02610},
  year={2019}
}

@inproceedings{SP20,
    title = "Breadth First Reasoning Graph for Multi-hop Question Answering",
    author = "Huang, Yongjie  and
      Yang, Meng",
    booktitle = "Proceedings of the 2021 Conference of the North American Chapter of the Association for Computational Linguistics: Human Language Technologies",
    month = jun,
    year = "2021",
    address = "Online",
    publisher = "Association for Computational Linguistics",
    url = "https://aclanthology.org/2021.naacl-main.464",
    doi = "10.18653/v1/2021.naacl-main.464",
    pages = "5810--5821",
}

@article{SP21,
  title={Coarse-grained decomposition and fine-grained interaction for multi-hop question answering},
  author={Cao, Xing and Liu, Yun},
  journal={Journal of Intelligent Information Systems},
  pages={1--21},
  year={2021},
  publisher={Springer}
}

@article{SP22,
  title={Iterative Hierarchical Attention for Answering Complex Questions over Long Documents},
  author={Sun, Haitian and Cohen, William W and Salakhutdinov, Ruslan},
  journal={arXiv preprint arXiv:2106.00200},
  year={2021}
}

@inproceedings{SP23,
  title={Generative Context Pair Selection for Multi-hop Question Answering},
  author={Dua, Dheeru and dos Santos, Cicero and Ng, Patrick and Athiwaratkun, Ben and Xiang, Bing and Gardner, Matt and Singh, Sameer},
  booktitle={Proceedings of the 2021 Conference on Empirical Methods in Natural Language Processing},
  pages={7009--7015},
  year={2021}
}

@inproceedings{SP24,
    title = "Answering Complex Open-domain Questions Through Iterative Query Generation",
    author = "Qi, Peng  and
      Lin, Xiaowen  and
      Mehr, Leo  and
      Wang, Zijian  and
      Manning, Christopher D.",
    booktitle = "Proceedings of the 2019 Conference on Empirical Methods in Natural Language Processing and the 9th International Joint Conference on Natural Language Processing (EMNLP-IJCNLP)",
    month = nov,
    year = "2019",
    address = "Hong Kong, China",
    publisher = "Association for Computational Linguistics",
    url = "https://aclanthology.org/D19-1261",
    doi = "10.18653/v1/D19-1261",
    pages = "2590--2602",
}

@inproceedings{SP25,
    title = "Dynamically Fused Graph Network for Multi-hop Reasoning",
    author = "Qiu, Lin  and
      Xiao, Yunxuan  and
      Qu, Yanru  and
      Zhou, Hao  and
      Li, Lei  and
      Zhang, Weinan  and
      Yu, Yong",
    booktitle = "Proceedings of the 57th Annual Meeting of the Association for Computational Linguistics",
    month = jul,
    year = "2019",
    address = "Florence, Italy",
    publisher = "Association for Computational Linguistics",
    url = "https://aclanthology.org/P19-1617",
    doi = "10.18653/v1/P19-1617",
    pages = "6140--6150",
}

@inproceedings{SP26,
  title={Is Graph Structure Necessary for Multi-hop Question Answering?},
  author={Shao, Nan and Cui, Yiming and Liu, Ting and Wang, Shijin and Hu, Guoping},
  booktitle={Proceedings of the 2020 Conference on Empirical Methods in Natural Language Processing (EMNLP)},
  pages={7187--7192},
  year={2020}
}

@inproceedings{SP27,
  title={Commonsense for Generative Multi-Hop Question Answering Tasks},
  author={Bauer, Lisa and Wang, Yicheng and Bansal, Mohit},
  booktitle={Proceedings of the 2018 Conference on Empirical Methods in Natural Language Processing},
  pages={4220--4230},
  year={2018}
}

@inproceedings{SP28,
  title={Repurposing Entailment for Multi-Hop Question Answering Tasks},
  author={Trivedi, Harsh and Kwon, Heeyoung and Khot, Tushar and Sabharwal, Ashish and Balasubramanian, Niranjan},
  booktitle={Proceedings of the 2019 Conference of the North American Chapter of the Association for Computational Linguistics: Human Language Technologies, Volume 1 (Long and Short Papers)},
  pages={2948--2958},
  year={2019}
}

@inproceedings{SP29,
  title={What’s Missing: A Knowledge Gap Guided Approach for Multi-hop Question Answering},
  author={Khot, Tushar and Sabharwal, Ashish and Clark, Peter},
  booktitle={Proceedings of the 2019 Conference on Empirical Methods in Natural Language Processing and the 9th International Joint Conference on Natural Language Processing (EMNLP-IJCNLP)},
  pages={2814--2828},
  year={2019}
}

@inproceedings{SP30,
    title = "Combining Lexical and Dense Retrieval for Computationally Efficient Multi-hop Question Answering",
    author = "Sidiropoulos, Georgios  and
      Voskarides, Nikos  and
      Vakulenko, Svitlana  and
      Kanoulas, Evangelos",
    booktitle = "Proceedings of the Second Workshop on Simple and Efficient Natural Language Processing",
    month = nov,
    year = "2021",
    address = "Virtual",
    publisher = "Association for Computational Linguistics",
    url = "https://aclanthology.org/2021.sustainlp-1.7",
    doi = "10.18653/v1/2021.sustainlp-1.7",
    pages = "58--63",
}

@inproceedings{SP32,
  title={Select, answer and explain: Interpretable multi-hop reading comprehension over multiple documents},
  author={Tu, Ming and Huang, Kevin and Wang, Guangtao and Huang, Jing and He, Xiaodong and Zhou, Bowen},
  booktitle={Proceedings of the AAAI Conference on Artificial Intelligence},
  volume={34},
  number={05},
  pages={9073--9080},
  year={2020}
}

@article{SP37,
  title={Multi-paragraph reasoning with knowledge-enhanced graph neural network},
  author={Ye, Deming and Lin, Yankai and Liu, Zhenghao and Liu, Zhiyuan and Sun, Maosong},
  journal={arXiv preprint arXiv:1911.02170},
  year={2019}
}

@article{SP45,
  title={Exploring graph-structured passage representation for multi-hop reading comprehension with graph neural networks},
  author={Song, Linfeng and Wang, Zhiguo and Yu, Mo and Zhang, Yue and Florian, Radu and Gildea, Daniel},
  journal={arXiv preprint arXiv:1809.02040},
  year={2018}
}

@article{SP51,
  author    = {Wenhan Xiong and
               Xiang Lorraine Li and
               Srinivasan Iyer and
               Jingfei Du and
               Patrick S. H. Lewis and
               William Yang Wang and
               Yashar Mehdad and
               Wen{-}tau Yih and
               Sebastian Riedel and
               Douwe Kiela and
               Barlas Oguz},
  title     = {Answering Complex Open-Domain Questions with Multi-Hop Dense Retrieval},
  journal   = {CoRR},
  volume    = {abs/2009.12756},
  year      = {2020},
  url       = {https://arxiv.org/abs/2009.12756},
  eprinttype = {arXiv},
  eprint    = {2009.12756},
  bibsource = {dblp computer science bibliography, https://dblp.org}
}

@inproceedings{QG1,
  title={Difficulty-controllable multi-hop question generation from knowledge graphs},
  author={Kumar, Vishwajeet and Hua, Yuncheng and Ramakrishnan, Ganesh and Qi, Guilin and Gao, Lianli and Li, Yuan-Fang},
  booktitle={International Semantic Web Conference},
  pages={382--398},
  year={2019},
  organization={Springer}
}

@inproceedings{QG2,
  title={Multi-hop Question Generation with Graph Convolutional Network},
  author={Su, Dan and Xu, Yan and Dai, Wenliang and Ji, Ziwei and Yu, Tiezheng and Fung, Pascale},
  booktitle={Findings of the Association for Computational Linguistics: EMNLP 2020},
  pages={4636--4647},
  year={2020}
}

@inproceedings{QG3,
  title={Reinforced Multi-task Approach for Multi-hop Question Generation},
  author={Gupta, Deepak and Chauhan, Hardik and Akella, Ravi Tej and Ekbal, Asif and Bhattacharyya, Pushpak},
  booktitle={Proceedings of the 28th International Conference on Computational Linguistics},
  pages={2760--2775},
  year={2020}
}

@article{QG4,
  title={Stronger Transformers for Neural Multi-Hop Question Generation},
  author={Sachan, Devendra Singh and Wu, Lingfei and Sachan, Mrinmaya and Hamilton, William},
  journal={arXiv preprint arXiv:2010.11374},
  year={2020}
}

@inproceedings{QG5,
  title={Unsupervised Multi-hop Question Answering by Question Generation},
  author={Pan, Liangming and Chen, Wenhu and Xiong, Wenhan and Kan, Min-Yen and Wang, William Yang},
  booktitle={Proceedings of the 2021 Conference of the North American Chapter of the Association for Computational Linguistics: Human Language Technologies},
  pages={5866--5880},
  year={2021}
}

@inproceedings{QG6,
    title = "Low-Resource Generation of Multi-hop Reasoning Questions",
    author = "Yu, Jianxing  and
      Liu, Wei  and
      Qiu, Shuang  and
      Su, Qinliang  and
      Wang, Kai  and
      Quan, Xiaojun  and
      Yin, Jian",
    booktitle = "Proceedings of the 58th Annual Meeting of the Association for Computational Linguistics",
    month = jul,
    year = "2020",
    address = "Online",
    publisher = "Association for Computational Linguistics",
    url = "https://aclanthology.org/2020.acl-main.601",
    doi = "10.18653/v1/2020.acl-main.601",
    pages = "6729--6739",
}

@inproceedings{DP1,
  title={HotpotQA: A Dataset for Diverse, Explainable Multi-hop Question Answering},
  author={Yang, Zhilin and Qi, Peng and Zhang, Saizheng and Bengio, Yoshua and Cohen, William and Salakhutdinov, Ruslan and Manning, Christopher D},
  booktitle={Proceedings of the 2018 Conference on Empirical Methods in Natural Language Processing},
  pages={2369--2380},
  year={2018}
}

@article{DP2,
  title={Constructing datasets for multi-hop reading comprehension across documents},
  author={Welbl, Johannes and Stenetorp, Pontus and Riedel, Sebastian},
  journal={Transactions of the Association for Computational Linguistics},
  volume={6},
  pages={287--302},
  year={2018},
  publisher={MIT Press}
}

@article{DP3,
  title={The narrativeqa reading comprehension challenge},
  author={Ko{\v{c}}isk{\`y}, Tom{\'a}{\v{s}} and Schwarz, Jonathan and Blunsom, Phil and Dyer, Chris and Hermann, Karl Moritz and Melis, G{\'a}bor and Grefenstette, Edward},
  journal={Transactions of the Association for Computational Linguistics},
  volume={6},
  pages={317--328},
  year={2018},
  publisher={MIT Press}
}

@inproceedings{DP4,
  title={Can a Suit of Armor Conduct Electricity? A New Dataset for Open Book Question Answering},
  author={Mihaylov, Todor and Clark, Peter and Khot, Tushar and Sabharwal, Ashish},
  booktitle={Proceedings of the 2018 Conference on Empirical Methods in Natural Language Processing},
  pages={2381--2391},
  year={2018}
}

@inproceedings{DP5,
  title={Looking beyond the surface: A challenge set for reading comprehension over multiple sentences},
  author={Khashabi, Daniel and Chaturvedi, Snigdha and Roth, Michael and Upadhyay, Shyam and Roth, Dan},
  booktitle={Proceedings of the 2018 Conference of the North American Chapter of the Association for Computational Linguistics: Human Language Technologies, Volume 1 (Long Papers)},
  pages={252--262},
  year={2018}
}

@inproceedings{DP6,
  title={HybridQA: A Dataset of Multi-Hop Question Answering over Tabular and Textual Data},
  author={Chen, Wenhu and Zha, Hanwen and Chen, Zhiyu and Xiong, Wenhan and Wang, Hong and Wang, William Yang},
  booktitle={Findings of the Association for Computational Linguistics: EMNLP 2020},
  pages={1026--1036},
  year={2020}
}

@inproceedings{DP7,
  title={Qasc: A dataset for question answering via sentence composition},
  author={Khot, Tushar and Clark, Peter and Guerquin, Michal and Jansen, Peter and Sabharwal, Ashish},
  booktitle={Proceedings of the AAAI Conference on Artificial Intelligence},
  volume={34},
  number={05},
  pages={8082--8090},
  year={2020}
}

@inproceedings{AP1,
  title={Multi-hop Inference for Sentence-level TextGraphs: How Challenging is Meaningfully Combining Information for Science Question Answering?},
  author={Jansen, Peter},
  booktitle={Proceedings of the Twelfth Workshop on Graph-Based Methods for Natural Language Processing (TextGraphs-12)},
  pages={12--17},
  year={2018}
}

@inproceedings{AP2,
  title={Do Multi-hop Readers Dream of Reasoning Chains?},
  author={Wang, Haoyu and Yu, Mo and Guo, Xiaoxiao and Das, Rajarshi and Xiong, Wenhan and Gao, Tian},
  booktitle={Proceedings of the 2nd Workshop on Machine Reading for Question Answering},
  pages={91--97},
  year={2019}
}

@inproceedings{AP3,
    title = "Understanding Dataset Design Choices for Multi-hop Reasoning",
    author = "Chen, Jifan  and
      Durrett, Greg",
    booktitle = "Proceedings of the 2019 Conference of the North {A}merican Chapter of the Association for Computational Linguistics: Human Language Technologies, Volume 1 (Long and Short Papers)",
    month = jun,
    year = "2019",
    address = "Minneapolis, Minnesota",
    publisher = "Association for Computational Linguistics",
    url = "https://aclanthology.org/N19-1405",
    doi = "10.18653/v1/N19-1405",
    pages = "4026--4032",
}

@inproceedings{AP4,
    title = "Compositional Questions Do Not Necessitate Multi-hop Reasoning",
    author = "Min, Sewon  and
      Wallace, Eric  and
      Singh, Sameer  and
      Gardner, Matt  and
      Hajishirzi, Hannaneh  and
      Zettlemoyer, Luke",
    booktitle = "Proceedings of the 57th Annual Meeting of the Association for Computational Linguistics",
    month = jul,
    year = "2019",
    address = "Florence, Italy",
    publisher = "Association for Computational Linguistics",
    url = "https://aclanthology.org/P19-1416",
    doi = "10.18653/v1/P19-1416",
    pages = "4249--4257",
}

@inproceedings{AP5,
    title = "Is Multihop {QA} in {DiRe} Condition? Measuring and Reducing Disconnected Reasoning",
    author = "Trivedi, Harsh  and
      Balasubramanian, Niranjan  and
      Khot, Tushar  and
      Sabharwal, Ashish",
    booktitle = "Proceedings of the 2020 Conference on Empirical Methods in Natural Language Processing (EMNLP)",
    month = nov,
    year = "2020",
    address = "Online",
    publisher = "Association for Computational Linguistics",
    url = "https://aclanthology.org/2020.emnlp-main.712",
    doi = "10.18653/v1/2020.emnlp-main.712",
    pages = "8846--8863",
}

@inproceedings{AP6,
    title = "Learning to Explain: Datasets and Models for Identifying Valid Reasoning Chains in Multihop Question-Answering",
    author = "Jhamtani, Harsh  and
      Clark, Peter",
    booktitle = "Proceedings of the 2020 Conference on Empirical Methods in Natural Language Processing (EMNLP)",
    month = nov,
    year = "2020",
    address = "Online",
    publisher = "Association for Computational Linguistics",
    url = "https://aclanthology.org/2020.emnlp-main.10",
    doi = "10.18653/v1/2020.emnlp-main.10",
    pages = "137--150",
}

@inproceedings{AP7,
    title = "{R}4{C}: A Benchmark for Evaluating {RC} Systems to Get the Right Answer for the Right Reason",
    author = "Inoue, Naoya  and
      Stenetorp, Pontus  and
      Inui, Kentaro",
    booktitle = "Proceedings of the 58th Annual Meeting of the Association for Computational Linguistics",
    month = jul,
    year = "2020",
    address = "Online",
    publisher = "Association for Computational Linguistics",
    url = "https://aclanthology.org/2020.acl-main.602",
    doi = "10.18653/v1/2020.acl-main.602",
    pages = "6740--6750",
}

@inproceedings{AP8,
    title = "Do Multi-Hop Question Answering Systems Know How to Answer the Single-Hop Sub-Questions?",
    author = "Tang, Yixuan  and
      Ng, Hwee Tou  and
      Tung, Anthony",
    booktitle = "Proceedings of the 16th Conference of the European Chapter of the Association for Computational Linguistics: Main Volume",
    month = apr,
    year = "2021",
    address = "Online",
    publisher = "Association for Computational Linguistics",
    url = "https://aclanthology.org/2021.eacl-main.283",
    doi = "10.18653/v1/2021.eacl-main.283",
    pages = "3244--3249",
}

@article{facts, 
    title = "Balancing Spreads of Influence in a     Social Network",
    volume = "34", 
    url = "https://ojs.aaai.org/index.php/AAAI/article/view/5327",
    doi = "10.1609/aaai.v34i01.5327",
    journal = "Proceedings of the AAAI Conference on Artificial Intelligence",
    author = "Becker, Ruben and Corò, Federico and D’Angelo, Gianlorenzo and Gilbert, Hugo",
    year = "2020",
    month = apr,
    pages= "3--10" }

@article{archival,
  author    = {Jiexin Wang and
               Adam Jatowt and
               Masatoshi Yoshikawa},
  title     = {ArchivalQA: {A} Large-scale Benchmark Dataset for Open Domain Question
               Answering over Archival News Collections},
  journal   = {CoRR},
  volume    = {abs/2109.03438},
  year      = {2021},
  url       = {https://arxiv.org/abs/2109.03438},
  eprinttype = {arXiv},
  eprint    = {2109.03438},
  bibsource = {dblp computer science bibliography, https://dblp.org}
}

@inproceedings{worldtree,
    title = "{W}orld{T}ree: A Corpus of Explanation Graphs for Elementary Science Questions supporting Multi-hop Inference",
    author = "Jansen, Peter  and
      Wainwright, Elizabeth  and
      Marmorstein, Steven  and
      Morrison, Clayton",
    booktitle = "Proceedings of the Eleventh International Conference on Language Resources and Evaluation ({LREC} 2018)",
    month = may,
    year = "2018",
    address = "Miyazaki, Japan",
    publisher = "European Language Resources Association (ELRA)",
    url = "https://aclanthology.org/L18-1433",
}

@inproceedings{albert,
  title={ALBERT: A Lite BERT for Self-supervised Learning of Language Representations},
  author={Lan, Zhenzhong and Chen, Mingda and Goodman, Sebastian and Gimpel, Kevin and Sharma, Piyush and Soricut, Radu},
  booktitle={International Conference on Learning Representations},
  year={2019}
}

@inproceedings{squad2,
  title={SQuAD: 100,000+ Questions for Machine Comprehension of Text},
  author={Rajpurkar, Pranav and Zhang, Jian and Lopyrev, Konstantin and Liang, Percy},
  booktitle={Proceedings of the 2016 Conference on Empirical Methods in Natural Language Processing},
  pages={2383--2392},
  year={2016}
}

@inproceedings{squad,
    title = "{SQ}u{AD}: 100,000+ Questions for Machine Comprehension of Text",
    author = "Rajpurkar, Pranav  and
      Zhang, Jian  and
      Lopyrev, Konstantin  and
      Liang, Percy",
    booktitle = "Proceedings of the 2016 Conference on Empirical Methods in Natural Language Processing",
    month = nov,
    year = "2016",
    address = "Austin, Texas",
    publisher = "Association for Computational Linguistics",
    url = "https://aclanthology.org/D16-1264",
    doi = "10.18653/v1/D16-1264",
    pages = "2383--2392",
}

@article{complex-survey-1,
  title={A survey on complex question answering over knowledge base: Recent advances and challenges},
  author={Fu, Bin and Qiu, Yunqi and Tang, Chengguang and Li, Yang and Yu, Haiyang and Sun, Jian},
  journal={arXiv preprint arXiv:2007.13069},
  year={2020}
}

@inproceedings{complex-survey-2,
  title={A survey on complex knowledge base question answering: Methods, challenges and solutions},
  author={Lan, Yunshi and He, Gaole and JIANG, Jinhao and JIANG, Jing and ZHAO, Wayne Xin and WEN, Ji-Rong},
  year={2021},
  organization={IJCAI}
}

@inproceedings{visual-qa-survey,
  title={Visual question answering using deep learning: A survey and performance analysis},
  author={Srivastava, Yash and Murali, Vaishnav and Dubey, Shiv Ram and Mukherjee, Snehasis},
  booktitle={International Conference on Computer Vision and Image Processing},
  pages={75--86},
  year={2020},
  organization={Springer}
}

@article{min2020ambigqa,   title={AmbigQA: Answering ambiguous open-domain questions},   author={Min, Sewon and Michael, Julian and Hajishirzi, Hannaneh and Zettlemoyer, Luke},   journal={arXiv preprint arXiv:2004.10645},   year={2020} }

@article{hermann2015teaching,
  title={Teaching machines to read and comprehend},
  author={Hermann, Karl Moritz and Kocisky, Tomas and Grefenstette, Edward and Espeholt, Lasse and Kay, Will and Suleyman, Mustafa and Blunsom, Phil},
  journal={Advances in neural information processing systems},
  volume={28},
  year={2015}
}

@article{bidaf,
  author    = {Min Joon Seo and
               Aniruddha Kembhavi and
               Ali Farhadi and
               Hannaneh Hajishirzi},
  title     = {Bidirectional Attention Flow for Machine Comprehension},
  journal   = {CoRR},
  volume    = {abs/1611.01603},
  year      = {2016},
  url       = {http://arxiv.org/abs/1611.01603},
  eprinttype = {arXiv},
  eprint    = {1611.01603},
  bibsource = {dblp computer science bibliography, https://dblp.org}
}

@inproceedings{min-etal-2018-efficient,
    title = "Efficient and Robust Question Answering from Minimal Context over Documents",
    author = "Min, Sewon  and
      Zhong, Victor  and
      Socher, Richard  and
      Xiong, Caiming",
    booktitle = "Proceedings of the 56th Annual Meeting of the Association for Computational Linguistics (Volume 1: Long Papers)",
    month = jul,
    year = "2018",
    address = "Melbourne, Australia",
    publisher = "Association for Computational Linguistics",
    url = "https://aclanthology.org/P18-1160",
    doi = "10.18653/v1/P18-1160",
    pages = "1725--1735",
}

@article{zaib2021conversational,
  title={Conversational question answering: A survey},
  author={Zaib, Munazza and Zhang, Wei Emma and Sheng, Quan Z and Mahmood, Adnan and Zhang, Yang},
  journal={arXiv preprint arXiv:2106.00874},
  year={2021}
}

@article{samek2017explainable,
  title={Explainable artificial intelligence: Understanding, visualizing and interpreting deep learning models},
  author={Samek, Wojciech and Wiegand, Thomas and M{\"u}ller, Klaus-Robert},
  journal={arXiv preprint arXiv:1708.08296},
  year={2017}
}

@inproceedings{alvarez-melis-jaakkola-2017-causal,
    title = "A causal framework for explaining the predictions of black-box sequence-to-sequence models",
    author = "Alvarez-Melis, David  and
      Jaakkola, Tommi",
    booktitle = "Proceedings of the 2017 Conference on Empirical Methods in Natural Language Processing",
    month = sep,
    year = "2017",
    address = "Copenhagen, Denmark",
    publisher = "Association for Computational Linguistics",
    url = "https://aclanthology.org/D17-1042",
    doi = "10.18653/v1/D17-1042",
    pages = "412--421",
}

@article{arras2016relevant,
  title={What is Relevant in a Text Document?": An Interpretable Machine Learning Approach. CoRR abs/1612.07843 (2016)},
  author={Arras, Leila and Horn, Franziska and Montavon, Gr{\'e}goire and M{\"u}ller, Klaus-Robert and Samek, Wojciech},
  journal={arXiv preprint arXiv:1612.07843},
  year={2016}
}

@inproceedings{biran2017explanation,
  title={Explanation and justification in machine learning: A survey},
  author={Biran, Or and Cotton, Courtenay},
  booktitle={IJCAI-17 workshop on explainable AI (XAI)},
  volume={8},
  number={1},
  pages={8--13},
  year={2017}
}

@inproceedings{gilpin2018explaining,
  title={Explaining explanations: An overview of interpretability of machine learning},
  author={Gilpin, Leilani H and Bau, David and Yuan, Ben Z and Bajwa, Ayesha and Specter, Michael and Kagal, Lalana},
  booktitle={2018 IEEE 5th International Conference on data science and advanced analytics (DSAA)},
  pages={80--89},
  year={2018},
  organization={IEEE}
}

@inproceedings{bhagavatula2013methods,
  title={Methods for exploring and mining tables on wikipedia},
  author={Bhagavatula, Chandra Sekhar and Noraset, Thanapon and Downey, Doug},
  booktitle={Proceedings of the ACM SIGKDD workshop on interactive data exploration and analytics},
  pages={18--26},
  year={2013}
}

@article{wishart2008drugbank,
  title={DrugBank: a knowledgebase for drugs, drug actions and drug targets},
  author={Wishart, David S and Knox, Craig and Guo, An Chi and Cheng, Dean and Shrivastava, Savita and Tzur, Dan and Gautam, Bijaya and Hassanali, Murtaza},
  journal={Nucleic acids research},
  volume={36},
  number={suppl\_1},
  pages={D901--D906},
  year={2008},
  publisher={Oxford University Press}
}

@article{gatt2018survey,
  title={Survey of the state of the art in natural language generation: Core tasks, applications and evaluation},
  author={Gatt, Albert and Krahmer, Emiel},
  journal={Journal of Artificial Intelligence Research},
  volume={61},
  pages={65--170},
  year={2018}
}

@inproceedings{novikova-etal-2017-need,
    title = "Why We Need New Evaluation Metrics for {NLG}",
    author = "Novikova, Jekaterina  and
      Du{\v{s}}ek, Ond{\v{r}}ej  and
      Cercas Curry, Amanda  and
      Rieser, Verena",
    booktitle = "Proceedings of the 2017 Conference on Empirical Methods in Natural Language Processing",
    month = sep,
    year = "2017",
    address = "Copenhagen, Denmark",
    publisher = "Association for Computational Linguistics",
    url = "https://aclanthology.org/D17-1238",
    doi = "10.18653/v1/D17-1238",
    pages = "2241--2252",
}

@inproceedings{geva-etal-2019-modeling,
    title = "Are We Modeling the Task or the Annotator? An Investigation of Annotator Bias in Natural Language Understanding Datasets",
    author = "Geva, Mor  and
      Goldberg, Yoav  and
      Berant, Jonathan",
    booktitle = "Proceedings of the 2019 Conference on Empirical Methods in Natural Language Processing and the 9th International Joint Conference on Natural Language Processing (EMNLP-IJCNLP)",
    month = nov,
    year = "2019",
    address = "Hong Kong, China",
    publisher = "Association for Computational Linguistics",
    url = "https://aclanthology.org/D19-1107",
    doi = "10.18653/v1/D19-1107",
    pages = "1161--1166",
}

@inproceedings{dua-etal-2020-benefits,
    title = "Benefits of Intermediate Annotations in Reading Comprehension",
    author = "Dua, Dheeru  and
      Singh, Sameer  and
      Gardner, Matt",
    booktitle = "Proceedings of the 58th Annual Meeting of the Association for Computational Linguistics",
    month = jul,
    year = "2020",
    address = "Online",
    publisher = "Association for Computational Linguistics",
    url = "https://aclanthology.org/2020.acl-main.497",
    doi = "10.18653/v1/2020.acl-main.497",
    pages = "5627--5634",
}

@inproceedings{gururangan-etal-2018-annotation,
    title = "Annotation Artifacts in Natural Language Inference Data",
    author = "Gururangan, Suchin  and
      Swayamdipta, Swabha  and
      Levy, Omer  and
      Schwartz, Roy  and
      Bowman, Samuel  and
      Smith, Noah A.",
    booktitle = "Proceedings of the 2018 Conference of the North {A}merican Chapter of the Association for Computational Linguistics: Human Language Technologies, Volume 2 (Short Papers)",
    month = jun,
    year = "2018",
    address = "New Orleans, Louisiana",
    publisher = "Association for Computational Linguistics",
    url = "https://aclanthology.org/N18-2017",
    doi = "10.18653/v1/N18-2017",
    pages = "107--112",
}

@inproceedings{jia2017adversarial,
  title={Adversarial Examples for Evaluating Reading Comprehension Systems},
  author={Jia, Robin and Liang, Percy},
  booktitle={Proceedings of the 2017 Conference on Empirical Methods in Natural Language Processing},
  pages={2021--2031},
  year={2017}
}

@article{wangIRJ21,
  author    = {Jiexin Wang and
               Adam Jatowt and
               Michael F{\"{a}}rber and
               Masatoshi Yoshikawa},
  title     = {Improving question answering for event-focused questions in temporal
               collections of news articles},
  journal   = {Inf. Retr. J.},
  volume    = {24},
  number    = {1},
  pages     = {29--54},
  year      = {2021},
}

@inproceedings{wangSIGIR21,
  author    = {Jiexin Wang and
               Adam Jatowt and
               Masatoshi Yoshikawa},
  title     = {Event Occurrence Date Estimation based on Multivariate Time Series
               Analysis over Temporal Document Collections},
  booktitle = {{SIGIR} '21: The 44th International {ACM} {SIGIR} Conference on Research
               and Development in Information Retrieval, Virtual Event, Canada, July
               11-15, 2021},
  pages     = {398--407},
  publisher = {{ACM}},
  year      = {2021},
}

@inproceedings{bert,
  title={BERT: Pre-training of Deep Bidirectional Transformers for Language Understanding},
  author={Kenton, Jacob Devlin Ming-Wei Chang and Toutanova, Lee Kristina},
  booktitle={Proceedings of NAACL-HLT},
  pages={4171--4186},
  year={2019}
}

@inproceedings{elmo,
    title = "Deep Contextualized Word Representations",
    author = "Peters, Matthew E.  and
      Neumann, Mark  and
      Iyyer, Mohit  and
      Gardner, Matt  and
      Clark, Christopher  and
      Lee, Kenton  and
      Zettlemoyer, Luke",
    booktitle = "Proceedings of the 2018 Conference of the North {A}merican Chapter of the Association for Computational Linguistics: Human Language Technologies, Volume 1 (Long Papers)",
    month = jun,
    year = "2018",
    address = "New Orleans, Louisiana",
    publisher = "Association for Computational Linguistics",
    url = "https://aclanthology.org/N18-1202",
    doi = "10.18653/v1/N18-1202",
    pages = "2227--2237",
}

@inproceedings{chen-etal-2017-reading,
    title = "Reading {W}ikipedia to Answer Open-Domain Questions",
    author = "Chen, Danqi  and
      Fisch, Adam  and
      Weston, Jason  and
      Bordes, Antoine",
    booktitle = "Proceedings of the 55th Annual Meeting of the Association for Computational Linguistics (Volume 1: Long Papers)",
    month = jul,
    year = "2017",
    address = "Vancouver, Canada",
    publisher = "Association for Computational Linguistics",
    url = "https://aclanthology.org/P17-1171",
    doi = "10.18653/v1/P17-1171",
    pages = "1870--1879",
}

@inproceedings{glove,
  title={Glove: Global vectors for word representation},
  author={Pennington, Jeffrey and Socher, Richard and Manning, Christopher D},
  booktitle={Proceedings of the 2014 conference on empirical methods in natural language processing (EMNLP)},
  pages={1532--1543},
  year={2014}
}

@article{lstm,
  title={Long short-term memory},
  author={Hochreiter, Sepp and Schmidhuber, J{\"u}rgen},
  journal={Neural computation},
  volume={9},
  number={8},
  pages={1735--1780},
  year={1997},
  publisher={MIT Press}
}

@article{roberta,
  title={Roberta: A robustly optimized bert pretraining approach},
  author={Liu, Yinhan and Ott, Myle and Goyal, Naman and Du, Jingfei and Joshi, Mandar and Chen, Danqi and Levy, Omer and Lewis, Mike and Zettlemoyer, Luke and Stoyanov, Veselin},
  journal={arXiv preprint arXiv:1907.11692},
  year={2019}
}

@article{transformer,
  title={Attention is all you need},
  author={Vaswani, Ashish and Shazeer, Noam and Parmar, Niki and Uszkoreit, Jakob and Jones, Llion and Gomez, Aidan N and Kaiser, {\L}ukasz and Polosukhin, Illia},
  journal={Advances in neural information processing systems},
  volume={30},
  year={2017}
}

@inproceedings{pgn,
  title={Get To The Point: Summarization with Pointer-Generator Networks},
  author={See, Abigail and Liu, Peter J and Manning, Christopher D},
  booktitle={Proceedings of the 55th Annual Meeting of the Association for Computational Linguistics (Volume 1: Long Papers)},
  pages={1073--1083},
  year={2017}
}

@article{graves2014neural,
  title={Neural turing machines},
  author={Graves, Alex and Wayne, Greg and Danihelka, Ivo},
  journal={arXiv preprint arXiv:1410.5401},
  year={2014}
}

@inproceedings{gat,
  title={Graph Attention Networks},
  author={Veli{\v{c}}kovi{\'c}, Petar and Cucurull, Guillem and Casanova, Arantxa and Romero, Adriana and Li{\`o}, Pietro and Bengio, Yoshua},
  booktitle={International Conference on Learning Representations},
  year={2018}
}

@article{kim2018bilinear,
  title={Bilinear attention networks},
  author={Kim, Jin-Hwa and Jun, Jaehyun and Zhang, Byoung-Tak},
  journal={Advances in Neural Information Processing Systems},
  volume={31},
  year={2018}
}

@inproceedings{ainslie-etal-2020-etc,
    title = "{ETC}: Encoding Long and Structured Inputs in Transformers",
    author = "Ainslie, Joshua  and
      Ontanon, Santiago  and
      Alberti, Chris  and
      Cvicek, Vaclav  and
      Fisher, Zachary  and
      Pham, Philip  and
      Ravula, Anirudh  and
      Sanghai, Sumit  and
      Wang, Qifan  and
      Yang, Li",
    booktitle = "Proceedings of the 2020 Conference on Empirical Methods in Natural Language Processing (EMNLP)",
    month = nov,
    year = "2020",
    address = "Online",
    publisher = "Association for Computational Linguistics",
    url = "https://aclanthology.org/2020.emnlp-main.19",
    doi = "10.18653/v1/2020.emnlp-main.19",
    pages = "268--284",
}

@article{van2018representation,
  title={Representation learning with contrastive predictive coding},
  author={Van den Oord, Aaron and Li, Yazhe and Vinyals, Oriol},
  journal={arXiv e-prints},
  pages={arXiv--1807},
  year={2018}
}

@inproceedings{chen-etal-2017-enhanced,
    title = "Enhanced {LSTM} for Natural Language Inference",
    author = "Chen, Qian  and
      Zhu, Xiaodan  and
      Ling, Zhen-Hua  and
      Wei, Si  and
      Jiang, Hui  and
      Inkpen, Diana",
    booktitle = "Proceedings of the 55th Annual Meeting of the Association for Computational Linguistics (Volume 1: Long Papers)",
    month = jul,
    year = "2017",
    address = "Vancouver, Canada",
    publisher = "Association for Computational Linguistics",
    url = "https://aclanthology.org/P17-1152",
    doi = "10.18653/v1/P17-1152",
    pages = "1657--1668",
}

@inproceedings{bowman2015large,
  title={A large annotated corpus for learning natural language inference},
  author={Bowman, Samuel R and Angeli, Gabor and Potts, Christopher and Manning, Christopher D},
  booktitle={Conference on Empirical Methods in Natural Language Processing, EMNLP 2015},
  pages={632--642},
  year={2015},
  organization={Association for Computational Linguistics (ACL)}
}

@inproceedings{williams-etal-2018-broad,
    title = "A Broad-Coverage Challenge Corpus for Sentence Understanding through Inference",
    author = "Williams, Adina  and
      Nangia, Nikita  and
      Bowman, Samuel",
    booktitle = "Proceedings of the 2018 Conference of the North {A}merican Chapter of the Association for Computational Linguistics: Human Language Technologies, Volume 1 (Long Papers)",
    month = jun,
    year = "2018",
    address = "New Orleans, Louisiana",
    publisher = "Association for Computational Linguistics",
    url = "https://aclanthology.org/N18-1101",
    doi = "10.18653/v1/N18-1101",
    pages = "1112--1122",
}

@article{speer2016conceptnet,
  title={ConceptNet 5.5: An Open Multilingual Graph of General Knowledge. Singh 2002 (2016)},
  author={Speer, Robyn and Chin, Joshua and Havasi, Catherine},
  journal={arXiv preprint arxiv:1612.03975},
  year={2016}
}

@article{church-hanks-1990-word,
    title = "Word Association Norms, Mutual Information, and Lexicography",
    author = "Church, Kenneth Ward  and
      Hanks, Patrick",
    journal = "Computational Linguistics",
    volume = "16",
    number = "1",
    year = "1990",
    url = "https://aclanthology.org/J90-1003",
    pages = "22--29",
}

@inproceedings{cho2014learning,
  title={Learning phrase representations using RNN encoder-decoder for statistical machine translation},
  author={Cho, Kyunghyun and van Merrienboer, B and Gulcehre, Caglar and Bougares, F and Schwenk, H and Bengio, Yoshua},
  booktitle={Conference on Empirical Methods in Natural Language Processing (EMNLP 2014)},
  year={2014}
}

@article{wang2016machine,
  title={Machine comprehension using match-lstm and answer pointer},
  author={Wang, Shuohang and Jiang, Jing},
  journal={arXiv preprint arXiv:1608.07905},
  year={2016}
}

@inproceedings{saeidi2018interpretation,
  title={Interpretation of Natural Language Rules in Conversational Machine Reading},
  author={Saeidi, Marzieh and Bartolo, Max and Lewis, Patrick and Singh, Sameer and Rockt{\"a}schel, Tim and Sheldon, Mike and Bouchard, Guillaume and Riedel, Sebastian},
  booktitle={EMNLP},
  year={2018}
}

@article{clark2018think,
  title={Think you have solved question answering? try arc, the ai2 reasoning challenge},
  author={Clark, Peter and Cowhey, Isaac and Etzioni, Oren and Khot, Tushar and Sabharwal, Ashish and Schoenick, Carissa and Tafjord, Oyvind},
  journal={arXiv preprint arXiv:1803.05457},
  year={2018}
}

@inproceedings{dasigi-etal-2021-dataset,
    title = "A Dataset of Information-Seeking Questions and Answers Anchored in Research Papers",
    author = "Dasigi, Pradeep  and
      Lo, Kyle  and
      Beltagy, Iz  and
      Cohan, Arman  and
      Smith, Noah A.  and
      Gardner, Matt",
    booktitle = "Proceedings of the 2021 Conference of the North American Chapter of the Association for Computational Linguistics: Human Language Technologies",
    month = jun,
    year = "2021",
    address = "Online",
    publisher = "Association for Computational Linguistics",
    url = "https://aclanthology.org/2021.naacl-main.365",
    doi = "10.18653/v1/2021.naacl-main.365",
    pages = "4599--4610",
}

@inproceedings{papineni2002bleu,
  title={Bleu: a method for automatic evaluation of machine translation},
  author={Papineni, Kishore and Roukos, Salim and Ward, Todd and Zhu, Wei-Jing},
  booktitle={Proceedings of the 40th annual meeting of the Association for Computational Linguistics},
  pages={311--318},
  year={2002}
}

@inproceedings{banerjee-lavie-2005-meteor,
    title = "{METEOR}: An Automatic Metric for {MT} Evaluation with Improved Correlation with Human Judgments",
    author = "Banerjee, Satanjeev  and
      Lavie, Alon",
    booktitle = "Proceedings of the {ACL} Workshop on Intrinsic and Extrinsic Evaluation Measures for Machine Translation and/or Summarization",
    month = jun,
    year = "2005",
    address = "Ann Arbor, Michigan",
    publisher = "Association for Computational Linguistics",
    url = "https://aclanthology.org/W05-0909",
    pages = "65--72",
}

@inproceedings{lin-2004-rouge,
    title = "{ROUGE}: A Package for Automatic Evaluation of Summaries",
    author = "Lin, Chin-Yew",
    booktitle = "Text Summarization Branches Out",
    month = jul,
    year = "2004",
    address = "Barcelona, Spain",
    publisher = "Association for Computational Linguistics",
    url = "https://aclanthology.org/W04-1013",
    pages = "74--81",
}

@article{vedantam2014cider,
  title={Cider: consensus-based image description evaluation. CoRR},
  author={Vedantam, Ramakrishna and Zitnick, C Lawrence and Parikh, Devi},
  journal={arXiv preprint arXiv:1411.5726},
  year={2014}
}

@article{liu2009mean,
  title={Mean average precision},
  author={Liu, Ling and {\"O}zsu, M Tamer},
  journal={Encyclopedia of Database Systems2009},
  volume={1703},
  year={2009}
}

@inproceedings{WangJ0Y20,
  author    = {Jiexin Wang and
               Adam Jatowt and
               Michael F{\"{a}}rber and
               Masatoshi Yoshikawa},
  title     = {Answering Event-Related Questions over Long-Term News Article Archives},
  booktitle = {Advances in Information Retrieval - 42nd European Conference on {IR}
               Research, {ECIR} 2020, Lisbon, Portugal, April 14-17, 2020, Proceedings,
               Part {I}},
  series    = {Lecture Notes in Computer Science},
  volume    = {12035},
  pages     = {774--789},
  publisher = {Springer},
  year      = {2020}
}

@inproceedings{kadlec-etal-2017-knowledge,
    title = "Knowledge Base Completion: Baselines Strike Back",
    author = "Kadlec, Rudolf  and
      Bajgar, Ondrej  and
      Kleindienst, Jan",
    booktitle = "Proceedings of the 2nd Workshop on Representation Learning for {NLP}",
    month = aug,
    year = "2017",
    address = "Vancouver, Canada",
    publisher = "Association for Computational Linguistics",
    url = "https://aclanthology.org/W17-2609",
    doi = "10.18653/v1/W17-2609",
    pages = "69--74",
}

@article{10.1145/219717.219748,
author = {Miller, George A.},
title = {WordNet: A Lexical Database for English},
year = {1995},
issue_date = {Nov. 1995},
publisher = {Association for Computing Machinery},
address = {New York, NY, USA},
volume = {38},
number = {11},
issn = {0001-0782},
url = {https://doi.org/10.1145/219717.219748},
doi = {10.1145/219717.219748},
journal = {Commun. ACM},
month = {nov},
pages = {39–41},
numpages = {3}
}

@inproceedings{patel2021nlp,
  title={Are NLP Models really able to Solve Simple Math Word Problems?},
  author={Patel, Arkil and Bhattamishra, Satwik and Goyal, Navin},
  booktitle={Proceedings of the 2021 Conference of the North American Chapter of the Association for Computational Linguistics: Human Language Technologies},
  pages={2080--2094},
  year={2021}
}

@article{hendrycks2021measuring,
  title={Measuring mathematical problem solving with the math dataset},
  author={Hendrycks, Dan and Burns, Collin and Kadavath, Saurav and Arora, Akul and Basart, Steven and Tang, Eric and Song, Dawn and Steinhardt, Jacob},
  journal={arXiv preprint arXiv:2103.03874},
  year={2021}
}

@article{schubotz2018introducing,
  title={Introducing mathqa: a math-aware question answering system},
  author={Schubotz, Moritz and Scharpf, Philipp and Dudhat, Kaushal and Nagar, Yash and Hamborg, Felix and Gipp, Bela},
  journal={Information Discovery and Delivery},
  year={2018},
  publisher={Emerald Publishing Limited}
}

@inproceedings{kusner2015word,
  title={From word embeddings to document distances},
  author={Kusner, Matt and Sun, Yu and Kolkin, Nicholas and Weinberger, Kilian},
  booktitle={International conference on machine learning},
  pages={957--966},
  year={2015},
  organization={PMLR}
}

@inproceedings{yu-etal-2021-multi,
    title = "Multi-{T}ime{L}ine Summarization ({MTLS}): Improving Timeline Summarization by Generating Multiple Summaries",
    author = "Yu, Yi  and
      Jatowt, Adam  and
      Doucet, Antoine  and
      Sugiyama, Kazunari  and
      Yoshikawa, Masatoshi",
    booktitle = "Proceedings of the 59th Annual Meeting of the Association for Computational Linguistics and the 11th International Joint Conference on Natural Language Processing (Volume 1: Long Papers)",
    year = "2021",
    address = "Online",
    publisher = "Association for Computational Linguistics",
    pages = "377--387"
}

@article{lin2021medical,
  title={Medical Visual Question Answering: A Survey},
  author={Lin, Zhihong and Zhang, Donghao and Tac, Qingyi and Shi, Danli and Haffari, Gholamreza and Wu, Qi and He, Mingguang and Ge, Zongyuan},
  journal={arXiv preprint arXiv:2111.10056},
  year={2021}
}

@article{jin2022biomedical,
  title={Biomedical Question Answering: A Survey of Approaches and Challenges},
  author={Jin, Qiao and Yuan, Zheng and Xiong, Guangzhi and Yu, Qianlan and Ying, Huaiyuan and Tan, Chuanqi and Chen, Mosha and Huang, Songfang and Liu, Xiaozhong and Yu, Sheng},
  journal={ACM Computing Surveys (CSUR)},
  volume={55},
  number={2},
  pages={1--36},
  year={2022},
  publisher={ACM New York, NY}
}

@article{wu2017visual,
  title={Visual question answering: A survey of methods and datasets},
  author={Wu, Qi and Teney, Damien and Wang, Peng and Shen, Chunhua and Dick, Anthony and van den Hengel, Anton},
  journal={Computer Vision and Image Understanding},
  volume={163},
  pages={21--40},
  year={2017},
  publisher={Elsevier}
}

@article{allam2012question,
  title={The question answering systems: A survey},
  author={Allam, Ali Mohamed Nabil and Haggag, Mohamed Hassan},
  journal={International Journal of Research and Reviews in Information Sciences (IJRRIS)},
  volume={2},
  number={3},
  year={2012}
}

@article{bouziane2015question,
  title={Question answering systems: survey and trends},
  author={Bouziane, Abdelghani and Bouchiha, Djelloul and Doumi, Noureddine and Malki, Mimoun},
  journal={Procedia Computer Science},
  volume={73},
  pages={366--375},
  year={2015},
  publisher={Elsevier}
}

@article{mishra2016survey,
  title={A survey on question answering systems with classification},
  author={Mishra, Amit and Jain, Sanjay Kumar},
  journal={Journal of King Saud University-Computer and Information Sciences},
  volume={28},
  number={3},
  pages={345--361},
  year={2016},
  publisher={Elsevier}
}

@article{hoffner2017survey,
  title={Survey on challenges of question answering in the semantic web},
  author={H{\"o}ffner, Konrad and Walter, Sebastian and Marx, Edgard and Usbeck, Ricardo and Lehmann, Jens and Ngonga Ngomo, Axel-Cyrille},
  journal={Semantic Web},
  volume={8},
  number={6},
  pages={895--920},
  year={2017},
  publisher={IOS Press}
}

@article{zhu2021retrieving,
  title={Retrieving and reading: A comprehensive survey on open-domain question answering},
  author={Zhu, Fengbin and Lei, Wenqiang and Wang, Chao and Zheng, Jianming and Poria, Soujanya and Chua, Tat-Seng},
  journal={arXiv preprint arXiv:2101.00774},
  year={2021}
}

@article{diefenbach2018core,
  title={Core techniques of question answering systems over knowledge bases: a survey},
  author={Diefenbach, Dennis and Lopez, Vanessa and Singh, Kamal and Maret, Pierre},
  journal={Knowledge and Information systems},
  volume={55},
  number={3},
  pages={529--569},
  year={2018},
  publisher={Springer}
}

@article{soares2020literature,
  title={A literature review on question answering techniques, paradigms and systems},
  author={Soares, Marco Antonio Calijorne and Parreiras, Fernando Silva},
  journal={Journal of King Saud University-Computer and Information Sciences},
  volume={32},
  number={6},
  pages={635--646},
  year={2020},
  publisher={Elsevier}
}

@article{dimitrakis2020survey,
  title={A survey on question answering systems over linked data and documents},
  author={Dimitrakis, Eleftherios and Sgontzos, Konstantinos and Tzitzikas, Yannis},
  journal={Journal of intelligent information systems},
  volume={55},
  number={2},
  pages={233--259},
  year={2020},
  publisher={Springer}
}

@article{weiss2021extending,
  title={Extending Multi-Text Sentence Fusion Resources via Pyramid Annotations},
  author={Weiss, Daniela Brook and Roit, Paul and Ernst, Ori and Dagan, Ido},
  journal={arXiv preprint arXiv:2110.04517},
  year={2021}
}

@inproceedings{lebanoff2019analyzing,
  title={Analyzing Sentence Fusion in Abstractive Summarization},
  author={Lebanoff, Logan and Muchovej, John and Dernoncourt, Franck and Kim, Doo Soon and Kim, Seokhwan and Chang, Walter and Liu, Fei},
  booktitle={Proceedings of the 2nd Workshop on New Frontiers in Summarization},
  pages={104--110},
  year={2019}
}

@inproceedings{geva2019discofuse,
  title={DiscoFuse: A Large-Scale Dataset for Discourse-Based Sentence Fusion},
  author={Geva, Mor and Malmi, Eric and Szpektor, Idan and Berant, Jonathan},
  booktitle={Proceedings of the 2019 Conference of the North American Chapter of the Association for Computational Linguistics: Human Language Technologies, Volume 1 (Long and Short Papers)},
  pages={3443--3455},
  year={2019}
}

@inproceedings{nayeem2018abstractive,
  title={Abstractive unsupervised multi-document summarization using paraphrastic sentence fusion},
  author={Nayeem, Mir Tafseer and Fuad, Tanvir Ahmed and Chali, Yllias},
  booktitle={Proceedings of the 27th International Conference on Computational Linguistics},
  pages={1191--1204},
  year={2018}
}

@inproceedings{goldstein2000multi,
  title={Multi-document summarization by sentence extraction},
  author={Goldstein, Jade and Mittal, Vibhu O and Carbonell, Jaime G and Kantrowitz, Mark},
  booktitle={NAACL-ANLP 2000 workshop: automatic summarization},
  year={2000}
}

@inproceedings{haghighi2009exploring,
  title={Exploring content models for multi-document summarization},
  author={Haghighi, Aria and Vanderwende, Lucy},
  booktitle={Proceedings of human language technologies: The 2009 annual conference of the North American Chapter of the Association for Computational Linguistics},
  pages={362--370},
  year={2009}
}

@inproceedings{barzilay1999information,
  title={Information fusion in the context of multi-document summarization},
  author={Barzilay, Regina and McKeown, Kathleen and Elhadad, Michael},
  booktitle={Proceedings of the 37th annual meeting of the Association for Computational Linguistics},
  pages={550--557},
  year={1999}
}

@article{ma2020multi,
  title={Multi-document summarization via deep learning techniques: A survey},
  author={Ma, Congbo and Zhang, Wei Emma and Guo, Mingyu and Wang, Hu and Sheng, Quan Z},
  journal={arXiv preprint arXiv:2011.04843},
  year={2020}
}

@inproceedings{yan2011evolutionary,
  title={Evolutionary timeline summarization: a balanced optimization framework via iterative substitution},
  author={Yan, Rui and Wan, Xiaojun and Otterbacher, Jahna and Kong, Liang and Li, Xiaoming and Zhang, Yan},
  booktitle={Proceedings of the 34th international ACM SIGIR conference on Research and development in Information Retrieval},
  pages={745--754},
  year={2011}
}

@article{ghalandari2020examining,
  title={Examining the state-of-the-art in news timeline summarization},
  author={Ghalandari, Demian Gholipour and Ifrim, Georgiana},
  journal={arXiv preprint arXiv:2005.10107},
  year={2020}
}

@inproceedings{steen2019abstractive,
  title={Abstractive timeline summarization},
  author={Steen, Julius and Markert, Katja},
  booktitle={Proceedings of the 2nd Workshop on New Frontiers in Summarization},
  pages={21--31},
  year={2019}
}

@Inbook{Melo2013,
author="Melo, Francisco",
editor="Dubitzky, Werner
and Wolkenhauer, Olaf
and Cho, Kwang-Hyun
and Yokota, Hiroki",
title="Area under the ROC Curve",
bookTitle="Encyclopedia of Systems Biology",
year="2013",
publisher="Springer New York",
address="New York, NY",
pages="38--39",
isbn="978-1-4419-9863-7",
doi="10.1007/978-1-4419-9863-7_209",
url="https://doi.org/10.1007/978-1-4419-9863-7_209"
}

@article{10.1145/582415.582418,
author = {J\"{a}rvelin, Kalervo and Kek\"{a}l\"{a}inen, Jaana},
title = {Cumulated Gain-Based Evaluation of IR Techniques},
year = {2002},
issue_date = {October 2002},
publisher = {Association for Computing Machinery},
address = {New York, NY, USA},
volume = {20},
number = {4},
issn = {1046-8188},
url = {https://doi.org/10.1145/582415.582418},
doi = {10.1145/582415.582418},
journal = {ACM Trans. Inf. Syst.},
month = {oct},
pages = {422–446},
numpages = {25}
}

@inproceedings{yadav-etal-2019-alignment,
    title = "Alignment over Heterogeneous Embeddings for Question Answering",
    author = "Yadav, Vikas  and
      Bethard, Steven  and
      Surdeanu, Mihai",
    booktitle = "Proceedings of the 2019 Conference of the North {A}merican Chapter of the Association for Computational Linguistics: Human Language Technologies, Volume 1 (Long and Short Papers)",
    month = jun,
    year = "2019",
    address = "Minneapolis, Minnesota",
    publisher = "Association for Computational Linguistics",
    url = "https://aclanthology.org/N19-1274",
    doi = "10.18653/v1/N19-1274",
    pages = "2681--2691",
}

@inproceedings{zhao-etal-2018-paragraph,
    title = "Paragraph-level Neural Question Generation with Maxout Pointer and Gated Self-attention Networks",
    author = "Zhao, Yao  and
      Ni, Xiaochuan  and
      Ding, Yuanyuan  and
      Ke, Qifa",
    booktitle = "Proceedings of the 2018 Conference on Empirical Methods in Natural Language Processing",
    month = oct # "-" # nov,
    year = "2018",
    address = "Brussels, Belgium",
    publisher = "Association for Computational Linguistics",
    url = "https://aclanthology.org/D18-1424",
    doi = "10.18653/v1/D18-1424",
    pages = "3901--3910"
}

@inproceedings{welleck2019neural,
  title={Neural Text Generation With Unlikelihood Training},
  author={Welleck, Sean and Kulikov, Ilia and Roller, Stephen and Dinan, Emily and Cho, Kyunghyun and Weston, Jason},
  booktitle={International Conference on Learning Representations},
  year={2019}
}

@techreport{rumelhart1985learning,
  title={Learning internal representations by error propagation},
  author={Rumelhart, David E and Hinton, Geoffrey E and Williams, Ronald J},
  year={1985},
  institution={California Univ San Diego La Jolla Inst for Cognitive Science}
}

@article{schlichtkrull2017modeling,
  title={Modeling Relational Data with Graph Convolutional Networks (2017)},
  author={Schlichtkrull, Michael and Kipf, Thomas N and Bloem, Peter and Van Den Berg, Rianne and Titov, Ivan and Welling, Max},
  journal={Preprint},
  year={2017}
}

@inproceedings{kovaleva-etal-2019-revealing,
    title = "Revealing the Dark Secrets of {BERT}",
    author = "Kovaleva, Olga  and
      Romanov, Alexey  and
      Rogers, Anna  and
      Rumshisky, Anna",
    booktitle = "Proceedings of the 2019 Conference on Empirical Methods in Natural Language Processing and the 9th International Joint Conference on Natural Language Processing (EMNLP-IJCNLP)",
    month = nov,
    year = "2019",
    address = "Hong Kong, China",
    publisher = "Association for Computational Linguistics",
    url = "https://aclanthology.org/D19-1445",
    doi = "10.18653/v1/D19-1445",
    pages = "4365--4374",
}

@inproceedings{clark2019sentence,
  title={Sentence mover’s similarity: Automatic evaluation for multi-sentence texts},
  author={Clark, Elizabeth and Celikyilmaz, Asli and Smith, Noah A},
  booktitle={Proceedings of the 57th Annual Meeting of the Association for Computational Linguistics},
  pages={2748--2760},
  year={2019}
}

@article{newsqa,
  author    = {Adam Trischler and
               Tong Wang and
               Xingdi Yuan and
               Justin Harris and
               Alessandro Sordoni and
               Philip Bachman and
               Kaheer Suleman},
  title     = {NewsQA: {A} Machine Comprehension Dataset},
  journal   = {CoRR},
  volume    = {abs/1611.09830},
  year      = {2016},
  url       = {http://arxiv.org/abs/1611.09830},
  eprinttype = {arXiv},
  eprint    = {1611.09830},
  bibsource = {dblp computer science bibliography, https://dblp.org}
}

@article{quac,
  author    = {Eunsol Choi and
               He He and
               Mohit Iyyer and
               Mark Yatskar and
               Wen{-}tau Yih and
               Yejin Choi and
               Percy Liang and
               Luke Zettlemoyer},
  title     = {QuAC : Question Answering in Context},
  journal   = {CoRR},
  volume    = {abs/1808.07036},
  year      = {2018},
  url       = {http://arxiv.org/abs/1808.07036},
  eprinttype = {arXiv},
  eprint    = {1808.07036},
  bibsource = {dblp computer science bibliography, https://dblp.org}
}

@article{gnn,
  author    = {Thomas N. Kipf and
               Max Welling},
  title     = {Semi-Supervised Classification with Graph Convolutional Networks},
  journal   = {CoRR},
  volume    = {abs/1609.02907},
  year      = {2016},
  url       = {http://arxiv.org/abs/1609.02907},
  eprinttype = {arXiv},
  eprint    = {1609.02907},
  bibsource = {dblp computer science bibliography, https://dblp.org}
}

@article{gcn,
  author    = {Thomas N. Kipf and
               Max Welling},
  title     = {Semi-Supervised Classification with Graph Convolutional Networks},
  journal   = {CoRR},
  volume    = {abs/1609.02907},
  year      = {2016},
  url       = {http://arxiv.org/abs/1609.02907},
  eprinttype = {arXiv},
  eprint    = {1609.02907},
  bibsource = {dblp computer science bibliography, https://dblp.org}
}

@article{qg4edu,
    author = {Kurdi, Ghader and Leo, Jared and Parsia, Bijan and Sattler, Uli and Al-Emari, Salam},
    year = {2019},
    month = {11},
    pages = {},
    title = {A Systematic Review of Automatic Question Generation for Educational Purposes},
    volume = {30},
    journal = {International Journal of Artificial Intelligence in Education},
    doi = {10.1007/s40593-019-00186-y}
}

@inproceedings{easyques1,
    title = "Easy Questions First? A Case Study on Curriculum Learning for Question Answering",
    author = "Sachan, Mrinmaya  and Xing, Eric",
    booktitle = "Proceedings of the 54th Annual Meeting of the Association for Computational Linguistics (Volume 1: Long Papers)",
    month = aug,
    year = "2016",
    address = "Berlin, Germany",
    publisher = "Association for Computational Linguistics",
    url = "https://aclanthology.org/P16-1043",
    doi = "10.18653/v1/P16-1043",
    pages = "453--463",
}

@article{dqg,
  author    = {Yifan Gao and
               Jianan Wang and
               Lidong Bing and
               Irwin King and
               Michael R. Lyu},
  title     = {Difficulty Controllable Question Generation for Reading Comprehension},
  journal   = {CoRR},
  volume    = {abs/1807.03586},
  year      = {2018},
  url       = {http://arxiv.org/abs/1807.03586},
  eprinttype = {arXiv},
  eprint    = {1807.03586},
  bibsource = {dblp computer science bibliography, https://dblp.org}
}

@inproceedings{pgn2,
    title = "Pointing the Unknown Words",
    author = "Gulcehre, Caglar  and
      Ahn, Sungjin  and
      Nallapati, Ramesh  and
      Zhou, Bowen  and
      Bengio, Yoshua",
    booktitle = "Proceedings of the 54th Annual Meeting of the Association for Computational Linguistics (Volume 1: Long Papers)",
    month = aug,
    year = "2016",
    address = "Berlin, Germany",
    publisher = "Association for Computational Linguistics",
    url = "https://aclanthology.org/P16-1014",
    doi = "10.18653/v1/P16-1014",
    pages = "140--149",
}

@inproceedings{attention4nmt,
    title = "Effective Approaches to Attention-based Neural Machine Translation",
    author = "Luong, Thang  and
      Pham, Hieu  and
      Manning, Christopher D.",
    booktitle = "Proceedings of the 2015 Conference on Empirical Methods in Natural Language Processing",
    month = sep,
    year = "2015",
    address = "Lisbon, Portugal",
    publisher = "Association for Computational Linguistics",
    url = "https://aclanthology.org/D15-1166",
    doi = "10.18653/v1/D15-1166",
    pages = "1412--1421",
}

@InProceedings{scst,
author = {Rennie, Steven J. and Marcheret, Etienne and Mroueh, Youssef and Ross, Jerret and Goel, Vaibhava},
title = {Self-Critical Sequence Training for Image Captioning},
booktitle = {Proceedings of the IEEE Conference on Computer Vision and Pattern Recognition (CVPR)},
month = {July},
year = {2017}
}

@article{reinforce,
author = {Williams, Ronald J.},
title = {Simple Statistical Gradient-Following Algorithms for Connectionist Reinforcement Learning},
year = {1992},
issue_date = {May 1992},
publisher = {Kluwer Academic Publishers},
address = {USA},
volume = {8},
number = {3–4},
issn = {0885-6125},
url = {https://doi.org/10.1007/BF00992696},
doi = {10.1007/BF00992696},
journal = {Mach. Learn.},
month = {may},
pages = {229–256},
numpages = {28}
}

@inproceedings{qg4online,
    title = "Generating Natural Language Questions to Support Learning On-Line",
    author = "Lindberg, David  and
      Popowich, Fred  and
      Nesbit, John  and
      Winne, Phil",
    booktitle = "Proceedings of the 14th {E}uropean Workshop on Natural Language Generation",
    month = aug,
    year = "2013",
    address = "Sofia, Bulgaria",
    publisher = "Association for Computational Linguistics",
    url = "https://aclanthology.org/W13-2114",
    pages = "105--114",
}

@inproceedings{localKGS2Smulti,
    title = "Using Local Knowledge Graph Construction to Scale {S}eq2{S}eq Models to Multi-Document Inputs",
    author = "Fan, Angela  and
      Gardent, Claire  and
      Braud, Chlo{\'e}  and
      Bordes, Antoine",
    booktitle = "Proceedings of the 2019 Conference on Empirical Methods in Natural Language Processing and the 9th International Joint Conference on Natural Language Processing (EMNLP-IJCNLP)",
    month = nov,
    year = "2019",
    address = "Hong Kong, China",
    publisher = "Association for Computational Linguistics",
    url = "https://aclanthology.org/D19-1428",
    doi = "10.18653/v1/D19-1428",
    pages = "4186--4196"
}

@inproceedings{qbleu,
    title = "Towards a Better Metric for Evaluating Question Generation Systems",
    author = "Nema, Preksha  and
      Khapra, Mitesh M.",
    booktitle = "Proceedings of the 2018 Conference on Empirical Methods in Natural Language Processing",
    month = oct # "-" # nov,
    year = "2018",
    address = "Brussels, Belgium",
    publisher = "Association for Computational Linguistics",
    url = "https://aclanthology.org/D18-1429",
    doi = "10.18653/v1/D18-1429",
    pages = "3950--3959",
}

@article{gleu,
  author    = {Yonghui Wu and Mike Schuster and Zhifeng Chen and Quoc V. Le and Mohammad Norouzi and Wolfgang Macherey and Maxim Krikun and Yuan Cao and Qin Gao and Klaus Macherey and Jeff Klingner and Apurva Shah and Melvin Johnson and Xiaobing Liu and  Lukasz Kaiser and Stephan Gouws and Yoshikiyo Kato and Taku Kudo and Hideto Kazawa and Keith Stevens and George Kurian and Nishant Patil and Wei Wang and Cliff Young and Jason Smith and Dason Riesa and Alex Rudnick and Oriol Vinyals and Greg Corrado and Macduff Hughes and Jeffrey Dean},
  title     = {Google's Neural Machine Translation System: Bridging the Gap between
               Human and Machine Translation},
  journal   = {CoRR},
  volume    = {abs/1609.08144},
  year      = {2016},
  url       = {http://arxiv.org/abs/1609.08144},
  eprinttype = {arXiv},
  eprint    = {1609.08144},
  bibsource = {dblp computer science bibliography, https://dblp.org}
}

@inproceedings{dua-etal-2019-drop,
    title = "{DROP}: A Reading Comprehension Benchmark Requiring Discrete Reasoning Over Paragraphs",
    author = "Dua, Dheeru  and
      Wang, Yizhong  and
      Dasigi, Pradeep  and
      Stanovsky, Gabriel  and
      Singh, Sameer  and
      Gardner, Matt",
    booktitle = "Proceedings of the 2019 Conference of the North {A}merican Chapter of the Association for Computational Linguistics: Human Language Technologies, Volume 1 (Long and Short Papers)",
    month = jun,
    year = "2019",
    address = "Minneapolis, Minnesota",
    publisher = "Association for Computational Linguistics",
    url = "https://aclanthology.org/N19-1246",
    doi = "10.18653/v1/N19-1246",
    pages = "2368--2378",
}

@inproceedings{complexwebques,
    title = "The Web as a Knowledge-Base for Answering Complex Questions",
    author = "Talmor, Alon  and
      Berant, Jonathan",
    booktitle = "Proceedings of the 2018 Conference of the North {A}merican Chapter of the Association for Computational Linguistics: Human Language Technologies, Volume 1 (Long Papers)",
    month = jun,
    year = "2018",
    address = "New Orleans, Louisiana",
    publisher = "Association for Computational Linguistics",
    url = "https://aclanthology.org/N18-1059",
    doi = "10.18653/v1/N18-1059",
    pages = "641--651",
}

@inproceedings{maxout-qg,
    title = "Paragraph-level Neural Question Generation with Maxout Pointer and Gated Self-attention Networks",
    author = "Zhao, Yao  and
      Ni, Xiaochuan  and
      Ding, Yuanyuan  and
      Ke, Qifa",
    booktitle = "Proceedings of the 2018 Conference on Empirical Methods in Natural Language Processing",
    month = oct # "-" # nov,
    year = "2018",
    address = "Brussels, Belgium",
    publisher = "Association for Computational Linguistics",
    url = "https://aclanthology.org/D18-1424",
    doi = "10.18653/v1/D18-1424",
    pages = "3901--3910",
}

@inproceedings{ansagnosticQG,
    title = "Self-Attention Architectures for Answer-Agnostic Neural Question Generation",
    author = "Scialom, Thomas  and
      Piwowarski, Benjamin  and
      Staiano, Jacopo",
    booktitle = "Proceedings of the 57th Annual Meeting of the Association for Computational Linguistics",
    month = jul,
    year = "2019",
    address = "Florence, Italy",
    publisher = "Association for Computational Linguistics",
    url = "https://aclanthology.org/P19-1604",
    doi = "10.18653/v1/P19-1604",
    pages = "6027--6032",
}

@inproceedings{qg4rc,
    title = "Learning to Ask: Neural Question Generation for Reading Comprehension",
    author = "Du, Xinya  and Shao, Junru and Cardie, Claire",
    booktitle = "Proceedings of the 55th Annual Meeting of the Association for Computational Linguistics (Volume 1: Long Papers)",
    month = jul,
    year = "2017",
    address = "Vancouver, Canada",
    publisher = "Association for Computational Linguistics",
    url = "https://aclanthology.org/P17-1123",
    doi = "10.18653/v1/P17-1123",
    pages = "1342--1352",
}

@article{qaqgdual,
  author    = {Duyu Tang and
               Nan Duan and
               Tao Qin and
               Ming Zhou},
  title     = {Question Answering and Question Generation as Dual Tasks},
  journal   = {CoRR},
  volume    = {abs/1706.02027},
  year      = {2017},
  url       = {http://arxiv.org/abs/1706.02027},
  eprinttype = {arXiv},
  eprint    = {1706.02027},
  bibsource = {dblp computer science bibliography, https://dblp.org}
}

@inproceedings{teachingqg,
  title     = {Teaching Machines to Ask Questions},
  author    = {Kaichun Yao and Libo Zhang and Tiejian Luo and Lili Tao and Yanjun Wu},
  booktitle = {Proceedings of the Twenty-Seventh International Joint Conference on
               Artificial Intelligence, {IJCAI-18}},
  publisher = {International Joint Conferences on Artificial Intelligence Organization},             
  pages     = {4546--4552},
  year      = {2018},
  month     = {7},
  doi       = {10.24963/ijcai.2018/632},
  url       = {https://doi.org/10.24963/ijcai.2018/632},
}

@inproceedings{sentence-deeprl,
    title = "Sentence Simplification with Deep Reinforcement Learning",
    author = "Zhang, Xingxing  and
      Lapata, Mirella",
    booktitle = "Proceedings of the 2017 Conference on Empirical Methods in Natural Language Processing",
    month = sep,
    year = "2017",
    address = "Copenhagen, Denmark",
    publisher = "Association for Computational Linguistics",
    url = "https://aclanthology.org/D17-1062",
    doi = "10.18653/v1/D17-1062",
    pages = "584--594",
}

@inproceedings{adversarial-dialogue,
    title = "Adversarial Learning for Neural Dialogue Generation",
    author = "Li, Jiwei  and
      Monroe, Will  and
      Shi, Tianlin  and
      Jean, S{\'e}bastien  and
      Ritter, Alan  and
      Jurafsky, Dan",
    booktitle = "Proceedings of the 2017 Conference on Empirical Methods in Natural Language Processing",
    month = sep,
    year = "2017",
    address = "Copenhagen, Denmark",
    publisher = "Association for Computational Linguistics",
    url = "https://aclanthology.org/D17-1230",
    doi = "10.18653/v1/D17-1230",
    pages = "2157--2169",
}

@inproceedings{gu-incorporating-copy-s2s,
    title = "Incorporating Copying Mechanism in Sequence-to-Sequence Learning",
    author = "Gu, Jiatao  and
      Lu, Zhengdong  and
      Li, Hang  and
      Li, Victor O.K.",
    booktitle = "Proceedings of the 54th Annual Meeting of the Association for Computational Linguistics (Volume 1: Long Papers)",
    month = aug,
    year = "2016",
    address = "Berlin, Germany",
    publisher = "Association for Computational Linguistics",
    url = "https://aclanthology.org/P16-1154",
    doi = "10.18653/v1/P16-1154",
    pages = "1631--1640",
}

@inproceedings{cnn,
    title = "Abstractive Text Summarization using Sequence-to-sequence {RNN}s and Beyond",
    author = "Nallapati, Ramesh  and
      Zhou, Bowen  and
      dos Santos, Cicero  and
      Gulcehre, Caglar  and
      Xiang, Bing",
    editor = "Riezler, Stefan  and
      Goldberg, Yoav",
    booktitle = "Proceedings of the 20th {SIGNLL} Conference on Computational Natural Language Learning",
    month = aug,
    year = "2016",
    address = "Berlin, Germany",
    publisher = "Association for Computational Linguistics",
    url = "https://aclanthology.org/K16-1028",
    doi = "10.18653/v1/K16-1028",
    pages = "280--290",
}

@article{bart,
  author    = {Mike Lewis and
               Yinhan Liu and
               Naman Goyal and
               Marjan Ghazvininejad and
               Abdelrahman Mohamed and
               Omer Levy and
               Veselin Stoyanov and
               Luke Zettlemoyer},
  title     = {{BART:} Denoising Sequence-to-Sequence Pre-training for Natural Language
               Generation, Translation, and Comprehension},
  journal   = {CoRR},
  volume    = {abs/1910.13461},
  year      = {2019},
  url       = {http://arxiv.org/abs/1910.13461},
  eprinttype = {arXiv},
  eprint    = {1910.13461},
  bibsource = {dblp computer science bibliography, https://dblp.org}
}

@article{nlg-tablegen,
  author    = {Wenhu Chen and
               Jianshu Chen and
               Yu Su and
               Zhiyu Chen and
               William Yang Wang},
  title     = {Logical Natural Language Generation from Open-Domain Tables},
  journal   = {CoRR},
  volume    = {abs/2004.10404},
  year      = {2020},
  url       = {https://arxiv.org/abs/2004.10404},
  eprinttype = {arXiv},
  eprint    = {2004.10404},
  bibsource = {dblp computer science bibliography, https://dblp.org}
}

@article{qa2nli,
  author    = {Dorottya Demszky and
               Kelvin Guu and
               Percy Liang},
  title     = {Transforming Question Answering Datasets Into Natural Language Inference
               Datasets},
  journal   = {CoRR},
  volume    = {abs/1809.02922},
  year      = {2018},
  url       = {http://arxiv.org/abs/1809.02922},
  eprinttype = {arXiv},
  eprint    = {1809.02922},
  bibsource = {dblp computer science bibliography, https://dblp.org}
}

@article{t5,
  author    = {Colin Raffel and
               Noam Shazeer and
               Adam Roberts and
               Katherine Lee and
               Sharan Narang and
               Michael Matena and
               Yanqi Zhou and
               Wei Li and
               Peter J. Liu},
  title     = {Exploring the Limits of Transfer Learning with a Unified Text-to-Text
               Transformer},
  journal   = {CoRR},
  volume    = {abs/1910.10683},
  year      = {2019},
  url       = {http://arxiv.org/abs/1910.10683},
  eprinttype = {arXiv},
  eprint    = {1910.10683},
  bibsource = {dblp computer science bibliography, https://dblp.org}
}

@inproceedings{semi-markov,
  title={Recurrent Hidden Semi-Markov Model},
  author={Hanjun Dai and Bo Dai and Yanming Zhang and Shuang Li and Le Song},
  booktitle={ICLR},
  year={2017}
}

@article{qa-ie,
  author    = {Emanuela Boros and
               Jose G. Moreno and
               Antoine Doucet},
  title     = {Event Detection as Question Answering with Entity Information},
  journal   = {CoRR},
  volume    = {abs/2104.06969},
  year      = {2021},
  url       = {https://arxiv.org/abs/2104.06969},
  eprinttype = {arXiv},
  eprint    = {2104.06969},
  bibsource = {dblp computer science bibliography, https://dblp.org}
}

@article{lipton2015critical,
  title={A critical review of recurrent neural networks for sequence learning},
  author={Lipton, Zachary C and Berkowitz, John and Elkan, Charles},
  journal={arXiv preprint arXiv:1506.00019},
  year={2015}
}

@article{vaswani2017attention,
  title={Attention is all you need},
  author={Vaswani, Ashish and Shazeer, Noam and Parmar, Niki and Uszkoreit, Jakob and Jones, Llion and Gomez, Aidan N and Kaiser, {\L}ukasz and Polosukhin, Illia},
  journal={Advances in neural information processing systems},
  volume={30},
  year={2017}
}

@article{hochreiter1997long,
  title={Long short-term memory},
  author={Hochreiter, Sepp and Schmidhuber, J{\"u}rgen},
  journal={Neural computation},
  volume={9},
  number={8},
  pages={1735--1780},
  year={1997},
  publisher={MIT press}
}

@article{see2017get,
  title={Get to the point: Summarization with pointer-generator networks},
  author={See, Abigail and Liu, Peter J and Manning, Christopher D},
  journal={arXiv preprint arXiv:1704.04368},
  year={2017}
}

@article{wu2020comprehensive,
  title={A comprehensive survey on graph neural networks},
  author={Wu, Zonghan and Pan, Shirui and Chen, Fengwen and Long, Guodong and Zhang, Chengqi and Philip, S Yu},
  journal={IEEE transactions on neural networks and learning systems},
  volume={32},
  number={1},
  pages={4--24},
  year={2020},
  publisher={IEEE}
}

@article{LIN2022111,
title = {A survey of transformers},
journal = {AI Open},
volume = {3},
pages = {111-132},
year = {2022},
issn = {2666-6510},
doi = {https://doi.org/10.1016/j.aiopen.2022.10.001},
url = {https://www.sciencedirect.com/science/article/pii/S2666651022000146},
author = {Tianyang Lin and Yuxin Wang and Xiangyang Liu and Xipeng Qiu},
}

@misc{folio,
      title={FOLIO: Natural Language Reasoning with First-Order Logic}, 
      author={Simeng Han and Hailey Schoelkopf and Yilun Zhao and Zhenting Qi and Martin Riddell and Luke Benson and Lucy Sun and Ekaterina Zubova and Yujie Qiao and Matthew Burtell and David Peng and Jonathan Fan and Yixin Liu and Brian Wong and Malcolm Sailor and Ansong Ni and Linyong Nan and Jungo Kasai and Tao Yu and Rui Zhang and Shafiq Joty and Alexander R. Fabbri and Wojciech Kryscinski and Xi Victoria Lin and Caiming Xiong and Dragomir Radev},
      year={2022},
      eprint={2209.00840},
      archivePrefix={arXiv},
      primaryClass={cs.CL}
}

@misc{gpt-qa2,
      title={Decomposed Prompting: A Modular Approach for Solving Complex Tasks}, 
      author={Tushar Khot and Harsh Trivedi and Matthew Finlayson and Yao Fu and Kyle Richardson and Peter Clark and Ashish Sabharwal},
      year={2023},
      eprint={2210.02406},
      archivePrefix={arXiv},
      primaryClass={cs.CL}
}

@inproceedings{gpt-qa4,
    title={Least-to-Most Prompting Enables Complex Reasoning in Large Language Models},
    author={Denny Zhou and Nathanael Sch{\"a}rli and Le Hou and Jason Wei and Nathan Scales and Xuezhi Wang and Dale Schuurmans and Claire Cui and Olivier Bousquet and Quoc V Le and Ed H. Chi},
    booktitle={The Eleventh International Conference on Learning Representations },
    year={2023},
    url={https://openreview.net/forum?id=WZH7099tgfM}
    }

@misc{gpt-qa5,
      title={STaR: Bootstrapping Reasoning With Reasoning}, 
      author={Eric Zelikman and Yuhuai Wu and Jesse Mu and Noah D. Goodman},
      year={2022},
      eprint={2203.14465},
      archivePrefix={arXiv},
      primaryClass={cs.LG}
}

@misc{gpt-qa6,
  author = {Wang, Ben},
  title = {{Mesh-Transformer-JAX: Model-Parallel Implementation of Transformer Language Model with JAX}},
  howpublished = {\url{https://github.com/kingoflolz/mesh-transformer-jax}},
  year = 2021,
  month = May
}

@inproceedings{saparov,
    title={Language Models Are Greedy Reasoners: A Systematic Formal Analysis of Chain-of-Thought},
    author={Abulhair Saparov and He He},
    booktitle={The Eleventh International Conference on Learning Representations },
    year={2023},
    url={https://openreview.net/forum?id=qFVVBzXxR2V}
    }

@misc{gptsara,
      title={Can GPT-3 Perform Statutory Reasoning?}, 
      author={Andrew Blair-Stanek and Nils Holzenberger and Benjamin Van Durme},
      year={2023},
      eprint={2302.06100},
      archivePrefix={arXiv},
      primaryClass={cs.CL}
}

@inproceedings{finqa,
    title = "{F}in{QA}: A Dataset of Numerical Reasoning over Financial Data",
    author = "Chen, Zhiyu  and
      Chen, Wenhu  and
      Smiley, Charese  and
      Shah, Sameena  and
      Borova, Iana  and
      Langdon, Dylan  and
      Moussa, Reema  and
      Beane, Matt  and
      Huang, Ting-Hao  and
      Routledge, Bryan  and
      Wang, William Yang",
    booktitle = "Proceedings of the 2021 Conference on Empirical Methods in Natural Language Processing",
    month = nov,
    year = "2021",
    address = "Online and Punta Cana, Dominican Republic",
    publisher = "Association for Computational Linguistics",
    url = "https://aclanthology.org/2021.emnlp-main.300",
    doi = "10.18653/v1/2021.emnlp-main.300",
    pages = "3697--3711",
}

@inproceedings{gpt,
 author = {Brown, Tom and Mann, Benjamin and Ryder, Nick and Subbiah, Melanie and Kaplan, Jared D and Dhariwal, Prafulla and Neelakantan, Arvind and Shyam, Pranav and Sastry, Girish and Askell, Amanda and Agarwal, Sandhini and Herbert-Voss, Ariel and Krueger, Gretchen and Henighan, Tom and Child, Rewon and Ramesh, Aditya and Ziegler, Daniel and Wu, Jeffrey and Winter, Clemens and Hesse, Chris and Chen, Mark and Sigler, Eric and Litwin, Mateusz and Gray, Scott and Chess, Benjamin and Clark, Jack and Berner, Christopher and McCandlish, Sam and Radford, Alec and Sutskever, Ilya and Amodei, Dario},
 booktitle = {Advances in Neural Information Processing Systems},
 editor = {H. Larochelle and M. Ranzato and R. Hadsell and M.F. Balcan and H. Lin},
 pages = {1877--1901},
 publisher = {Curran Associates, Inc.},
 title = {Language Models are Few-Shot Learners},
 url = {https://proceedings.neurips.cc/paper_files/paper/2020/file/1457c0d6bfcb4967418bfb8ac142f64a-Paper.pdf},
 volume = {33},
 year = {2020}
}

@misc{cot,
      title={Chain-of-Thought Prompting Elicits Reasoning in Large Language Models}, 
      author={Jason Wei and Xuezhi Wang and Dale Schuurmans and Maarten Bosma and Brian Ichter and Fei Xia and Ed Chi and Quoc Le and Denny Zhou},
      year={2023},
      eprint={2201.11903},
      archivePrefix={arXiv},
      primaryClass={cs.CL}
}

@misc{gpt-success-qa,
      title={Evaluation of ChatGPT as a Question Answering System for Answering Complex Questions}, 
      author={Yiming Tan and Dehai Min and Yu Li and Wenbo Li and Nan Hu and Yongrui Chen and Guilin Qi},
      year={2023},
      eprint={2303.07992},
      archivePrefix={arXiv},
      primaryClass={cs.CL}
}

@misc{gpt-success-mt,
      title={How Good Are GPT Models at Machine Translation? A Comprehensive Evaluation}, 
      author={Amr Hendy and Mohamed Abdelrehim and Amr Sharaf and Vikas Raunak and Mohamed Gabr and Hitokazu Matsushita and Young Jin Kim and Mohamed Afify and Hany Hassan Awadalla},
      year={2023},
      eprint={2302.09210},
      archivePrefix={arXiv},
      primaryClass={cs.CL}
}

@misc{gpt-success-entail,
      title={Entailment as Few-Shot Learner}, 
      author={Sinong Wang and Han Fang and Madian Khabsa and Hanzi Mao and Hao Ma},
      year={2021},
      eprint={2104.14690},
      archivePrefix={arXiv},
      primaryClass={cs.CL}
}

@article{fewshot,
    author = {Fei-Fei, Li and Fergus, Rob and Perona, Pietro},
    year = {2006},
    month = {05},
    pages = {594-611},
    title = {One-Shot Learning of Object Categories},
    volume = {28},
    journal = {IEEE transactions on pattern analysis and machine intelligence},
    doi = {10.1109/TPAMI.2006.79}
}

@misc{LL1,
      title={Learning to Decompose: Hypothetical Question Decomposition Based on Comparable Texts}, 
      author={Ben Zhou and Kyle Richardson and Xiaodong Yu and Dan Roth},
      year={2022},
      eprint={2210.16865},
      archivePrefix={arXiv},
      primaryClass={cs.CL}
}

@misc{LL2,
      title={Is a Question Decomposition Unit All We Need?}, 
      author={Pruthvi Patel and Swaroop Mishra and Mihir Parmar and Chitta Baral},
      year={2022},
      eprint={2205.12538},
      archivePrefix={arXiv},
      primaryClass={cs.CL}
}

@misc{LL3,
      title={Mastering the ABCDs of Complex Questions: Answer-Based Claim Decomposition for Fine-grained Self-Evaluation}, 
      author={Nishant Balepur and Jie Huang and Samraj Moorjani and Hari Sundaram and Kevin Chen-Chuan Chang},
      year={2023},
      eprint={2305.14750},
      archivePrefix={arXiv},
      primaryClass={cs.CL}
}

@misc{LL6,
      title={Verify-and-Edit: A Knowledge-Enhanced Chain-of-Thought Framework}, 
      author={Ruochen Zhao and Xingxuan Li and Shafiq Joty and Chengwei Qin and Lidong Bing},
      year={2023},
      eprint={2305.03268},
      archivePrefix={arXiv},
      primaryClass={cs.CL}
}

@misc{LL12,
      title={Interleaving Retrieval with Chain-of-Thought Reasoning for Knowledge-Intensive Multi-Step Questions}, 
      author={Harsh Trivedi and Niranjan Balasubramanian and Tushar Khot and Ashish Sabharwal},
      year={2023},
      eprint={2212.10509},
      archivePrefix={arXiv},
      primaryClass={cs.CL}
}

@inproceedings{LL13,
  title={Navigating the Fermi Multiverse: Assessing LLMs for Complex Multi-hop Queries},
  author={Rahgouy, Mostafa and Giglou, Hamed Babaei and Feng, Dongji and Rahgooy, Taher and Dozier, Gerry and Seals, Cheryl D},
  year={2023}
}

@misc{LL14,
      title={Exploiting Reasoning Chains for Multi-hop Science Question Answering}, 
      author={Weiwen Xu and Yang Deng and Huihui Zhang and Deng Cai and Wai Lam},
      year={2021},
      eprint={2109.02905},
      archivePrefix={arXiv},
      primaryClass={cs.CL}
}

@misc{LL15,
      title={Modeling Multi-hop Question Answering as Single Sequence Prediction}, 
      author={Semih Yavuz and Kazuma Hashimoto and Yingbo Zhou and Nitish Shirish Keskar and Caiming Xiong},
      year={2022},
      eprint={2205.09226},
      archivePrefix={arXiv},
      primaryClass={cs.CL}
}

@INPROCEEDINGS{LL16,
  author={Haji, Shosuke and Suekane, Keiichi and Sano, Hirofumi and Takagi, Tomohiro},
  booktitle={2023 IEEE 17th International Conference on Semantic Computing (ICSC)}, 
  title={Exploratory Inference Chain: Exploratorily Chaining Multi-hop Inferences with Large Language Models for Question-Answering}, 
  year={2023},
  volume={},
  number={},
  pages={175-182},
  doi={10.1109/ICSC56153.2023.00036}
}

@misc{LL17,
      title={S$^3$HQA: A Three-Stage Approach for Multi-hop Text-Table Hybrid Question Answering}, 
      author={Fangyu Lei and Xiang Li and Yifan Wei and Shizhu He and Yiming Huang and Jun Zhao and Kang Liu},
      year={2023},
      eprint={2305.11725},
      archivePrefix={arXiv},
      primaryClass={cs.CL}
}

@InProceedings{LL18,
    author="Li, Jiawei
    and Ren, Mucheng
    and Gao, Yang
    and Yang, Yizhe",
    editor="Sun, Maosong
    and Qin, Bing
    and Qiu, Xipeng
    and Jing, Jiang
    and Han, Xianpei
    and Rao, Gaoqi
    and Chen, Yubo",
    title="Ask to Understand: Question Generation for Multi-hop Question Answering",
    booktitle="Chinese Computational Linguistics",
    year="2023",
    publisher="Springer Nature Singapore",
    address="Singapore",
    pages="19--36",
    isbn="978-981-99-6207-5"
}

@misc{LL19,
      title={Leveraging Structured Information for Explainable Multi-hop Question Answering and Reasoning}, 
      author={Ruosen Li and Xinya Du},
      year={2023},
      eprint={2311.03734},
      archivePrefix={arXiv},
      primaryClass={cs.CL}
}

@misc{LL20,
      title={Drilling Down into the Discourse Structure with LLMs for Long Document Question Answering}, 
      author={Inderjeet Nair and Shwetha Somasundaram and Apoorv Saxena and Koustava Goswami},
      year={2023},
      eprint={2311.13565},
      archivePrefix={arXiv},
      primaryClass={cs.CL}
}

@inproceedings{LL21,
    title={Improving Multi-Hop Reasoning in {LLM}s by Learning from Rich Human Feedback},
    author={Nitish Joshi and Hanlin Zhang and Koushik Kalyanaraman and Zhiting Hu and Kumar Chellapilla and He He and Li Erran Li},
    booktitle={Neuro-Symbolic Learning and Reasoning in the era of Large Language Models},
    year={2023},
    url={https://openreview.net/forum?id=wxfqhp9bNR}
}

@inproceedings{LL23,
    title={Improving Multi-Hop Reasoning in {LLM}s by Learning from Rich Human Feedback},
    author={Nitish Joshi and Hanlin Zhang and Koushik Kalyanaraman and Zhiting Hu and Kumar Chellapilla and He He and Li Erran Li},
    booktitle={Neuro-Symbolic Learning and Reasoning in the era of Large Language Models},
    year={2023},
    url={https://openreview.net/forum?id=wxfqhp9bNR}
}

@misc{LL24,
      title={Don't Hallucinate, Abstain: Identifying LLM Knowledge Gaps via Multi-LLM Collaboration}, 
      author={Shangbin Feng and Weijia Shi and Yike Wang and Wenxuan Ding and Vidhisha Balachandran and Yulia Tsvetkov},
      year={2024},
      eprint={2402.00367},
      archivePrefix={arXiv},
      primaryClass={cs.CL}
}

@misc{LL25,
      title={Exploring Hybrid Question Answering via Program-based Prompting}, 
      author={Qi Shi and Han Cui and Haofeng Wang and Qingfu Zhu and Wanxiang Che and Ting Liu},
      year={2024},
      eprint={2402.10812},
      archivePrefix={arXiv},
      primaryClass={cs.CL}
}

@misc{LL26,
      title={GenDec: A robust generative Question-decomposition method for Multi-hop reasoning}, 
      author={Jian Wu and Linyi Yang and Yuliang Ji and Wenhao Huang and Börje F. Karlsson and Manabu Okumura},
      year={2024},
      eprint={2402.11166},
      archivePrefix={arXiv},
      primaryClass={cs.CL}
}

@misc{LL33,
      title={Self-Consistency Improves Chain of Thought Reasoning in Language Models}, 
      author={Xuezhi Wang and Jason Wei and Dale Schuurmans and Quoc Le and Ed Chi and Sharan Narang and Aakanksha Chowdhery and Denny Zhou},
      year={2023},
      eprint={2203.11171},
      archivePrefix={arXiv},
      primaryClass={cs.CL}
}

@misc{LL34,
      title={Self-Refine: Iterative Refinement with Self-Feedback}, 
      author={Aman Madaan and Niket Tandon and Prakhar Gupta and Skyler Hallinan and Luyu Gao and Sarah Wiegreffe and Uri Alon and Nouha Dziri and Shrimai Prabhumoye and Yiming Yang and Shashank Gupta and Bodhisattwa Prasad Majumder and Katherine Hermann and Sean Welleck and Amir Yazdanbakhsh and Peter Clark},
      year={2023},
      eprint={2303.17651},
      archivePrefix={arXiv},
      primaryClass={cs.CL}
}

@inproceedings{LL35,
  title     = {Interpretable AMR-Based Question Decomposition for Multi-hop Question Answering},
  author    = {Deng, Zhenyun and Zhu, Yonghua and Chen, Yang and Witbrock, Michael and Riddle, Patricia},
  booktitle = {Proceedings of the Thirty-First International Joint Conference on
               Artificial Intelligence, {IJCAI-22}},
  publisher = {International Joint Conferences on Artificial Intelligence Organization},
  editor    = {Lud De Raedt},
  pages     = {4093--4099},
  year      = {2022},
  month     = {7},
  note      = {Main Track},
  doi       = {10.24963/ijcai.2022/568},
  url       = {https://doi.org/10.24963/ijcai.2022/568},
}

@inproceedings{amr,
    title = "{A}bstract {M}eaning {R}epresentation for Sembanking",
    author = "Banarescu, Laura  and
      Bonial, Claire  and
      Cai, Shu  and
      Georgescu, Madalina  and
      Griffitt, Kira  and
      Hermjakob, Ulf  and
      Knight, Kevin  and
      Koehn, Philipp  and
      Palmer, Martha  and
      Schneider, Nathan",
    editor = "Pareja-Lora, Antonio  and
      Liakata, Maria  and
      Dipper, Stefanie",
    booktitle = "Proceedings of the 7th Linguistic Annotation Workshop and Interoperability with Discourse",
    month = aug,
    year = "2013",
    address = "Sofia, Bulgaria",
    publisher = "Association for Computational Linguistics",
    url = "https://aclanthology.org/W13-2322",
    pages = "178--186",
}

@misc{llm-survey,
      title={A Survey of Large Language Models}, 
      author={Wayne Xin Zhao and Kun Zhou and Junyi Li and Tianyi Tang and Xiaolei Wang and Yupeng Hou and Yingqian Min and Beichen Zhang and Junjie Zhang and Zican Dong and Yifan Du and Chen Yang and Yushuo Chen and Zhipeng Chen and Jinhao Jiang and Ruiyang Ren and Yifan Li and Xinyu Tang and Zikang Liu and Peiyu Liu and Jian-Yun Nie and Ji-Rong Wen},
      year={2023},
      eprint={2303.18223},
      archivePrefix={arXiv},
      primaryClass={cs.CL}
}

@misc{flant5,
  doi = {10.48550/ARXIV.2210.11416},
  url = {https://arxiv.org/abs/2210.11416},
  author = {Chung, Hyung Won and Hou, Le and Longpre, Shayne and Zoph, Barret and Tay, Yi and Fedus, William and Li, Eric and Wang, Xuezhi and Dehghani, Mostafa and Brahma, Siddhartha and Webson, Albert and Gu, Shixiang Shane and Dai, Zhuyun and Suzgun, Mirac and Chen, Xinyun and Chowdhery, Aakanksha and Narang, Sharan and Mishra, Gaurav and Yu, Adams and Zhao, Vincent and Huang, Yanping and Dai, Andrew and Yu, Hongkun and Petrov, Slav and Chi, Ed H. and Dean, Jeff and Devlin, Jacob and Roberts, Adam and Zhou, Denny and Le, Quoc V. and Wei, Jason},
  title = {Scaling Instruction-Finetuned Language Models},
  publisher = {arXiv},
  year = {2022},
  copyright = {Creative Commons Attribution 4.0 International}
}

@inproceedings{mavi,
    title = "Retrieval-Augmented Chain-of-Thought in Semi-structured Domains",
    author = "Mavi, Vaibhav  and
      Saparov, Abulhair  and
      Zhao, Chen",
    editor = "Preo{\textcommabelow{t}}iuc-Pietro, Daniel  and
      Goanta, Catalina  and
      Chalkidis, Ilias  and
      Barrett, Leslie  and
      Spanakis, Gerasimos (Jerry)  and
      Aletras, Nikolaos",
    booktitle = "Proceedings of the Natural Legal Language Processing Workshop 2023",
    month = dec,
    year = "2023",
    address = "Singapore",
    publisher = "Association for Computational Linguistics",
    url = "https://aclanthology.org/2023.nllp-1.18",
    doi = "10.18653/v1/2023.nllp-1.18",
    pages = "178--191",
}

@misc{deberta,
      title={DeBERTa: Decoding-enhanced BERT with Disentangled Attention}, 
      author={Pengcheng He and Xiaodong Liu and Jianfeng Gao and Weizhu Chen},
      year={2021},
      eprint={2006.03654},
      archivePrefix={arXiv},
      primaryClass={cs.CL}
}

@misc{moreinfo,
      title={Knowledge Card: Filling LLMs' Knowledge Gaps with Plug-in Specialized Language Models}, 
      author={Shangbin Feng and Weijia Shi and Yuyang Bai and Vidhisha Balachandran and Tianxing He and Yulia Tsvetkov},
      year={2024},
      eprint={2305.09955},
      archivePrefix={arXiv},
      primaryClass={cs.CL}
}

@misc{selfreflect,
      title={Language Models (Mostly) Know What They Know}, 
      author={Saurav Kadavath and Tom Conerly and Amanda Askell and Tom Henighan and Dawn Drain and Ethan Perez and Nicholas Schiefer and Zac Hatfield-Dodds and Nova DasSarma and Eli Tran-Johnson and Scott Johnston and Sheer El-Showk and Andy Jones and Nelson Elhage and Tristan Hume and Anna Chen and Yuntao Bai and Sam Bowman and Stanislav Fort and Deep Ganguli and Danny Hernandez and Josh Jacobson and Jackson Kernion and Shauna Kravec and Liane Lovitt and Kamal Ndousse and Catherine Olsson and Sam Ringer and Dario Amodei and Tom Brown and Jack Clark and Nicholas Joseph and Ben Mann and Sam McCandlish and Chris Olah and Jared Kaplan},
      year={2022},
      eprint={2207.05221},
      archivePrefix={arXiv},
      primaryClass={cs.CL}
}

@misc{instr,
      title={Training language models to follow instructions with human feedback}, 
      author={Long Ouyang and Jeff Wu and Xu Jiang and Diogo Almeida and Carroll L. Wainwright and Pamela Mishkin and Chong Zhang and Sandhini Agarwal and Katarina Slama and Alex Ray and John Schulman and Jacob Hilton and Fraser Kelton and Luke Miller and Maddie Simens and Amanda Askell and Peter Welinder and Paul Christiano and Jan Leike and Ryan Lowe},
      year={2022},
      eprint={2203.02155},
      archivePrefix={arXiv},
      primaryClass={cs.CL}
}

@inproceedings{hidden,
    title = "The Curious Case of Hallucinatory (Un)answerability: Finding Truths in the Hidden States of Over-Confident Large Language Models",
    author = "Slobodkin, Aviv  and
      Goldman, Omer  and
      Caciularu, Avi  and
      Dagan, Ido  and
      Ravfogel, Shauli",
    editor = "Bouamor, Houda  and
      Pino, Juan  and
      Bali, Kalika",
    booktitle = "Proceedings of the 2023 Conference on Empirical Methods in Natural Language Processing",
    month = dec,
    year = "2023",
    address = "Singapore",
    publisher = "Association for Computational Linguistics",
    url = "https://aclanthology.org/2023.emnlp-main.220",
    doi = "10.18653/v1/2023.emnlp-main.220",
    pages = "3607--3625",
}

@misc{holisticeval,
      title={Holistic Evaluation of Language Models}, 
      author={Percy Liang and Rishi Bommasani and Tony Lee and Dimitris Tsipras and Dilara Soylu and Michihiro Yasunaga and Yian Zhang and Deepak Narayanan and Yuhuai Wu and Ananya Kumar and Benjamin Newman and Binhang Yuan and Bobby Yan and Ce Zhang and Christian Cosgrove and Christopher D. Manning and Christopher Ré and Diana Acosta-Navas and Drew A. Hudson and Eric Zelikman and Esin Durmus and Faisal Ladhak and Frieda Rong and Hongyu Ren and Huaxiu Yao and Jue Wang and Keshav Santhanam and Laurel Orr and Lucia Zheng and Mert Yuksekgonul and Mirac Suzgun and Nathan Kim and Neel Guha and Niladri Chatterji and Omar Khattab and Peter Henderson and Qian Huang and Ryan Chi and Sang Michael Xie and Shibani Santurkar and Surya Ganguli and Tatsunori Hashimoto and Thomas Icard and Tianyi Zhang and Vishrav Chaudhary and William Wang and Xuechen Li and Yifan Mai and Yuhui Zhang and Yuta Koreeda},
      year={2023},
      eprint={2211.09110},
      archivePrefix={arXiv},
      primaryClass={cs.CL}
}

@article{mtlearner,
  author = {Radford, Alec and Wu, Jeffrey and Child, Rewon and Luan, David and Amodei, Dario and Sutskever, Ilya and others},
  journal = {OpenAI blog},
  number = 8,
  pages = 9,
  title = {Language models are unsupervised multitask learners},
  volume = 1,
  year = 2019
}

@misc{tian2023just,
      title={Just Ask for Calibration: Strategies for Eliciting Calibrated Confidence Scores from Language Models Fine-Tuned with Human Feedback}, 
      author={Katherine Tian and Eric Mitchell and Allan Zhou and Archit Sharma and Rafael Rafailov and Huaxiu Yao and Chelsea Finn and Christopher D. Manning},
      year={2023},
      eprint={2305.14975},
      archivePrefix={arXiv},
      primaryClass={cs.CL}
}

@misc{cobbe2021training,
      title={Training Verifiers to Solve Math Word Problems}, 
      author={Karl Cobbe and Vineet Kosaraju and Mohammad Bavarian and Mark Chen and Heewoo Jun and Lukasz Kaiser and Matthias Plappert and Jerry Tworek and Jacob Hilton and Reiichiro Nakano and Christopher Hesse and John Schulman},
      year={2021},
      eprint={2110.14168},
      archivePrefix={arXiv},
      primaryClass={cs.LG}
}

@article{surveyodqa,
  title={The state of the art in open domain complex question answering: a survey},
  author={Etezadi, Romina and Shamsfard, Mehrnoush},
  journal={Applied Intelligence},
  volume={53},
  number={4},
  pages={4124--4144},
  year={2023},
  publisher={Springer}
}

@misc{llm-arith-1,
      title={MathPrompter: Mathematical Reasoning using Large Language Models}, 
      author={Shima Imani and Liang Du and Harsh Shrivastava},
      year={2023},
      eprint={2303.05398},
      archivePrefix={arXiv},
      primaryClass={cs.CL}
}

@misc{llm-arith-2,
      title={ArthModel: Enhance Arithmetic Skills to Large Language Model}, 
      author={Yingdi Guo},
      year={2023},
      eprint={2311.18609},
      archivePrefix={arXiv},
      primaryClass={cs.CL}
}

@misc{llm-arith-3,
      title={PAL: Program-aided Language Models}, 
      author={Luyu Gao and Aman Madaan and Shuyan Zhou and Uri Alon and Pengfei Liu and Yiming Yang and Jamie Callan and Graham Neubig},
      year={2023},
      eprint={2211.10435},
      archivePrefix={arXiv},
      primaryClass={cs.CL}
}

@misc{llm-arith-4,
      title={Arithmetic with Language Models: from Memorization to Computation}, 
      author={Davide Maltoni and Matteo Ferrara},
      year={2024},
      eprint={2308.01154},
      archivePrefix={arXiv},
      primaryClass={cs.AI}
}

\end{document}